\documentclass[10pt,journal,compsoc]{IEEEtran}

\usepackage{amsmath,amssymb}
\usepackage{float}
\usepackage{caption}
\usepackage{mathtools}
\usepackage{colortbl}
\usepackage{hyperref}
\usepackage{mathabx}
\usepackage{multirow}
\usepackage{nicefrac}
\usepackage{graphicx}
\usepackage{booktabs}
\usepackage{braket}
\usepackage{enumitem}
\usepackage{pifont}
\usepackage{subfigure}
\usepackage{xcolor}

\ifCLASSOPTIONcompsoc
  \usepackage[nocompress]{cite}
\else
  \usepackage{cite}
\fi

\newcommand*{\recommendbox}[1]{\colorbox{light-gray}{\parbox{.98\linewidth}{#1}}}
\newcommand{\tablerecommendations}{
\begin{table*}
\label{tab:recommendations}
\recommendbox{
\centering
\vspace{0.1cm}
\begin{tabular}{p{0.45\linewidth}p{0.49\linewidth}}
\multicolumn{2}{c}{\textsc{Major findings of our performance evaluation on Class-Incremental Learning}\vspace{0.1cm}}\\
\begin{minipage}[t]{\linewidth} 
  \begin{itemize}
    \item For exemplar-free \mbox{class-IL}, data regularization methods outperform weight regularization methods (see Table~\ref{tab:regul-methods}).
    \item Finetuning with exemplars (FT-E) yields a good baseline that outperforms more complex methods on several experimental settings (see Figs.~\ref{fig:vggface2} and~\ref{fig:multi_plot}).
    \item Weight regularization combines better with exemplars than data regularization for some scenarios (see Figs.~\ref{fig:cifar100_fixd_plot_cifar100_fixd_lft_plot},~\ref{fig:vggface2},~\ref{fig:imagenet_plot}).
  \end{itemize} 
\end{minipage} &
\begin{minipage}[t]{\linewidth} 
  \begin{itemize}
    \item Random exemplar sampling yields results competitive with herding, especially for shorter sequences (see Table~\ref{tab:ex-sampling}).
    \item Methods that explicitly address task-recency bias outperform those that do not.
    \item Network architecture greatly influences the performance of \mbox{class-IL} methods, in particular the presence or absence of skip connections has a significant impact (see Fig.~\ref{fig:diff-nets}).
  \end{itemize} 
\end{minipage} \\
\end{tabular}
\vspace{0.1cm}
}
\end{table*}
}

\newcommand{\figcrosstask}[2]{
\begin{figure}[tb]
\begin{center}
  \includegraphics[width=#1\linewidth]{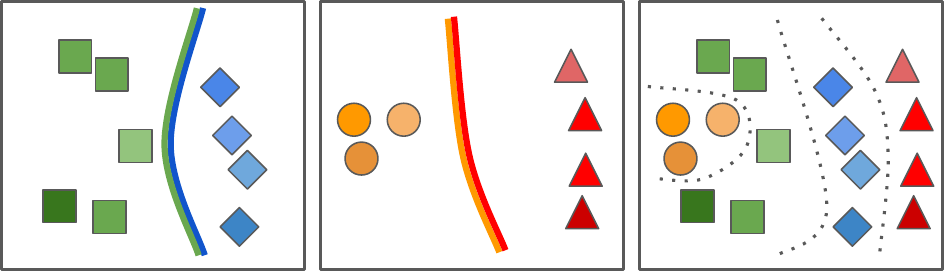}
  \caption{#2}
  \label{fig:cross-task}
\end{center}
\end{figure}
}

\newcommand{\figCM}[2]{
\begin{figure}[tb]
\begin{center}
  \includegraphics[width=#1\linewidth]{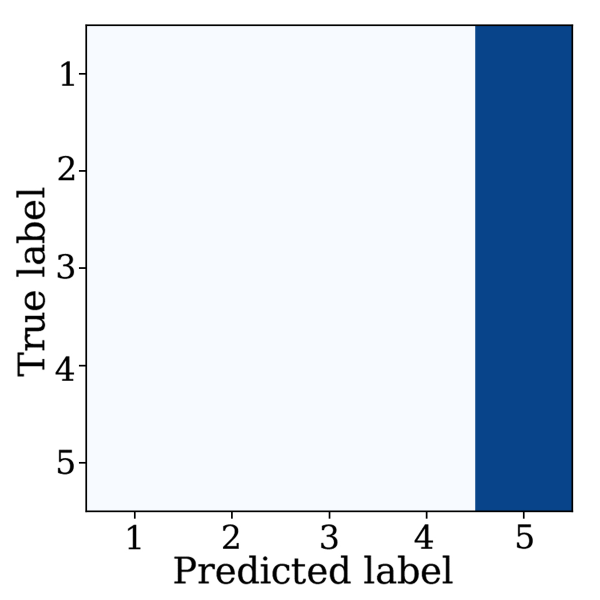}\qquad
  \includegraphics[width=#1\linewidth]{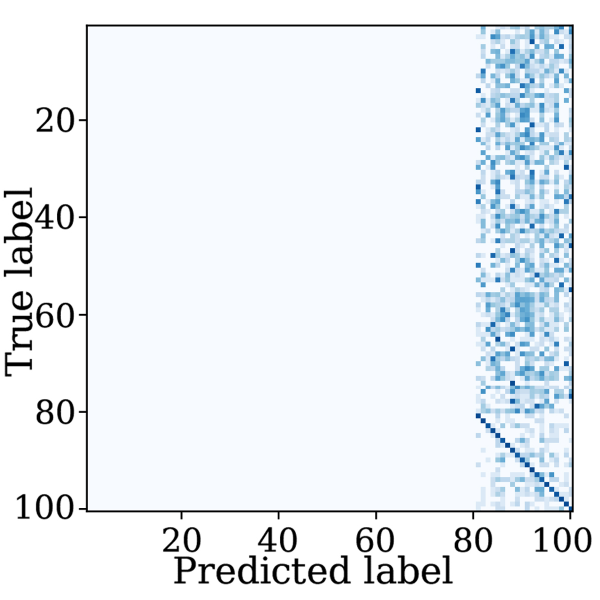} \\
    \includegraphics[width=#1\linewidth]{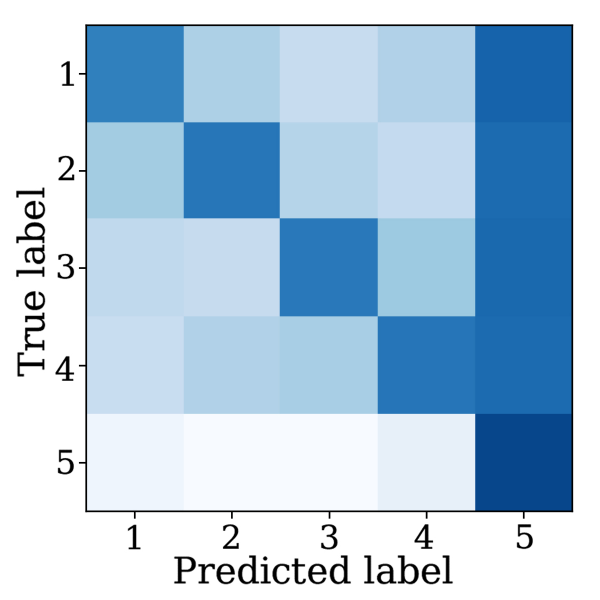}\qquad
  \includegraphics[width=#1\linewidth]{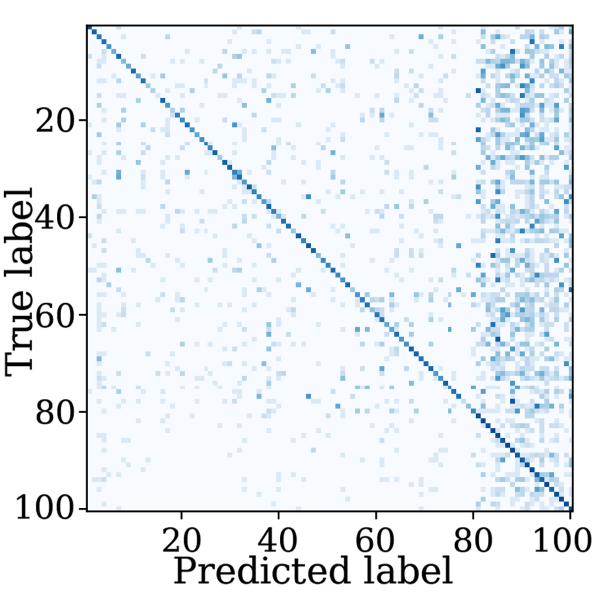}
  \caption{#2}
  \label{fig:cm}
\end{center}
\end{figure}
}

\newcommand{\figbiasWeights}[2]{
\begin{figure}[tb]
\begin{center}
  \includegraphics[width=#1\linewidth]{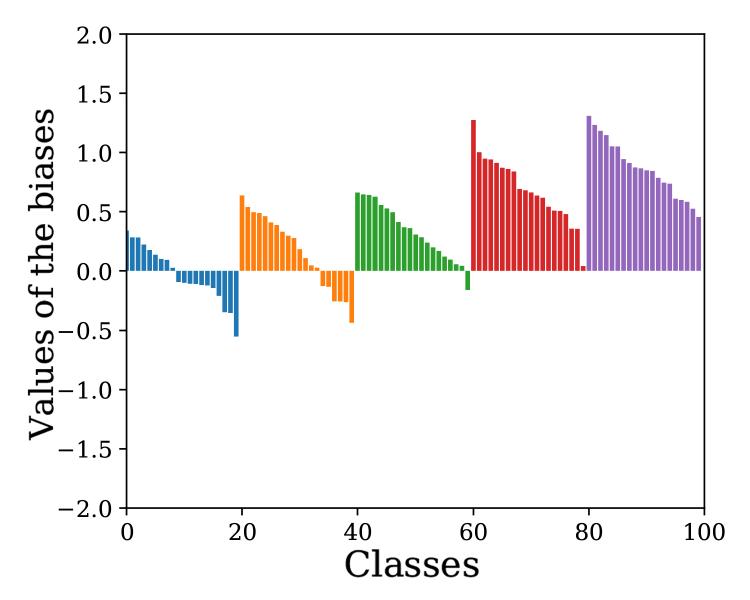}
  \includegraphics[width=#1\linewidth]{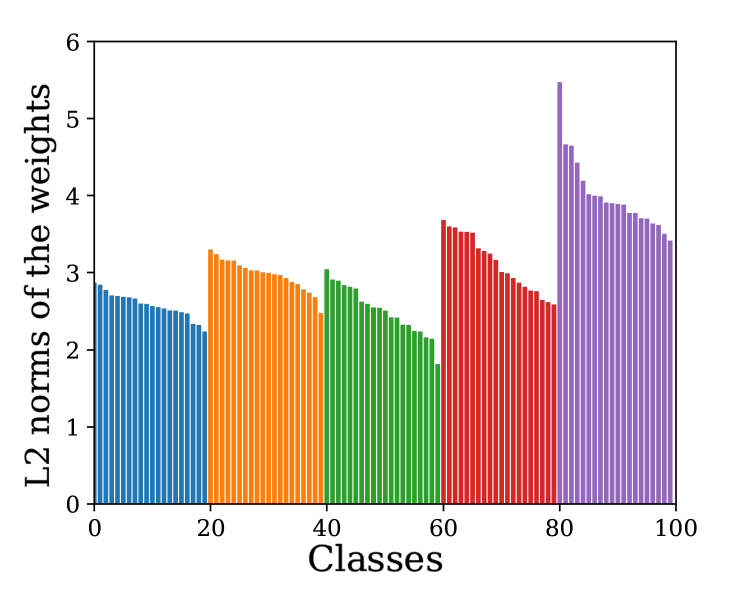}
  \caption{#2}
  \label{fig:bias_icarl}
\end{center}
\end{figure}
}

\newcommand{\tabbaselines}{
\begin{table}[tb]
\caption{Average accuracy for different baseline variants on \mbox{CIFAR-100 (10/10)}. E denotes using 2,000 exemplars (fixed memory) or 20 exemplars per class (grow) selected with herding. All baselines start with 80.7 accuracy after task 1.}
\label{tab:baselines}
\vspace{-1em}
\setlength{\tabcolsep}{4pt}
\begin{center}
\begin{tabular}{lccccccccc}
\toprule
 & T2 & T3 & T4 & T5 & T6 & T7 & T8 & T9 & T10 \\
\midrule
FT       & 33.9 & 27.9 & 19.1 & 17.7 & 12.2 & 11.6 & 10.2 & 9.0 & 7.9 \\
FT+      & 38.2 & 31.0 & 22.6 & 18.6 & 15.7 & 14.4 & 13.1 & 11.6 & 10.1 \\
FZ       & 24.1 & 18.4 & 12.8 & 12.7 & 9.2 & 8.2 & 7.8 & 6.3 & 5.3 \\
FZ+      & \textbf{42.2} & \textbf{31.3} & \textbf{24.5} & \textbf{23.1} & \textbf{20.5} & \textbf{18.3} & \textbf{17.0} & \textbf{15.6} & \textbf{14.4} \\
\midrule
FT-E (fixed) & \textbf{65.7} & \textbf{61.7} & \textbf{55.0} & \textbf{51.7} & \textbf{48.3} & \textbf{46.2} & \textbf{41.1} & \textbf{38.7} & \textbf{37.9} \\
FT-E (grow)  & 49.9 & 46.0 & 36.9 & 38.9 & 37.1 & 37.0 & 34.5 & 34.5 & 34.6 \\
FZ-E (fixed) & 50.0 & 37.1 & 26.1 & 24.2 & 19.5 & 19.4 & 15.3 & 14.3 & 11.3 \\
FZ-E (grow)  & 40.5 & 31.2 & 22.0 & 20.9 & 16.6 & 17.6 & 13.7 & 13.8 & 11.3 \\
\bottomrule
\end{tabular}
\end{center}
\end{table}
}

\newcommand{\tabregularizationmethods}{
\begin{table*}[tb]
\caption{Average accuracy for regularization-based methods on CIFAR-100 (10/10) on ResNet-32 trained from scratch.}
\label{tab:regul-methods}
\vspace{-0.5em}
\begin{center}
\begin{tabular}{cc@{\hspace{1cm}}cccccccc}
\toprule
 & avg. acc. after & FT+ & LwF & EWC & PathInt & MAS & RWalk & LwM \\
\midrule
\multirow{3}{*}{\shortstack{No exemplars\\ (task-IL)}}
 & task 2  & 59.8 & 72.0 & 61.5 & 63.8 & 63.4 & 63.4 & \textbf{74.2} \\
 & task 5  & 49.8 & \textbf{76.7} & 60.2 & 57.3 & 61.8 & 56.3 & 76.2 \\
 & task 10 & 38.3 & \textbf{76.6} & 56.7 & 53.1 & 58.6 & 49.3 & 70.4 \\
\midrule
\multirow{3}{*}{\shortstack{No exemplars\\ (class-IL)}}
 & task 2  & 38.2 & 55.4 & 39.8 & 41.2 & 39.9 & 40.3 & \textbf{57.8} \\
 & task 5  & 18.6 & \textbf{41.6} & 21.9 & 23.5 & 22.1 & 22.9 & 37.4 \\
 & task 10 & 10.1 & \textbf{30.2} & 13.1 & 13.6 & 13.9 & 14.0 & 21.9 \\
\midrule
 & avg. acc. after & FT-E & LwF-E & EWC-E & PathInt-E & MAS-E & RWalk & LwM-E \\
\midrule
\multirow{3}{*}{\shortstack{2,000 exemplars\\ fixed memory (class-IL)}}
 & task 2  & \textbf{65.7} & 63.4 & 61.5 & 56.8 & 57.6 & 56.9 & 65.5 \\
 & task 5  & 51.7 & 46.2 & 42.7 & 34.5 & 29.3 & 36.5 & \textbf{52.7} \\
 & task 10 & \textbf{37.9} & 30.8 & 28.1 & 18.5 & 18.9 & 22.7 & 37.4 \\
\midrule
\multirow{3}{*}{\shortstack{20 exemplars per class\\ growing memory (class-IL)}}
 & task 2  & 49.9 & 51.0 & 47.9 & 45.1 & 45.3 & 44.1 & \textbf{53.7} \\
 & task 5  & 38.9 & 32.6 & 32.1 & 26.3 & 22.9 & 27.0 & \textbf{39.4} \\
 & task 10 & \textbf{34.6} & 27.2 & 25.4 & 17.3 & 15.9 & 20.3 & 32.3 \\
\bottomrule
\end{tabular}
\end{center}
\end{table*}
}

\newcommand{\tabexemplarsampling}{
\begin{table}[tb]
\setlength\tabcolsep{4pt}
\caption{CIFAR-100 (10/10) for different sampling strategies with fixed memory of 2,000 exemplars on ResNet-32.}
\label{tab:ex-sampling}
\vspace{-0.5em}
\begin{center}
\resizebox{\linewidth}{!}{
\begin{tabular}{c@{\hspace{0.5cm}}c@{\hspace{1cm}}ccccc}
\toprule
avg. acc. & sampling & \multirow{2}{*}{FT-E} & \multirow{2}{*}{LwF-E} & \multirow{2}{*}{EWC-E} & \multirow{2}{*}{EEIL} & \multirow{2}{*}{BiC} \\
after & strategy & & & & & \\
\midrule
\multirow{4}{*}{task 2}  & random   & \textbf{65.2} & \textbf{61.5} & \textbf{59.4} & 64.6 & \textbf{65.3} \\
                         & herding  & 64.0 & \textbf{61.5} & 58.9 & \textbf{65.1} & 65.0 \\
                         & entropy  & 61.1 & 60.6 & 57.7 & 62.7 & 62.9 \\
                         & distance & 60.7 & 59.7 & 56.8 & 63.8 & 62.9 \\
\midrule
\multirow{4}{*}{task 5}  & random   & 49.4 & 45.2 & 39.7 & 54.5 & 54.2 \\
                         & herding  & \textbf{51.4} & \textbf{45.9} & \textbf{40.8} & \textbf{55.7} & \textbf{54.6} \\
                         & entropy  & 39.7 & 38.2 & 32.6 & 48.4 & 49.6 \\
                         & distance & 39.8 & 37.9 & 30.5 & 47.0 & 47.9 \\
\midrule
\multirow{4}{*}{task 10} & random   & 36.7 & 29.0 & 25.1 & 39.7 & 41.3 \\
                         & herding  & \textbf{37.8} & \textbf{31.0} & \textbf{26.6} & \textbf{41.8} & \textbf{42.0} \\
                         & entropy  & 23.6 & 19.3 & 15.1 & 30.4 & 33.5 \\
                         & distance & 22.1 & 19.2 & 12.6 & 28.7 & 31.1 \\
\bottomrule
\end{tabular}}
\end{center}
\vspace{-0.5em}
\end{table}
}

\newcommand{\figBiasCorrection}{
\begin{figure*}[tb]
\centering
  \begin{tabular}{ccccc}
  \includegraphics[width=0.18\linewidth]{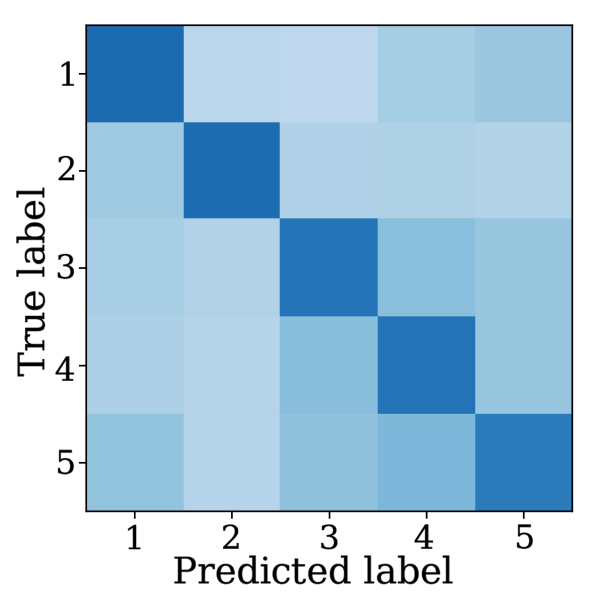} &
  \includegraphics[width=0.18\linewidth]{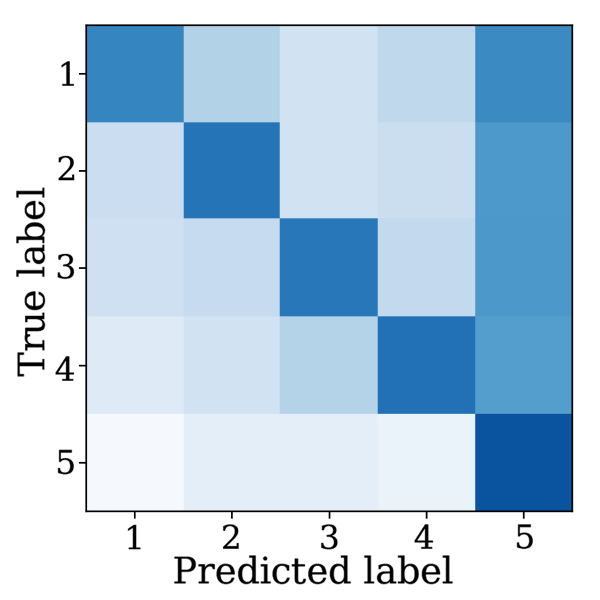} &
  \includegraphics[width=0.18\linewidth]{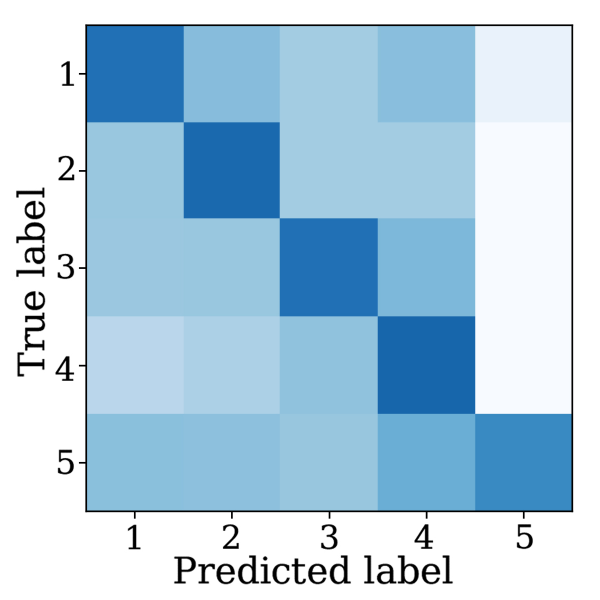} &
  \includegraphics[width=0.18\linewidth]{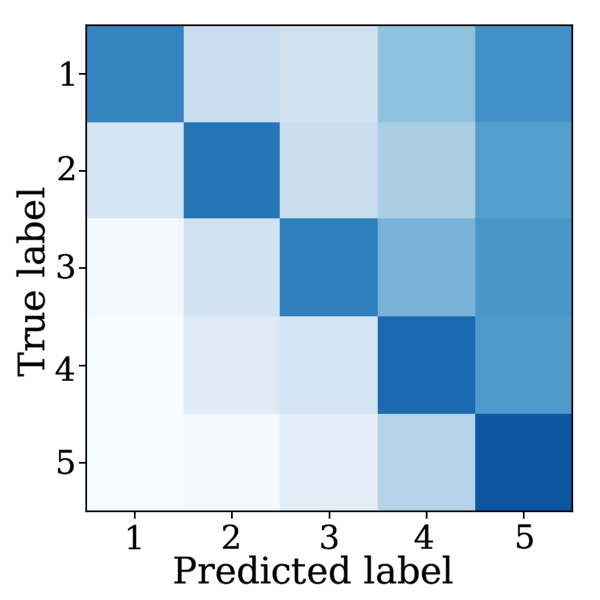} &
  \includegraphics[width=0.18\linewidth]{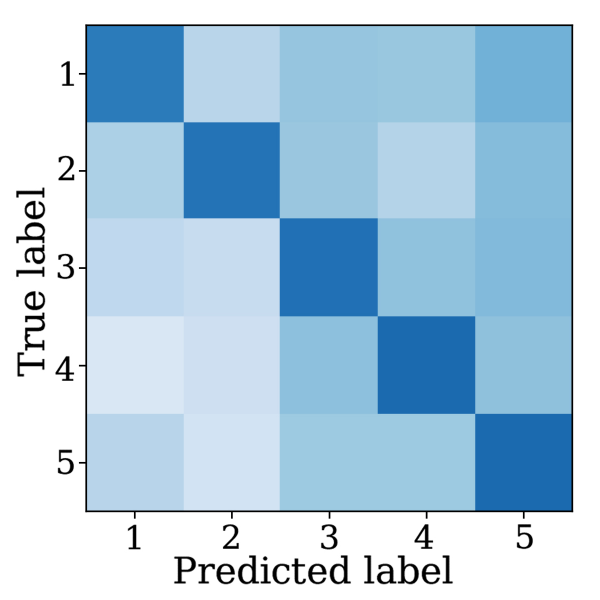} \\
  iCaRL & EEIL & BiC & LUCIR & IL2M \\
  \end{tabular}
  \caption{Task confusion matrices for CIFAR-100 (5/20) with 2,000 exemplars selected with herding.}
  \label{fig:bias-correction-exp}
\end{figure*}
}

\newcommand{\figInterspersed}[1]{
\begin{figure*}[tb]
\begin{center}
  \includegraphics[trim=0 25 0 35, clip, width=1.0\linewidth]{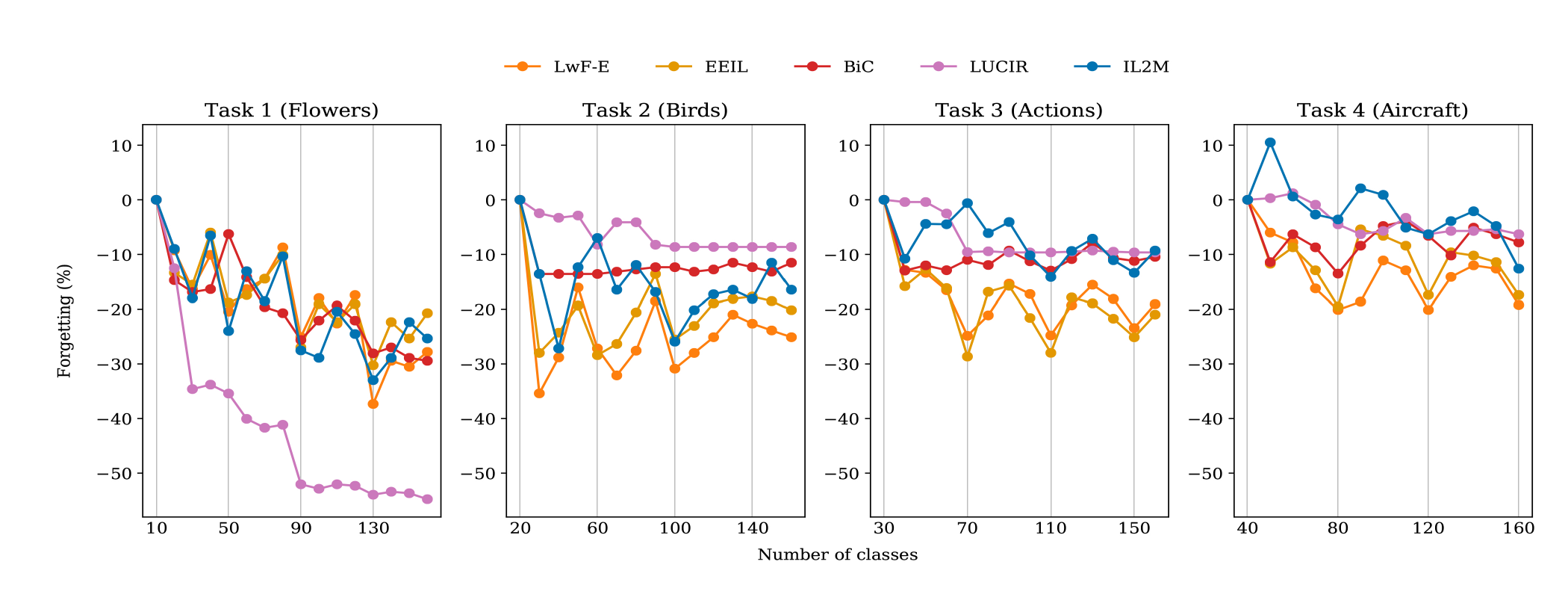}
  \caption{#1}
  \label{fig:interspersed_forg}
\end{center}
\vspace{-0.5em}
\end{figure*}
}

\newcommand{\figdifferentnetworks}[2]{
\begin{figure}[tb]
\vspace{-0.3em}
\begin{center}
  \includegraphics[trim=#1, clip, width=1.0\linewidth]{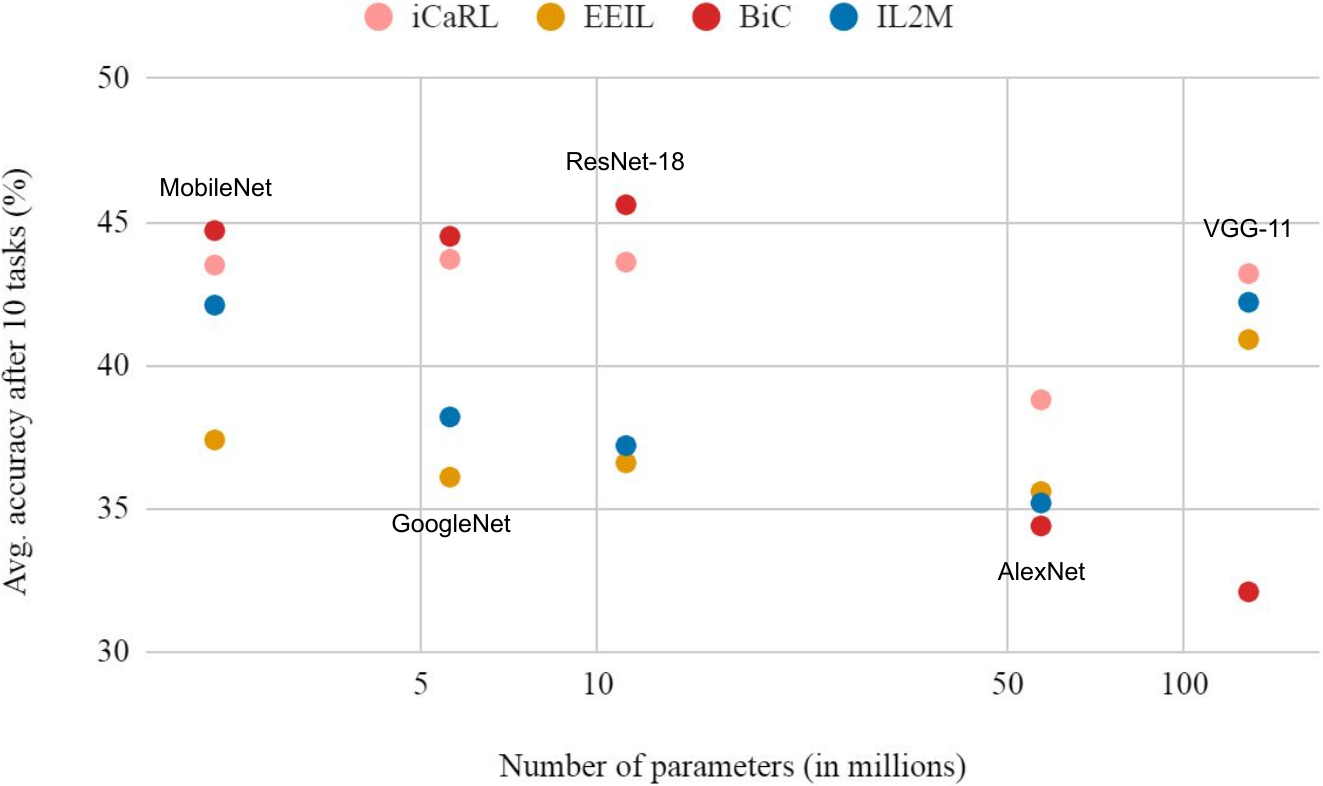}
  \caption{#2}
  \label{fig:diff-nets}
\end{center}
\vspace{-0.7em}
\end{figure}
}

\newcommand{\figExpSingleColumn}[4]{
\begin{figure}[tb]
\begin{center}
  \includegraphics[trim=#2, clip, width=#1\linewidth]{./figures/#3.pdf}
  \caption{#4}
  \label{fig:#3}
\end{center}
\end{figure}
}
\newcommand{\figExpDoubleColumnTwoImage}[4]{
\begin{figure*}[tb]
\begin{center}
    \subfigure{%
  {\centering\includegraphics[trim=0 12 0 23, clip, width=#1\linewidth]{./figures/#2.pdf}}%
  \label{fig:#2}
  }\,
  \subfigure{%
  {\centering\includegraphics[trim=0 12 0 5, clip, width=#1\linewidth]{./figures/#3.pdf}}%
  \label{fig:#3}
  }
  \caption{#4}
  \label{fig:#2_#3}
\end{center}
\end{figure*}
}

\newcommand{\tabexemplarsamplinggrowing}{
\begin{table}[tb]
\setlength\tabcolsep{4pt}
\caption{CIFAR-100 (11/50-5) with different sampling strategies and growing memory of 20 exemplars per class on ResNet-32 trained from scratch.}
\label{tab:ex-sampling-2}
\begin{center}
\begin{tabular}{c@{\hspace{0.5cm}}c@{\hspace{1cm}}ccccc}
\toprule
acc. & sampling & \multirow{2}{*}{FT-E} & \multirow{2}{*}{LwF-E} & \multirow{2}{*}{EWC-E} & \multirow{2}{*}{EEIL} & \multirow{2}{*}{BiC} \\
after & strategy & & & & & \\
\midrule
\multirow{6}{*}{task 2} & random       & 42.4 & 49.0 & \textbf{47.2} & 44.5 & 55.5 \\
                        & herding      & \textbf{48.0} & \textbf{51.7} & 45.1 & \textbf{47.9} & 53.5 \\
                        & entropy      & 39.6 & 43.6 & 38.6 & 38.4 & 46.1 \\
                        & distance     & 36.0 & 44.0 & 33.3 & 37.4 & 43.6 \\
                        & inv-entropy  & 41.4 & 44.5 & 45.5 & 43.3 & \textbf{55.6} \\
                        & inv-distance & 44.3 & 48.2 & 43.9 & 40.3 & 47.9 \\
\midrule
\multirow{6}{*}{task 5} & random       & \textbf{38.5} & 34.2 & 30.4 & \textbf{41.3} & 43.2 \\
                        & herding      & 36.5 & \textbf{36.6} & \textbf{34.1} & 40.8 & \textbf{44.6} \\
                        & entropy      & 27.3 & 24.4 & 20.2 & 28.2 & 31.4 \\
                        & distance     & 25.1 & 25.2 & 20.0 & 27.6 & 31.2 \\
                        & inv-entropy  & 34.5 & 32.4 & 30.0 & 35.9 & 41.6 \\
                        & inv-distance & 33.1 & 32.5 & 30.0 & 37.0 & 38.3 \\
\midrule
\multirow{6}{*}{task 10} & random       & \textbf{32.5} & 26.0 & 22.7 & 37.3 & 36.1 \\
                         & herding      & 32.0 & \textbf{26.3} & \textbf{23.6} & \textbf{38.8} & \textbf{39.1} \\
                         & entropy      & 16.1 & 14.8 & 10.7 & 23.0 & 25.9 \\
                         & distance     & 17.1 & 13.5 &  8.5 & 23.0 & 22.7 \\
                         & inv-entropy  & 28.7 & 22.2 & 21.8 & 30.1 & 32.8 \\
                         & inv-distance & 29.2 & 23.3 & 20.6 & 27.1 & 35.4 \\
\bottomrule
\end{tabular}
\end{center}
\end{table}
}

\newcommand{\tabexemplarsamplingstats}{
\begin{table}[tb]
\setlength\tabcolsep{4pt}
\caption{CIFAR-100 (10/10) with different sampling strategies and fixed memory of 2,000 exemplars on ResNet-32 trained from scratch.}
\label{tab:ex-sampling-3}
\begin{center}
\resizebox{\linewidth}{!}{
\begin{tabular}{ccccccc}
\toprule
acc. & samp & \multirow{2}{*}{FT-E} & \multirow{2}{*}{LwF-E} & \multirow{2}{*}{EWC-E} & \multirow{2}{*}{EEIL} & \multirow{2}{*}{BiC} \\
after & strat & & & & & \\
\midrule
                         & rand & \textbf{65.2 $\pm$ 4.0} & \textbf{61.5 $\pm$ 3.2} & \textbf{59.4 $\pm$ 3.9} & $64.6 \pm 3.7$ & \textbf{65.3 $\pm$ 4.0} \\
                    task & herd & $64.0 \pm 4.8$ & \textbf{61.5 $\pm$ 3.1} & $58.9 \pm 4.5$ & \textbf{65.1 $\pm$ 4.2} & $65.0 \pm 4.2$ \\
                    2    & entr & $61.1 \pm 5.4$ & $60.6 \pm 4.1$ & $57.7 \pm 3.9$ & $62.7 \pm 5.1$ & $62.9 \pm 3.1$ \\
                         & dist & $60.7 \pm 5.7$ & $59.7 \pm 4.0$ & $56.8 \pm 4.2$ & $63.8 \pm 5.9$ & $62.9 \pm 3.6$ \\
\midrule
                         & rand & $49.4 \pm 2.9$ & $45.2 \pm 3.2$ & $39.7 \pm 3.2$ & $54.5 \pm 3.3$ & $54.2 \pm 3.4$ \\
                    task & herd & \textbf{51.4 $\pm$ 3.1} & \textbf{45.9 $\pm$ 2.8} & \textbf{40.8 $\pm$ 3.2} & \textbf{55.7 $\pm$ 3.5} & \textbf{54.6 $\pm$ 3.3} \\
                    5    & entr & $39.7 \pm 4.0$ & $38.2 \pm 3.9$ & $32.6 \pm 3.7$ & $48.4 \pm 4.4$ & $49.6 \pm 3.6$ \\
                         & dist & $39.8 \pm 4.0$ & $37.9 \pm 3.0$ & $30.5 \pm 3.0$ & $47.0 \pm 5.0$ & $47.9 \pm 4.4$ \\
\midrule
                         & rand & $36.7 \pm 1.8$ & $29.0 \pm 1.9$ & $25.1 \pm 3.0$ & $39.7 \pm 4.0$ & $41.3 \pm 3.3$ \\
                    task & herd & \textbf{37.8 $\pm$ 1.9} & \textbf{31.0 $\pm$ 2.1} & \textbf{26.6 $\pm$ 2.8} & \textbf{41.8 $\pm$ 3.7} & \textbf{42.0 $\pm$ 3.5} \\
                    10   & entr & $23.6 \pm 1.7$ & $19.3 \pm 2.5$ & $15.1 \pm 2.1$ & $30.4 \pm 3.1$ & $33.5 \pm 3.3$ \\
                         & dist & $22.1 \pm 2.1$ & $19.2 \pm 1.9$ & $12.6 \pm 2.2$ & $28.7 \pm 3.5$ & $31.1 \pm 4.3$ \\
\bottomrule
\end{tabular}}
\end{center}
\end{table}
}

\newcommand{\tabledatasets}[1]{
\begin{table}
\caption{#1}
\label{tab:datasets}
\centering
\begin{tabular}{ccccc}
\toprule
\textbf{Dataset} & \textbf{\#Train} & \textbf{\#Eval} & \textbf{\#Classes} \\
\midrule
CIFAR-100~\cite{krizhevsky2009learning}            &    50,000 & 10,000 & 100 \\
Oxford Flowers~\cite{nilsback2008automated}        &     2,040 & 6,149 & 102 \\
MIT Indoor Scenes~\cite{quattoni2009recognizing}   &     5,360 & 1,340 & 67 \\
CUB-200-2011 Birds~\cite{wah2011caltech}           &     5,994 & 5,794 & 200 \\
Stanford Cars~\cite{krause20133d}                  &     8,144 & 8,041 & 196 \\
FGVC Aircraft~\cite{maji2013fine}                  &     6,667 & 3,333 & 100 \\
Stanford Actions~\cite{yao2011human}               &     4,000 & 5,532 & 40 \\
VGGFace2~\cite{cao2018vggface2}                    &   491,746 & 50,000 & 1,000 \\
ImageNet ILSVRC2012~\cite{russakovsky2015imagenet} & 1,281,167 & 50,000 & 1,000 \\
\bottomrule
\end{tabular}
\end{table}
}

\newcommand{\figrpsnets}[1]{
\begin{table}[tb]
\caption{#1}
\label{tab:rpsnets}
\centering
\begin{tabular}{ccccc}
\toprule
 & \#paths & \#params & avg. acc. & time / epoch\\
\midrule
RPS (ResNet-18) & 8 & 89.56M & 57.0 & 36.4s\\
\midrule
\multirow{4}{*}{\shortstack{\textbf{RPS}\\ (ResNet-32)}}
 & 8 & 3.72M & 42.1 & 21.1s\\
 & 4 & 1.86M & 41.3 & 18.0s\\
 & 2 & 0.93M & 37.8 & 13.5s\\
 & 1 & 0.47M & 33.0 & 12.4s\\
\midrule
FT-E & 1 & 0.47M & \textbf{36.5} & 12.1s\\
FZ-E & 1 & 0.47M & 10.7 & 11.4s\\
LwF-E & 1 & 0.47M & 31.8 & 12.8s\\
\bottomrule
\end{tabular}
\end{table}
}

\newcommand{\figClassOrdering}{
\begin{figure*}[tb]
  \centering
  \setlength\tabcolsep{3pt}%
  \begin{tabular}{cccc}
    \includegraphics[width=0.24\linewidth]{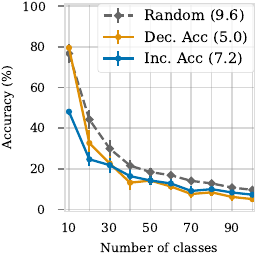} &
    \includegraphics[width=0.24\linewidth]{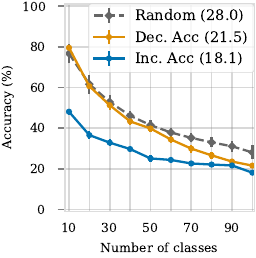} &
    \includegraphics[width=0.24\linewidth]{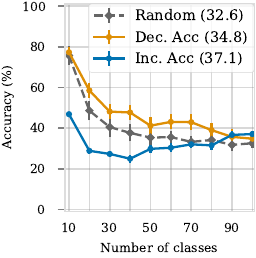} &
    \includegraphics[width=0.24\linewidth]{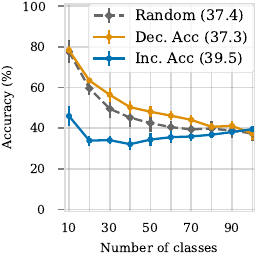} \\
    FT & LwF & FT-E & BiC \\
  \end{tabular}
  \caption{Class ordering results for CIFAR-100 on ResNet-32 trained from scratch. For FT-E and BiC, 20 exemplars per class are sampled using herding. Error bars indicate standard deviation over six runs.}
  \label{fig:class-ordering-exp}
\end{figure*}
}

\newcommand{\tabdifferentnetworks}[1]{
\begin{table}[tb]
\caption{#1}
\label{tab:diff-nets}
\centering
\resizebox{\linewidth}{!}{
\begin{tabular}{cccccc}
\toprule
 & & task 2 & task 5 & task 9 & $A_{10}$ \\
\midrule
\multirow{4}{*}{\shortstack{\textbf{AlexNet}\\ 60m params\\ 2012}}
 & iCaRL & 39.6 (-23.2) & 30.0 (-8.4)  & 33.0 (-5.2)  & \textbf{38.8} \\
 & EEIL  & 27.4 (-55.0) & 25.2 (-49.0) & 22.6 (-49.4) & 35.6 \\
 & BiC   & 30.6 (-31.8) & 26.4 (+14.0) & 21.2 (+16.8) & 34.4 \\
 & IL2M  & 27.4 (-52.4) & 21.6 (-41.2) & 44.0 (-25.2) & 35.2 \\
\midrule
\multirow{4}{*}{\shortstack{\textbf{VGG-11}\\ 133m params\\ 2014}}
 & iCaRL & 32.4 (-30.0) & 34.0 (-24.8) & 42.6 (-8.2)  & \textbf{43.2} \\
 & EEIL  & 29.6 (-56.0) & 29.0 (-50.4) & 32.8 (-45.6) & 40.9 \\
 & BiC   & 32.4 (-33.8) & 19.6 (+3.4)  & 31.0 (-3.2)  & 32.1 \\
 & IL2M  & 27.8 (-58.2) & 31.0 (-19.6) & 54.0 (-17.4) & 42.2 \\
\midrule
\multirow{4}{*}{\shortstack{\textbf{GoogleNet}\\ 6.8m params\\ 2014}}
 & iCaRL & 35.0 (-30.0) & 29.2 (-24.0) & 43.6 (-12.2) & 43.7 \\
 & EEIL  & 18.2 (-68.4) & 26.0 (-49.2) & 31.8 (-45.0) & 36.1 \\
 & BiC   & 27.2 (-51.2) & 39.8 (-14.2) & 49.0 (-4.4)  & \textbf{44.5} \\
 & IL2M  & 23.6 (-59.0) & 23.0 (-36.6) & 40.0 (-36.0) & 38.2 \\
\midrule
\multirow{4}{*}{\shortstack{\textbf{ResNet-18}\\ 11m params\\ 2015}}
 & iCaRL & 38.4 (-31.8) & 29.6 (-21.8) & 43.8 (-9.8)  & 43.6 \\
 & EEIL  & 26.0 (-59.4) & 26.8 (-52.8) & 28.2 (-48.8) & 36.6 \\
 & BiC   & 31.2 (-48.6) & 41.0 (+0.4)  & 49.4 (+4.4)  & \textbf{45.6} \\
 & IL2M  & 26.2 (-60.8) & 24.0 (-47.8) & 35.0 (-44.4) & 37.2 \\
\midrule
\multirow{4}{*}{\shortstack{\textbf{WideResNet-50}\\ 66.8m params\\ 2016}}
 & iCaRL & 34.2 (-32.4) & 33.4 (-22.2) & 41.2 (-19.8) & 42.7 \\
 & EEIL  & 25.6 (-61.4) & 25.8 (-55.6) & 23.0 (-55.6) & 37.0 \\
 & BiC   & 40.8 (-41.8) & 34.4 (-21.4) & 54.0 (-7.4)  & \textbf{45.0} \\
 & IL2M  & 27.4 (-53.0) & 29.8 (-24.0) & 41.6 (-33.2) & 40.0 \\
\midrule
\multirow{4}{*}{\shortstack{\textbf{MobileNet}\\ 4.2m params\\ 2017}}
 & iCaRL & 38.4 (-33.4) & 33.6 (-21.6) & 40.2 (-23.8) & 43.5 \\
 & EEIL  & 21.2 (-68.4) & 29.0 (-52.4) & 25.4 (-54.8) & 37.4 \\
 & BiC   & 39.4 (-44.2) & 41.2 (-11.4) & 45.2 (-14.0) & \textbf{44.7} \\
 & IL2M  & 35.0 (-46.6) & 24.2 (-24.2) & 42.6 (-30.0) & 42.1 \\
\bottomrule
\end{tabular}}
\end{table}
}

\newcommand{\tabmorecifar}[1]{
\begin{table}[tb]
\caption{#1}
\label{tab:more-cifar}
\centering
\resizebox{\linewidth}{!}{
\begin{tabular}{ccccc}
\toprule
\multirow{2}{*}{Approach} & \multicolumn{2}{c}{CIFAR-100 (10/10)} & \multicolumn{2}{c}{CIFAR-100 (11/50-5)} \\
                          & fixd mem. & grow mem. & fixd mem. & grow mem. \\
\midrule
FT-E & $37.9 \pm 2.1$ & $34.6 \pm 2.3$ & $39.0 \pm 1.7$ & $37.5 \pm 3.2$ \\
FZ-E & $11.3 \pm 0.6$ & $11.3 \pm 0.6$ & $39.8 \pm 1.3$ & $38.9 \pm 2.1$ \\
Joint & $66.3 \pm 2.2$ & $66.3 \pm 2.2$ & $65.8 \pm 2.5$ & $65.8 \pm 2.5$ \\
\midrule
EWC-E & $28.1 \pm 2.2$ & $25.4 \pm 2.1$ & $42.9 \pm 1.5$ & $41.7 \pm 1.4$ \\
MAS-E & $18.9 \pm 2.2$ & $15.9 \pm 2.2$ & $32.3 \pm 7.8$ & $33.5 \pm 5.9$ \\
PathInt-E & $18.5 \pm 1.5$ & $17.3 \pm 2.2$ & $27.4 \pm 6.8$ & $41.0 \pm 5.6$ \\
RWalk & $22.7 \pm 1.3$ & $20.3 \pm 3.0$ & $38.3 \pm 8.5$ & $35.2 \pm 8.5$ \\
LwM-E & $37.4 \pm 1.7$ & $32.3 \pm 2.4$ & $38.3 \pm 1.3$ & $35.9 \pm 2.0$ \\
DMC* & $25.9 \pm 1.3$ & $25.9 \pm 1.3$ & - & - \\
DMC & $20.6 \pm 2.8$ & $15.4 \pm 1.5$ & - & - \\
GD* & $44.6 \pm 1.0$ & $41.3 \pm 0.7$ & $44.7 \pm 0.6$ & $43.5 \pm 1.4$ \\
GD & $43.7 \pm 1.6$ & $40.5 \pm 2.0$ & $44.5 \pm 1.4$ & $42.4 \pm 1.5$ \\
LwF-E & $30.8 \pm 2.1$ & $27.2 \pm 2.0$ & $36.9 \pm 1.3$ & $34.0 \pm 1.2$ \\
iCaRL & $33.5 \pm 1.7$ & $34.6 \pm 1.3$ & $43.4 \pm 4.7$ & $42.4 \pm 5.1$ \\
EEIL & $41.9 \pm 3.0$ & $38.7 \pm 2.7$ & $42.6 \pm 1.0$ & $40.8 \pm 1.6$ \\
BiC & $42.0 \pm 2.6$ & $36.5 \pm 3.5$ & $47.0 \pm 1.1$ & $45.1 \pm 1.6$ \\
LUCIR & $36.1 \pm 3.5$ & $31.8 \pm 3.5$ & $43.4 \pm 3.0$ & $41.7 \pm 2.9$ \\
IL2M & $41.8 \pm 1.8$ & $38.5 \pm 2.2$ & $41.0 \pm 1.6$ & $40.0 \pm 1.5$ \\
\midrule
GDumb & $19.8 \pm 3.1$ & $18.6 \pm 2.9$ & - & - \\
ER & $37.1 \pm 1.3$ & $31.4 \pm 1.5$ & $39.8 \pm 1.5$ & $36.9 \pm 2.2$ \\
MIR & $36.1 \pm 0.7$ & $31.1 \pm 0.7$ & $38.2 \pm 1.0$ & $37.1 \pm 0.9$ \\
\bottomrule
\end{tabular}}
\end{table}
}

\newcommand{\tablesmalldomain}[1]{
\begin{table}
\caption{#1}
\label{tab:smalldomain}
\centering
\begin{tabular}{ccccc}
\toprule
\textbf{Approach} & \textbf{FZ-E} & \textbf{LwF-E} & \textbf{LwM-E} & \textbf{iCaRL} \\
\midrule
Accuracy at task 40 & 34.3 & 43.5 & 55.0 & 55.4 \\
\bottomrule
\end{tabular}
\end{table}
}

\newcommand{\tabmoreimagenet}[1]{
\begin{table}[tb]
\setlength\tabcolsep{4pt}
\caption{#1}
\label{tab:more-imagenet}
\centering
\resizebox{\linewidth}{!}{
\begin{tabular}{ccccccccccc}
\toprule
Approach & T1 & T2 & T3 & T4 & T5 & T6 & T7 & T8 & T9 & T10 \\
\midrule
FT-E & 78.1 & 25.3 & 20.8 & 18.4 & 17.9 & 16.6 & 15.9 & 15.3 & 15.6 & 15.6 \\
\midrule
EWC-E & 78.1 & 26.0 & 21.5 & 17.9 & 17.1 & 15.9 & 15.4 & 14.4 & 14.3 & 14.6 \\
GD    & 78.1 & 28.2 & 24.7 & 21.8 & 21.6 & 20.0 & 19.2 & 18.6 & 18.4 & 18.4\\
EEIL  & 78.1 & 27.2 & 25.9 & 23.0 & 21.0 & 20.1 & 18.8 & 18.3 & 18.6 & 18.4 \\
BiC   & 78.9 & 35.2 & 32.3 & 29.3 & 26.8 & 24.3 & 21.8 & 20.2 & 19.3 & 17.3 \\
LUCIR & 79.4 & 34.4 & 30.0 & 25.7 & 23.0 & 21.7 & 20.5 & 19.3 & 19.0 & 18.0 \\
IL2M  & 78.1 & 30.3 & 25.3 & 23.5 & 22.9 & 21.0 & 20.6 & 20.1 & 20.0 & 20.0 \\
\bottomrule
\end{tabular}}
\end{table}
}

\newcommand{\tabmoreimagenetlarge}[1]{
\begin{table}[tb]
\caption{#1}
\label{tab:more-imagenet-large}
\centering
\resizebox{\linewidth}{!}{
\begin{tabular}{cccccccccccc}
\toprule
\multirow{2}{*}{Approach} & T1 & T2 & T3 & T4 & T5 & T6 & T7 & T8 & T9 & T10 & T11 \\
 & (500) & (50) & (50) & (50) & (50) & (50) & (50) & (50) & (50) & (50) & (50) \\
\midrule
FT-E & 71.6 & 47.2 & 41.9 & 39.2 & 37.0 & 37.2 & 33.6 & 31.6 & 32.4 & 31.6 & 31.3 \\
\midrule
EWC-E & 71.6 & 53.4 & 50.0 & 46.4 & 44.6 & 44.3 & 41.0 & 40.3 & 38.6 & 35.7 & 34.6 \\
GD    & 71.6 & 55.0 & 49.3 & 46.7 & 45.1 & 43.2 & 39.7 & 39.0 & 39.4 & 38.8 & 35.9 \\
LwF-E & 71.6 & 54.9 & 45.5 & 40.7 & 37.0 & 35.4 & 32.1 & 30.1 & 29.7 & 29.5 & 29.3 \\
EEIL  & 71.6 & 53.7 & 45.6 & 42.4 & 42.5 & 40.9 & 38.6 & 37.3 & 35.1 & 35.6 & 34.9 \\
BiC   & 71.6 & 66.9 & 62.9 & 59.5 & 55.9 & 51.8 & 48.7 & 45.7 & 42.7 & 39.3 & 35.5 \\
IL2M  & 71.6 & 53.5 & 47.0 & 44.8 & 41.8 & 41.0 & 37.7 & 38.0 & 36.8 & 35.9 & 34.1 \\
\bottomrule
\end{tabular}}
\end{table}
}

\newcommand{\tabsamplingstats}[1]{
\begin{table}[tb]
\caption{#1}
\label{tab:sampling-stats}
\centering
\resizebox{\linewidth}{!}{
\begin{tabular}{ccccccc}
\toprule
\multirow{2}{*}{Approach} & \multicolumn{2}{c}{after task 2} & \multicolumn{2}{c}{after task 5} & \multicolumn{2}{c}{after task 10} \\
 & stat. & $p$-value & stat. & $p$-value & stat. & $p$-value \\
\midrule
FT-E  & 55.5 & 0.675 & 30.0 & 0.070 & 36.5 & 0.163 \\
LwF-E & 53.0 & 0.604 & 43.0 & 0.312 & 26.5 & 0.041 \\
EWC-E & 51.5 & 0.560 & 38.5 & 0.203 & 32.0 & 0.093 \\
EEIL  & 40.0 & 0.236 & 37.0 & 0.172 & 23.5 & 0.024 \\
BiC   & 52.0 & 0.575 & 46.5 & 0.410 & 40.0 & 0.236 \\
\bottomrule
\end{tabular}}
\end{table}
}

\newcommand{\minisection}[1]{\vspace{0.04in} \noindent {\bf #1}\ \ }
\newcommand{\myeq}{\mkern1.5mu{=}\mkern1.5mu}

\definecolor{light-gray}{gray}{0.95}
\definecolor{somegray}{rgb}{0.5, 0.5, 0.5}
\newcommand{\darkgrayed}[1]{\textcolor{somegray}{#1}}
\makeatletter
\newcommand*\titleheader[1]{\gdef\@titleheader{#1}}
\AtBeginDocument{%
  \let\st@red@title\@title
  \def\@title{%
    \vskip-1.5em
    \bgroup\normalfont\large\centering\@titleheader\par\egroup
    \vskip1.5em\st@red@title}
}
\makeatother

\titleheader{\darkgrayed{This paper has been accepted for publication at the\\
IEEE Transactions on Pattern Analysis and Machine Intelligence, 2022.
\copyright IEEE}\vspace{-1cm}}

\title{Class-incremental learning: survey  and performance evaluation on image classification}

\begin{document}

\author{Marc~Masana, Xialei~Liu, Bart{\l}omiej~Twardowski, Mikel~Menta, Andrew~D.~Bagdanov, Joost~van~de~Weijer

\thanks{Marc Masana, Xialei Liu, Bart{\l}omiej Twardowski, Mikel Menta and Joost van de Weijer are from the LAMP team at the Computer Vision Center, Barcelona, Spain (e-mail: \{mmasana, xialei, btwardowski, mikel.menta, joost\}@cvc.uab.es). Andrew D. Bagdanov is from the Media Integration and Communication Center, Florence, Italy (e-mail: andrew.bagdanov@unifi.it).}
\thanks{(Corresponding author: Xialei Liu)}
}

\markboth{}%
{Masana \MakeLowercase{\textit{et al.}}: Class-incremental learning: survey and performance evaluation on image classification}

\IEEEtitleabstractindextext{%
\begin{abstract}
For future learning systems, incremental learning is desirable because it allows for: efficient resource usage by eliminating the need to retrain from scratch at the arrival of new data; reduced memory usage by preventing or limiting the amount of data required to be stored -- also important when privacy limitations are imposed; and learning that more closely resembles human learning. The main challenge for incremental learning is catastrophic forgetting, which refers to the precipitous drop in performance on previously learned tasks after learning a new one. Incremental learning of deep neural networks has seen explosive growth in recent years. Initial work focused on task-incremental learning, where a task-ID is provided at inference time. Recently, we have seen a shift towards class-incremental learning where the learner must discriminate at inference time between all classes seen in previous tasks without recourse to a task-ID. In this paper, we provide a complete survey of existing class-incremental learning methods for image classification, and in particular, we perform an extensive experimental evaluation on thirteen class-incremental methods. We consider several new experimental scenarios, including a comparison of class-incremental methods on multiple large-scale image classification datasets, an investigation into small and large domain shifts, and a comparison of various network architectures.
\end{abstract}

}

\maketitle

\IEEEdisplaynontitleabstractindextext

\IEEEpeerreviewmaketitle

%%%%%%%%%%%%%%%%%%%%%%%%%%%%%%%%%%%%%
\IEEEraisesectionheading{\section{Introduction}
\label{sec:introduction}}
\IEEEPARstart{I}{ncremental learning} aims to develop artificially intelligent systems that can continuously learn to address new tasks from new data while preserving knowledge learned from previously learned tasks~\cite{rebuffi2017icarl, thrun1996learning}. In most incremental learning (IL) scenarios, tasks are presented to a learner in a sequence of delineated \emph{training sessions} during which only data from a single task is available for learning. After each training session, the learner should be capable of performing all previously seen tasks on unseen data. The biological inspiration for this learning model is clear, as it reflects how humans acquire and integrate new knowledge: when presented with new tasks to learn, we leverage knowledge from previous ones and integrate newly learned knowledge into previous tasks~\cite{french1999catastrophic}.

This contrasts markedly with the prevailing supervised learning paradigm in which labeled data for all tasks is jointly available during a single training session of a deep network. Incremental learners only have access to data from a single task at a time while being evaluated on all learned tasks so far. The main challenge in incremental learning is to learn from data from the current task in a way that prevents forgetting of previously learned tasks. The naive approach of finetuning, so fruitfully applied to domain transfer problems, suffers from the lack of data from previous tasks and the resulting classifier is unable to classify data from them. This drastic drop in performance on previously learned tasks is a phenomenon known as \emph{catastrophic forgetting}~\cite{goodfellow2013empirical,kirkpatrick2017overcoming,mccloskey1989catastrophic}. Incremental learning aims to prevent catastrophic forgetting, while at the same time avoiding the problem of \emph{intransigence} which inhibits adaptation to new tasks~\cite{chaudhry2018riemannian}.

\tablerecommendations

We adopt the viewpoint on incremental learning first proposed
along with the iCaRL approach~\cite{rebuffi2017icarl} and the terminology used by Van de Ven and Tolias~\cite{vandeven2019three}. In incremental learning, the training is divided into a sequence of tasks, and in any training session the learner has only access to the data of the current task, optionally, some methods can consider a small amount of stored data from previous tasks. Most early methods considered the scenario, known as \emph{task-incremental learning} (task-IL), in which the algorithm has access to a task-ID at inference time. This has the clear advantage that methods do not have to discriminate between classes coming from different tasks. More recently, methods have started addressing the more difficult scenario of \emph{class-incremental learning} (class-IL), where the learner does not have access to the \mbox{task-ID} at inference time, and therefore must be able to distinguish between all classes from all tasks (see Fig.~\ref{fig:incremental_learning}). Scenarios for which the \mbox{task-ID} is typically absent at inference time include those that incrementally increase the granularity of their capacity (e.g. detecting a growing number of object classes in images). In the last few years a wide variety of methods for \mbox{class-IL} have been proposed, and the time is ripe to provide a broad overview and experimental comparison of them.

In this survey, we set out to identify the main challenges for \mbox{class-IL}, and we organize the proposed solutions in three main categories: \emph{regularization-based} solutions that aim to minimize the impact of learning new tasks on the weights that are important for previous tasks; \emph{exemplar-based} solutions that store a limited set of exemplars to prevent forgetting of previous tasks; and solutions that directly address the problem of \emph{task-recency bias}, a phenomenon occurring in \mbox{class-IL} methods that refers to the bias towards recently-learned tasks. In addition to an overview of progress in \mbox{class-IL} in recent years, we also provide an extensive experimental evaluation of existing methods. We evaluate several of the more popular regularization methods (often proposed for task-IL) and extend them with exemplars for a more fair comparison to recently developed methods. In addition, we perform extensive experiments comparing thirteen methods on several scenarios and also evaluate \mbox{class-IL} methods on a new, more challenging multi-dataset setting. Finally, we are the first to compare these methods on a wide range of network architectures. We summarize the outcomes of our survey in the ``recommendations box'' at the top of this page. Our extensible \mbox{class-IL} evaluation framework, including code to reproduce results, is publicly available at \url{https://github.com/mmasana/FACIL}.

This paper is organized as follows. In Sec.~\ref{sec:class-incremental}, we define class-incremental learning and the main challenges which need to be addressed. In Sec.~\ref{sec:approaches}, we start by defining the scope of methods we consider for our experimental evaluation based on a list of desiderata. Then we introduce the main approaches that have been proposed for \mbox{class-IL}.  In Sec.~\ref{sec:relatedwork}, we describe related work. In Sec.~\ref{sec:exp-setup}, we outline our experimental setup and follow with an extensive experimental evaluation in Sec.~\ref{sec:results}. In Sec.~\ref{sec:trends}, we discuss several emerging trends in \mbox{class-IL} and then finish with conclusions in Sec.~\ref{sec:conclusions}.

\figExpSingleColumn{0.92}{0 0 0 0}{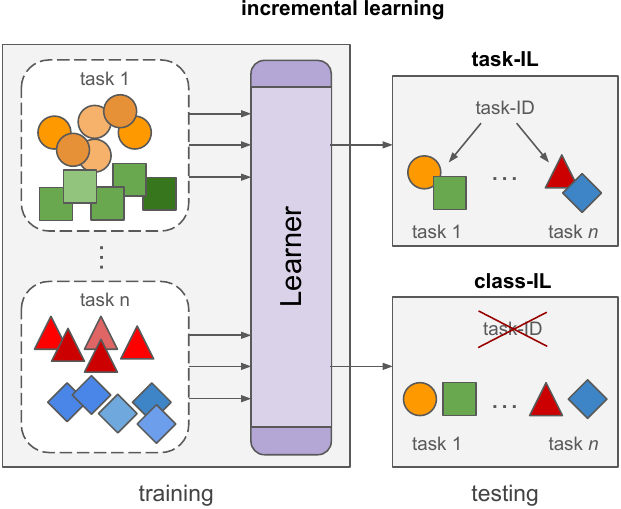}{In incremental learning, disjoint tasks are learned sequentially. Task-IL has access to the task-ID during evaluation, while the more challenging setting of \mbox{class-IL} does not. \mbox{Class-IL} is the subject of this survey.}

%%%%%%%%%%%%%%%%%%%%%%%%%%%%%%%%%%%%%
\section{Class-incremental learning}
\label{sec:class-incremental}
Incremental learning is related to several research topics, including lifelong and continual learning. Lifelong learning can be thought of as the problem of building intelligent systems capable of learning throughout an extended life-cycle in which new knowledge must be acquired and integrated to accomplish new tasks~\cite{thrun1995lifelong,chen2018lifelong}. Continual learning is one of the characteristics of a lifelong learning system, however, even this distinction is often blurred~\cite{aljundi2016expert, chaudhry2018efficient}. Continual learning and incremental learning are often used interchangeably in the literature~\cite{vandeven2019three, delange2021continual}, and neither are restricted to only fully-supervised problems. For example, Elastic Weight Consolidation (EWC)~\cite{kirkpatrick2017overcoming} was applied to Deep Reinforcement Learning for video games.

Given this lack of coherent terminology in the literature, for the purpose of this survey we define \emph{incremental learning} as continual learning systems that learn from a sequence of tasks consisting of new, supervised data.

\subsection{The practical importance of incremental learning}
The notable increase in attention IL has received in the last few years has been fueled by a demand from applications in industry and society. There are several problems for which incremental knowledge assimilation offers a solution.

\minisection{Memory restrictions:} Systems that have physical limitations for the data that they can store cannot resort to joint training strategies because they simply cannot store all seen data. Such systems can only store a limited set of examples for the tasks they perform, and therefore learning must be done incrementally. This scenario is especially common in robotics~\cite{lesort2020continual}, where robots are faced with different tasks at different times or locations, but should still be able to perform all previously learned tasks.

\minisection{Data security/privacy restrictions:} For systems that learn from data that should not be permanently stored, incremental learning can provide a solution. Government legislation could restrict data storage from clients at a central location (e.g. for applications on mobile devices). Privacy considerations are also common in health applications where legislation prevents the long-term storage of data from patients, and thus incremental learning is key~\cite{mcclure2018distributed}.

\minisection{Sustainable artificial intelligence:} The cost of training deep learning algorithms can be exorbitant. Examples include GPT-2 (1 week training on 32 TPUv3~\cite{strubell2019energy}). The carbon footprint of retraining such systems for every new data update is considerable, and will likely grow in the coming years~\cite{sharir2020cost}. Incremental learning provides algorithms that are much more computationally efficient and only require processing of new data when updating the system.

\subsection{General class-incremental learning setup}
Class-IL methods learn from a stream of data drawn from a non-stationary distribution. These methods should scale to a large number of tasks without excessive computational and memory growth. They aim to exploit knowledge from previous classes to improve learning new ones (\emph{forward transfer}), as well as exploiting new data to improve performance on previous tasks (\emph{backward transfer})~\cite{lopez2017gradient}. Our investigation focuses on class-IL scenarios in which the algorithm must learn a sequence of tasks (see Section~\ref{sec:trends} for discussion of task-free scenarios). By \emph{task}, we refer to a set of classes disjoint from classes in other (previous or future) tasks. In each \emph{training session}, the learner only has access to data from a single task. We optionally consider a small memory that can be used to store some exemplars from previous tasks. Tasks consist of a number of classes, and learners are allowed to process the training data of the current task multiple times during the training session (also called the \emph{offline learning} setting). We do not consider the online learning setting used in some papers~\cite{lopez2017gradient, mai2021online} in which each data sample is only seen once.

More formally, an incremental learning problem $\mathcal{T}$ consists of a sequence of $n$ tasks:
\begin{eqnarray}
  \mathcal{T} = [(C^1, D^1), (C^2, D^2), \ldots, (C^n, D^n)],
\end{eqnarray}
where each task $t$ is represented by a set of classes $C^t\myeq\{c^t_1,c^t_2...,c^t_{n^t}\}$ and training data $D^t$. We use $N^t$ to represent the total number of classes in all tasks up to and including task $t$: $N^t\myeq\sum_{i=1}^{t} |C^i|$.  We consider class-incremental classification problems in which $D^t\myeq\set{(\mathbf{x}_1, \mathbf{y}_1),(\mathbf{x}_2, \mathbf{y}_2),\ldots,(\mathbf{x}_{m^t}, \mathbf{y}_{m^t})}$, where ${\bf x}$ are input features for a training sample, and ${\bf y_i}\!\in\!\set{0, 1}^{N^t}$ is a one-hot ground truth label vector corresponding to $\mathbf{x}_i$. During training for task $t$, the learner only has access to $D^{t}$. In our experimental evaluation, we do not allow class overlap between tasks (i.e., $C^i\!\cap\!C^j\!=\!\emptyset$ if $i\!\neq\!j$). his setting is used in the majority of the compared methods~\cite{rebuffi2017icarl, li2017learning, hou2019learning, wu2019large, castro2018end} and we therefore adopt it, even though class overlap between tasks could occur in some real-world applications.

We consider incremental learners that are deep networks parameterized by weights $\theta$ and we use ${\mathbf{o}}(\mathbf{x}) = h(\mathbf{x}; \theta)$ to indicate the output logits of the network on input $\mathbf{x}$. We further split the neural network in a feature extractor $f$ with weights $\phi$ and linear classifier $g$ with weights $V$ according to ${\mathbf{o}}(\mathbf{x}) = g(f(\mathbf{x}; \phi);V)$. We use  $\hat{\mathbf{y}} = \sigma(h(\mathbf{x}; \theta))$ to identify the network predictions, where $\sigma$ indicates the softmax function. After training on task $t$, we evaluate the performance of the network on all classes $\bigcup_{i=1}^{t} C^i$. This contrasts with task-IL, where the task-ID $t$ is known and evaluation is only over task $C^{t}$ at inference time.

Most \mbox{class-IL} classifiers are trained with a cross-entropy loss. When training only on data from the current task $t$, we can consider two cross-entropy variants. We can consider a cross-entropy over \emph{all} classes up to the current task:
\begin{eqnarray}
\mathcal{L}_{c}(\mathbf{x}, \mathbf{y};\theta^t) = \sum_{k=1}^{N^{t}} y_k \log \frac{\exp(\mathbf{o}_k)}{\sum_{i=1}^{N^t} \exp(\mathbf{o}_i)}.
\label{eq:CE}
\end{eqnarray}
Note that in this case, since the softmax normalization is performed over \emph{all} previously seen classes from all previous tasks, errors during training are backpropagated from all outputs -- including those which do not correspond to classes belonging to the current task. Instead, we can consider only network outputs for the classes belonging to the current task $t$ and define the following cross-entropy loss~\cite{castro2018end, belouadah2019il2m}:
\begin{eqnarray}
  \mathcal{L}_{c^*}(\mathbf{x}, \mathbf{y}; \theta^t) = \sum_{k=1}^{|C^t|} y_{N^{t\textit{-}1}+k} \log \frac{\exp(\mathbf{o}_{N^{t\textit{-}1}+k})}{\sum_{i=1}^{|C^{t}|} \exp(\mathbf{o}_{N^{t\textit{-}1}+i})}
\label{eq:CE2}
\end{eqnarray}
This loss only considers the softmax-normalized predictions for classes from the \emph{current} task. As a consequence, errors are backpropagated only from the probabilities related to these classes from task $t$, leading to less forgetting (see Sec.~\ref{sec:baselines}). When using exemplars from previous tasks, it is natural to apply Eq.~\ref{eq:CE} which considers the estimated probabilities on both previous and new classes. However, in Sec.~\ref{sec:baselines}, we confirm that, when no exemplars are used, using the loss in Eq.~\ref{eq:CE2} results in less forgetting and a much stronger baseline than finetuning with Eq.~\ref{eq:CE}.

\subsection{Scope of our experimental evaluation}
The literature on IL is vast and growing, and several definitions and interpretations of \mbox{class-IL} have been proposed in recent years~\cite{delange2021continual, rebuffi2017icarl, schwarz2018progress, vandeven2019three}. In order to narrow the scope of this survey to a broad group of usefully comparable methods, we consider \mbox{class-IL} methods that are:
\begin{enumerate}
\item \textbf{Task-agnostic in evaluation}: incremental learners able to predict classes from all previously learned tasks without recourse to a task oracle at inference providing a subset of possible classes.
\item \textbf{Offline}: methods in which data is presented in \emph{training sessions} whose data is \textit{i.i.d} and can be processed multiple times before moving on to the next task.
\item \textbf{Fixed network architecture}: methods using a fixed architecture for all tasks, without adding significant amount of parameters to the architecture for new tasks.
\item \textbf{Tabula rasa}: incremental learners trained from scratch which do not require pretraining on large labeled datasets. This property eliminates potential biases introduced by the class distributions seen during pretraining and any exploits derivable from that knowledge.
\item \textbf{Mature}: methods applicable to complex image classification problems.
\end{enumerate}
Property 1 distinguishes class-IL from task-IL~\cite{delange2021continual}. Properties 2--5 are characteristics that we use to select methods for our evaluation: property 2 excludes online methods~\cite{lopez2017gradient,mai2021online}, property 3 excludes growing architectures~\cite{rusu2016progressive, rajasegaran2019random}, property 4 excludes methods that require a pretrained network~\cite{hayes2020remind, mallya2018packnet}, and property 5 excludes methods that are not yet scalable to larger datasets~\cite{shin2017continual,nguyen2018variational}.

Finally, we consider one additional (optional) property:
\begin{enumerate}
    \setcounter{enumi}{5}
    \item \textbf{Exemplar-free}: methods not requiring storage of image data from previous tasks. This is an important characteristic of methods which should be privacy-preserving.
\end{enumerate}
Methods not requiring any data storage are seeing increased attention in a world where data privacy and security are fundamental for many users and are under increased legislative control.

\subsection{Challenges of class-incremental learning}
The fundamental obstacles to effective class-incremental learning are conceptually simple, but in practice very challenging to overcome. These challenges originate from the sequential training of tasks and the requirement that at any moment the learner must be able to classify all classes from all previously learned tasks. Incremental learning methods must balance retaining knowledge from previous tasks while learning new knowledge for the current task. This problem is called the \emph{stability-plasticity dilemma}~\cite{mermillod2013stability}. A naive approach to \mbox{class-IL} which focuses solely on learning the new task will suffer from \emph{catastrophic forgetting}: a drastic drop in the performance on previous tasks~\cite{goodfellow2013empirical,mccloskey1989catastrophic}. Preventing catastrophic forgetting leads to a second important problem of \mbox{class-IL}, that of \emph{intransigence}: the resistance to learn new tasks~\cite{chaudhry2018riemannian}. There are several causes of catastrophic forgetting in class-incremental learners:
\begin{itemize}
\item \textbf{Weight drift}: While learning new tasks, the network weights relevant to \emph{old} tasks are updated to minimize a loss on the \emph{new} task. As a result, performance on previous tasks suffers -- often dramatically.
\item \textbf{Activation drift}: Closely related to weight drift, changing weights result in changes to activations, and consequently in changes to the network output. Focusing on activations rather than on weights can be less restrictive since this allows weights to change as long as they result in minimal changes in layer activations.
\item \textbf{Inter-task confusion}: in \mbox{class-IL}, the objective is to discriminate all classes from all tasks. However, since classes are never jointly trained, the network weights cannot optimally discriminate all classes (see Fig.~\ref{fig:cross-task}). This holds for all layers in the network.
\item \textbf{Task-recency bias}: Separately learned tasks might have incomparable classifier outputs. Typically, the most dominant task bias is towards more recent task classes. This effect is clearly observed in confusion matrices which illustrate the tendency to miss-classify inputs as belonging to the most recently seen task (see Fig.~\ref{fig:cm}).
\end{itemize}

\figcrosstask{1.0}{A network trained continually to discriminate between task 1 (left) and task 2 (middle) is unlikely to have learned features to discriminate between the four classes (right). We call this problem \emph{inter-task confusion}.\vspace{-1em}}

\figCM{0.4}{Examples of task and class confusion matrices for Finetuning (top row) and Finetuning with 2,000 exemplars (bottom row) on \mbox{CIFAR-100}. Note the large bias towards the classes of the last task for Finetuning. By exploiting exemplars, the resulting classifier is clearly less biased.}

\noindent The first two sources of forgetting are related to network drift and have been broadly considered in the task-IL literature. Regularization-based methods either focus on preventing the drift of important weights~\cite{aljundi2018memory, chaudhry2018riemannian, kirkpatrick2017overcoming, zenke2017continual} or the drift of activations~\cite{jung2016less, li2017learning}.

The last two points are specific to \mbox{class-IL} since they have no access to a task-ID at inference time. Most research has focused on reducing task imbalance~\cite{belouadah2019il2m, hou2019learning, wu2019large}, which addresses the task-recency bias. To prevent inter-task confusion and learn representations which are optimal to discriminate between all classes, rehearsal~\cite{castro2018end, rebuffi2017icarl} or pseudo-rehearsal~\cite{shin2017continual, wu2018memory, xiang2019incremental} are commonly used.

%%%%%%%%%%%%%%%%%%%%%%%%%%%%%%%%%%%%%
\section{Approaches}
\label{sec:approaches}
In this section, we describe several approaches to address the above mentioned challenges of \mbox{class-IL}. We divide them into three main categories: regularization-based methods, rehearsal-based methods, and bias-correction methods.

\subsection{Regularization approaches}
Several approaches use regularization terms together with the classification loss in order to mitigate catastrophic forgetting. Some regularize on the weights and estimate an importance metric for each parameter in the network~\cite{aljundi2018memory, chaudhry2018riemannian, kirkpatrick2017overcoming, lee2017overcoming, liu2018rotate, zenke2017continual}, while others focus on the importance of remembering feature representations~\cite{jung2016less, li2017learning, rannen2017encoder, silver2002task, zhang2020class, lee2019overcoming}. Most of these approaches have been developed within the context of task-IL and have been reviewed by other works~\cite{delange2021continual}. Because of their importance also for \mbox{class-IL}, we discuss them briefly. Regularization of feature representations in particular is widely used in \mbox{class-IL}. Finally, we will describe several regularization techniques developed recently specifically for \mbox{class-IL}.

\minisection{Weight regularization.}
The first class of approaches focuses on preventing weight drift determined to be relevant for previous tasks. They do so by estimating a prior importance of each parameter in the network (which are assumed to be independent) after learning each task. When training on new tasks, the importance of each parameter is used to penalize changes to them. That is, in addition to the cross-entropy classification loss, an additional loss is introduced:
\begin{equation}
\mathcal{L}_{\text{reg}}(\theta^{t}) = \frac{1}{2} \sum_{i=1}^{|\theta^{t\textit{-}1}|} \Omega_{i} (\theta_{i}^{t\textit{-}1} - \theta^{t}_i)^2,
\end{equation}
where $\theta_i^t$ is weight $i$ of the network currently being trained,  $\theta^{t\textit{-}1}_i$ is the value of this parameter at the end of training on task $t\,\textit{-}\,1$, $|\theta^{t\textit{-}1}|$ is the number of weights in the network, and $\Omega_{i}$ contains importance values for each network weight.

Kirkpatrick et al.~\cite{kirkpatrick2017overcoming} proposed \emph{Elastic Weight Consolidation} (EWC) in which $\Omega_{i}$ is calculated as a diagonal approximation of the Fisher Information Matrix. However, this captures the importance of the model at the minimum after each task is learned, while ignoring the influence of those parameters along the learning trajectory in weight space. Liu et al.~\cite{liu2018rotate} improve EWC by rotating the parameter space to one that provides a better approximation of the Fisher Information Matrix. However, the model has to be extended with fixed parameters during training, which does not increase the capacity of the network but incurs in a computational and memory cost. In a similar vein of improving the approximation of the Fisher Information Matrix in EWC, Lee et al.~\cite{lee2020continual} propose an extension of the Kronecker factorization technique for block-diagonal approximation of the Fisher Information Matrix. They additionally demonstrate how such Kronecker factorizations make accommodating batch normalization possible.

In contrast, Zenke et al.~\cite{zenke2017continual} proposed the \emph{Path Integral} approach (PathInt), that accumulates the changes in each parameter online  along the entire learning trajectory. As noted by the authors, batch updates to the weights might lead to overestimating the importance, while starting from pretrained models might lead to underestimating it. To address this, \emph{Memory Aware Synapses} (MAS)~\cite{aljundi2018memory} also proposes to calculate $\Omega_{i}$ online by accumulating the sensitivity of the learned function (the magnitude of the gradient). Further, \emph{Riemanian Walk} (RWalk)~\cite{chaudhry2018riemannian} fuses both Fisher Information Matrix approximation and online path integral in a single algorithm to calculate the importance for each parameter. In addition, RWalk uses exemplars to further improve results.

\minisection{Data regularization.}
The second class of regularization-based approaches aims to prevent activation drift and is based on knowledge distillation~\cite{bucilua2006model, hinton2014distilling} which was originally designed to learn a more compact student network from a larger teacher network. Li et al.~\cite{li2017learning} proposed to use the technique to keep the representations of previous data from drifting too much while learning new tasks. Their method, called \emph{Learning without Forgetting} (LwF) applies the following loss:
\begin{equation}
\mathcal{L}_{dis}\left(\mathbf{x};\theta^t\right)=\sum_{k=1}^{N^{t\text{-}1}} \pi _k^{t\text{-}1}\left(\mathbf{x} \right) \log  \pi^{t}_k( \mathbf{x} ),
   \label{eq:LwF}
\end{equation}
where $\pi_k( \mathbf{x} )$ are temperature-scaled logits of the network:
\begin{equation}
\pi_k \left(\mathbf{x} \right) = \frac{e^{\nicefrac{ \mathbf{o}_k \left(\mathbf{x}\right)}{T}}}{\sum_{l=1}^{N^{t\text{-}1}} e^{\nicefrac{ \mathbf{o}_l \left(\mathbf{x}\right)}{T}}},
   \label{eq:temp_scale}
\end{equation}
and ${\mathbf{o}}(\mathbf{x})$ is the output of the network before the softmax is applied, and $T$ is the temperature scaling parameter. We use ${\mathbf{\pi}}^{t\text{-}1} $ to refer to the predictions of the network after training task $t\,\textit{-}\,1$. Temperature scaling was introduced to help with the problem of having the probability of the correct class too high~\cite{hinton2014distilling}.

The learning without forgetting loss in Eq.~\ref{eq:LwF} was originally proposed for a task-IL setting. However, it has since been a key ingredient of many \mbox{class-IL} methods~\cite{castro2018end, hou2019learning, liu2020mnemonics, rebuffi2017icarl, wu2019large, zhang2020class}. When the LwF method is combined with exemplars the distillation loss is typically also applied to the exemplars of previous classes~\cite{castro2018end, hou2019learning, rebuffi2017icarl, wu2019large}. Finally, some works have observed that the loss works especially well when the domain shift between tasks is small (as is typically the case for \mbox{class-IL}), however, when domain shifts are large its efficacy drops significantly~\cite{aljundi2016expert}.

A very similar approach, called \emph{less-forgetting learning} (LFL), was proposed by Jung et al.~\cite{jung2016less}. LFL preserves previous knowledge by freezing the last layer and penalizing differences between the activations before the classifier layer. However, since this can introduce larger issues when the domain shift is too large, other approaches introduced modifications to deal with it. Encoder-based lifelong learning~\cite{rannen2017encoder} extends LwF by optimizing an undercomplete autoencoder which projects features to a manifold with fewer dimensions. One autoencoder is learned per task, which makes the growth linear, although the autoencoders are small compared to the total model size.

\minisection{Recent developments in regularization.}
Several new regularization techniques have been proposed in recent work on \mbox{class-IL}. Zagoruyko and Komodakis~\cite{zagoruyko2016paying} proposed to use the attention of the teacher network to guide the student network. \emph{Learning without Memorizing} (LwM)~\cite{dhar2019learning} applies this technique to \mbox{class-IL}. The main idea is that the attention used by the network trained on previous tasks should not change while training the new task. Features contributing to the decision of a certain class label are expected to remain the same. This is enforced by the attention distillation loss:
\begin{equation}
\mathcal{L}_{AD}\left(\mathbf{x};\theta^t\right)=\left\lVert \frac{Q^{t-1}\left(\mathbf{x}\right)}{\lVert Q^{t-1}\left(\mathbf{x}\right) \rVert_2}-\frac{Q^{t}\left(\mathbf{x}\right)}{\lVert Q^{t}\left(\mathbf{x}\right) \rVert_2} \right\rVert_1,
\end{equation}
where the attention map $Q$ is given by:
\begin{equation}
Q^t\left(\mathbf{x}\right)=\text{Grad-CAM}\left( \mathbf{x}, \theta^{t}, c \right)
\end{equation}
\begin{equation}
Q^{t\text{-}1}\left(\mathbf{x}\right)=\text{Grad-CAM}\left( \mathbf{x}, \theta^{t\text{-}1}, c \right)
\end{equation}
and is generated with the Grad-CAM algorithm~\cite{selvaraju2017grad}. Grad-CAM computes the gradient with respect to a target class $c$ to produce a coarse localization map indicating the image regions which contributed to the prediction. Here, we cannot use the target class label, because this label did not exist when training the previous model $\theta^{t-1}$. Instead, the authors propose to use the previous class predicted with the highest probability to compute the attention maps: $c=\rm{argmax}\;h\left({\bf x};\theta^{t-1}\right)$.

Another recent method building upon LwF is \emph{Deep Model Consolidation} (DMC)~\cite{zhang2020class}. It is based on the observation that there exists an asymmetry between previous and new classes when training: new classes have explicit and strong supervision, whereas supervision for old classes is weaker and communicated by means of knowledge distillation. To remove this asymmetry, they propose to apply a \emph{double distillation} loss on the model $\theta^{t\text{-}1}$ trained on previous classes and a newly trained model $\theta^{t}$ for the new classes (allowing this model to forget previous tasks):
\begin{equation}
\mathcal{L}_{DD} ( \mathbf{u};\theta )=\frac{1}{N^t}\sum_{k=1}^{N^t} \left(\mathbf{o}_k ( \mathbf{u} )-\mathring{\mathbf{o}}_k ( \mathbf{u} ) \right)^2,
\end{equation}
where $\mathring{\mathbf{o}}_k$ are normalized logits:
\begin{equation}
  \mathring{\mathbf{o}}_k( \mathbf{u} ) =
\begin{cases}
  \mathbf{o}_k^{t\text{-}1}( \mathbf{u} ) - \displaystyle \frac{1}{N^{t\text{-}1}} \displaystyle \sum_{l=1}^{N^{ t\text{-}1}}{\mathbf{o}}_l^{t\text{-}1} ( \mathbf{u} ) & \mbox{if } 1\le k\le N^{t\text{-}1}\\
  \mathbf{o}_k^t ( \mathbf{u} ) - \displaystyle \frac{1}{N^t} \displaystyle \sum_{l=1}^{N^t}{\mathbf{o}}_l^{t} ( \mathbf{u} ) & \mbox{if } N^{t\text{-}1}< k\le N^{t}.
\end{cases}
\end{equation}
Here $\mathbf{o}^{t\text{-}1}(\mathbf{u})$ refers to the logits from the network trained on previous tasks, and $\mathbf{o}^{t}(\mathbf{u})$ the ones trained on the new classes. Because the algorithm does not have access to data of previous tasks, they propose to use auxiliary data $\mathbf{u}$, which can be any unlabeled data from a similar domain.

Similarly, \emph{Global Distillation} (GD)~\cite{lee2019overcoming} also proposes to use external data to distill knowledge from previous tasks. They first train a teacher on the current task and calibrate its confidence using the external data and exemplars. Then they triple-distill a new model from the teacher, the previous model, and their ensemble (plus the cross-entropy loss for current task and exemplars). The teacher and the previous task are used with both current task data and external dataset data, while the ensemble is only used with the external data. Finally, they perform finetuning while avoiding task-recency bias by weighting the loss according to the amount of data. The external dataset sampling method is based on the predictions of the model. They also propose a version that does not require external data by replacing it with stored exemplars from previous tasks.

In Hou et al.~\cite{Hou_2018_ECCV}, current task data is also not trained directly, but rather used to train an expert teacher first. The method additionally distills an old model with a reserved small fraction of previous task data to preserve the performance on old tasks, similar to LwF but using stored exemplars instead of new data. Based on an analysis of iCaRL, Javed and Shafait~\cite{javed2018revisiting} propose a \emph{dynamic threshold moving} algorithm to fix the nearest exemplar mean classifier bias trained with distillation by maintaining an up-to-date scaling vector.

Finally, the \emph{less-forget constraint}~\cite{hou2019learning} is a variant of LwF. Instead of regularizing network predictions, they propose to regularize on the cosine similarity between the L2-normalized logits of the previous and current network:
\begin{equation}
\mathcal{L}_{lf}( \mathbf{x}; \theta ) =
1 - \frac{\langle \mathbf{o}^{t\text{-}1}( \mathbf{x} ), \mathbf{o}^t ( \mathbf{x} ) \rangle}
{||\mathbf{o}^{t\text{-}1}( \mathbf{x} )||_2 ||\mathbf{o}^{t}( \mathbf{x} )||_2},
\label{eq:lf}
\end{equation}
where $\langle \cdot, \cdot \rangle$ is the inner product between vectors. This regularization is less sensitive to task imbalance because the comparison is between normalized vectors. The authors show that this loss reduces bias towards new classes.

\subsection{Rehearsal approaches}
\label{subsec:rehearsal}
Rehearsal methods keep a small number of exemplars~\cite{castro2018end, rebuffi2017icarl, wu2019large} (exemplar rehearsal),  or generate synthetic images~\cite{shin2017continual, ostapenko2019learning} or features~\cite{kemker2017fearnet, xiang2019incremental} (pseudo-rehearsal). By replaying the stored or generated data from previous tasks rehearsal methods aim to prevent the forgetting of previous tasks. Most rehearsal methods combine the usage of exemplars to tackle the inter-task confusion with approaches that deal with other causes of catastrophic forgetting. The usage of exemplar rehearsal for \mbox{class-IL} was first proposed in \textit{Incremental Classifier and Representation Learning} (iCaRL)~\cite{rebuffi2017icarl}. This technique has since been applied in the majority of \mbox{class-IL} methods. In this section, we focus on the choices which need to be taken when applying exemplars.

\minisection{Memory types.}
Exemplar memory must be extended at the end of a training session after the model has already been adapted to the new task. If the memory has a fixed maximum size across all tasks (fixed memory), some exemplars must first be removed to make space for new ones. This ensures that the memory capacity stays the same and the capacity is fixed. The more tasks and classes that are learned, the less representation each class has for rehearsal. After learning a certain amount of tasks, the memory could be expanded to better accommodate the new distributions. However, previously removed samples will be lost, thus the decision of when to expand is an important one.
If the memory is allowed to grow (growing memory), then only new samples from the current task need to be added. This enforces the classes to have a stable representation during rehearsal across all tasks, at the cost of having a linear increase of memory, which might not be suitable in some applications. In both cases, the number of exemplars per class is enforced to be the same to ensure an equal representation of all classes.

\minisection{Sampling strategies.}
The simplest way to select exemplars to add to the memory is by randomly sampling from the available data (random), which has been shown to be very effective without much computational cost~\cite{chaudhry2018riemannian, rebuffi2017icarl}.

Inspired by Welling~\cite{welling2009herding}, iCaRL selects exemplars based on their corresponding feature space representations (herding). Representations are extracted for all samples and the mean for each class is calculated. The method iteratively selects exemplars for each of the classes. At each step, an exemplar is selected so that, when added to the exemplars of its class, the resulting exemplar mean is closest to the real class mean. The order in which exemplars are added is important, and taken into account when some need to be removed. Although this iterative selection procedure can outperform random, it increases computational cost.

In RWalk~\cite{chaudhry2018riemannian}, two other sampling strategies are proposed. The first one calculates the entropy of the softmax outputs and selects exemplars with higher entropy (entropy). This enforces selection of samples that have more distributed scores across all classes. Similarly, the second one selects exemplars based on how close they are to the decision boundary (distance), assuming that the feature space and the decision boundaries do not change too much. For a given sample $(\mathbf{x}_{i},\mathbf{y}_{i})$, the pseudo-distance to the decision boundary is calculated by $f(\mathbf{x}_{i};\phi)^T V_{\mathbf{y}_{i}}$, meaning that the smaller the distance, the closer to the decision boundary.

For these sampling strategies (except for random), the order which exemplars are chosen is recorded following a decreasing order of importance. If a fixed memory is used and some memory must be freed to make space for new exemplars, the exemplars with the lower importance are the ones removed first.

\minisection{Task balancing.}
When applying rehearsal during the training of a new task, the weight of the new classes compared to the previous ones is defined by the trade-off between the two parts of the loss, as well as the number of samples from each class at each training step. Most approaches sample the training batches from the joint pool between new data and rehearsal exemplars~\cite{chaudhry2018riemannian, hou2019learning, rebuffi2017icarl, wu2019large}. This means that batches are clearly over-represented by new samples and rely on the trade-off between the cross-entropy loss and the other losses that prevent forgetting. In contrast, \emph{End-to-End Incremental Learning} (EEIL)~\cite{castro2018end} proposes having a more balanced training where batches are equally distributed between new and previous classes. This seems to have quite beneficial effects in compensating for the task imbalance during training. GDumb~\cite{prabhu2020gdumb} is a baseline that ensures a balanced training set. Their method greedily stores samples in memory in a class-balanced way as they arrive. For testing, they train a model from scratch using only samples in the memory. This extremely simple baseline was shown to obtain competitive results on several IL scenarios, but for class-IL it was outperformed by methods that exploit information from samples not present in the memory.

\minisection{Combining rehearsal and data regularization.} Several methods~\cite{rebuffi2017icarl,castro2018end,wu2019large,hou2019learning} use the distillation loss from Learning without Forgetting~\cite{li2017learning} to deal with the activation drift in combination with exemplars. However, Beloudah and Popescu~\cite{belouadah2019il2m} do the important observation that this distillation term actually hurts performance when using exemplars.
The results in Table.~\ref{tab:regul-methods} confirm this, however in some scenarios a combination of weight regularization and exemplar rehearsal can be beneficial (see Figs.~\ref{fig:cifar100_fixd_plot_cifar100_fixd_lft_plot},~\ref{fig:vggface2},~\ref{fig:imagenet_plot}). Additionally, when combined with auxiliary unlabelled data using double distillation~\cite{lee2019overcoming}, results improve further.

Recent research on task-IL~\cite{titsias2020functional} shows that data regularization (referred to as functional regularization) can provide a natural way to select data for rehearsal by choosing the inducing points of the Gaussian process used to approximate the posterior belief over the underlying task-specific function (network output). This direction was further explored in~\cite{pan2020continual}, however the usefulness of these approaches for class-IL is still to be determined.

\subsection{Bias-correction approaches}
\label{sec:bias_correction}
Bias-correction methods aim to address the problem of task-recency bias, which refers to the tendency of incrementally learned network to be biased towards classes in the most recently learned task. This is mainly caused by the fact that, at the end of training, the network has seen many examples of the classes in the last task but none (or very few in case of rehearsal) from earlier tasks. One direct consequence of this, as observed by Hou et al.~\cite{hou2019learning}, is that the classifier norm is larger for new classes than for the previous ones and that the classifier bias favors the more recent classes. This effect is shown in Fig.~\ref{fig:bias_icarl}, where the lower biases and reduced norm of the classifier make less likely for the network to select any of the previous classes. In this section, we discuss several approaches to address this problem.

The earlier mentioned iCaRL method~\cite{rebuffi2017icarl} combines exemplars and Learning without Forgetting, using a classifier layer and cross-entropy loss during training. To prevent the task-recency bias, they do not use the classifier at inference. Instead they compute the class mean of the exemplars in the feature representation, and then apply a nearest exemplar-mean for classification. Since this process is independent of the weights and biases of the final layer, the method was shown to be much less prone to the task-recency bias.

One simple yet effective approach to prevent task-recency bias has been proposed by Castro et al.~\cite{castro2018end} in their method EEIL. They suggest introducing an additional stage, called \emph{balanced training}, at the end of each training session. In this phase, an equal number of exemplars from all classes is used for a limited number of iterations. To avoid forgetting the new classes, they introduce a distillation loss on the classification layer only for the classes from the current task. Balanced training could come at the cost of overfitting  to the exemplars that have been stored, when these do not completely represent the distribution.

\figbiasWeights{0.49}{Bias and weight analysis for iCaRL with 2,000 exemplars on \mbox{CIFAR-100}. We show the ordered biases and norm of the last classification layer of the network for different tasks. Note how the bias and the norm of the weights are higher for the last tasks. This results in a \emph{task-recency bias}.}

Another simple and effective approach to preventing task-recency bias was proposed by Wu et al.~\cite{wu2019large}, who call their method  \emph{Bias Correction} (BiC). They add an additional layer dedicated to correcting task bias to the network. A training session is divided into two stages. During the first stage they train the new task with the cross-entropy loss and the distillation loss (see Eq.~\ref{eq:LwF}). Then they use a split of a very small part of the training data to serve as a validation set during the second phase. They propose to learn a linear transformation on top of the logits, $\mathbf{o}_k$, to compensate for the task-recency bias. The transformed logits are given by:
\begin{equation}
    \mathbf{q}_k = \alpha_s \mathbf{o}_k + \beta_s,\;\; c_k \in C^s\label{eq:bic}
\end{equation}
where $\alpha_s$ and $\beta_s$ are the parameters which compensate for the bias in task $s$. For each task there are only two parameters which are shared for all classes in that task (initialized to $\alpha_1=1$ and $\beta_1=0)$. In the second phase, all the parameters in the network are frozen, except for the parameters of the current task $\alpha_t$ and $\beta_t$. These are optimized with a standard softmax on the transformed logits $\mathbf{q}_k$ using the set-aside validation set. Finally, they only apply a weight decay on $\beta$ parameters and not on the $\alpha$ parameters.

As mentioned earlier, task-recency bias was also observed by Hou et al.~\cite{hou2019learning}. In  their method \emph{Learning a Unified Classifier Incrementally via Rebalancing} (LUCIR), they propose to replace the standard softmax layer $\sigma$ with a cosine normalization layer according to:
\begin{equation}
    \mathcal{L}_{cos}(\mathbf{x};\theta^{t}) = \sum_{k=1}^{N^{t}} y_k \log \frac{\exp(\eta   \langle \frac{f(\mathbf{x})}{||f(\mathbf{x})||}, \frac{V_k}{||V_k||}
     \rangle )}{\sum_{i=1}^{N^t} \exp(\eta   \langle \frac{f(\mathbf{x})}{||f(\mathbf{x})||}, \frac{V_i}{||V_i||} \rangle)}
\label{eq:cos}
\end{equation}
where $f(\mathbf{x})$ are the feature extractor outputs, $\langle \cdot, \cdot \rangle$ is the inner product, $V_k$ are the classifier weights (also called class embedding) related to class $k$, and $\eta$ is a learnable parameter which controls the peakiness of the probability distribution.

Hou et al.~\cite{hou2019learning} also address the problem of inter-task confusion. To prevent new classes from occupying a similar location as classes from previous tasks, they apply the \emph{margin ranking loss}. This loss pushes the current embedding away from the embeddings of the $K$ most similar classes according to:
\begin{eqnarray}
 \mathcal{L}_{mr}(\mathbf{x}) =\!\sum_{k=1}^{K} \max \left(m-{\scriptstyle \langle \frac{f(\mathbf{x})}{||f(\mathbf{x})||}, \frac{V_y}{||V_y||} \rangle} +  {\scriptstyle \langle \frac{f(\mathbf{x})}{||f(\mathbf{x})||}, \frac{V_k}{||V_k||} \rangle} , 0 \right)
\label{eq:marginloss}
\end{eqnarray}
where $\hat V_y$ refers to the ground truth class embedding of $\mathbf{x}$, $\hat V_k$ is the embedding of the closest classes, and $m$ is the margin.

Finally, another approach that addresses task-recency bias was proposed by Belouadah and Popescu~\cite{belouadah2019il2m} with their method called \emph{\mbox{Class-IL} with dual memory} (IL2M). Their method is similar to BiC~\cite{wu2019large} in the sense that they propose to rectify the network predictions. However, where BiC learns to rectify the predictions by adding an additional layer, IL2M rectifies based on the saved certainty statistics of predictions of classes from previous tasks. Defining $m = \arg\max \hat{\mathbf{y}}(\mathbf{x})$, they compute the rectified predictions of the previous classes $k$ as:
\begin{equation}
  \hat y_k^r (\mathbf{x}) =
  \begin{cases}
    \hat y_k (\mathbf{x})\times \frac{\overline{y}_k^p}{\overline{y}_k^t} \times \frac{\overline{y}^t}{\overline{y}^p}  & \mbox{if } m \in C^t \\
    \hat y_k(\mathbf{x}) & \mbox{otherwise}.
  \end{cases}
  \label{eq:il2m}
\end{equation}
Here $\overline{y}_k^p$ (superscript $p$ refers to past) is the mean of the predictions $\hat y_k$ for all images of class $c_k$ after training the task in which class $c_k$ is first learned ($c_k\in C^p$), and $\overline{y}^p$ is the mean of the predictions for all classes in that task. Both $\overline{y}_k^p$ and $\overline{y}^p$ are stored directly after their corresponding training session. $\overline{y}_k^t$ is the mean of the predictions $\hat y_k$ for all images of class $c_k$ after training the new task (this is computed based on the exemplars). Similarly, $\overline{y}^t$ is the mean of the predictions for all classes in the new task. As can be seen the rectification is only applied when the predicted class is a new class ($m\in C^t$). If the predicted class is an old class, the authors argue that no rectification is required since the prediction does not suffer from task-imbalance.

\figExpSingleColumn{1.0}{20 0 20 0}{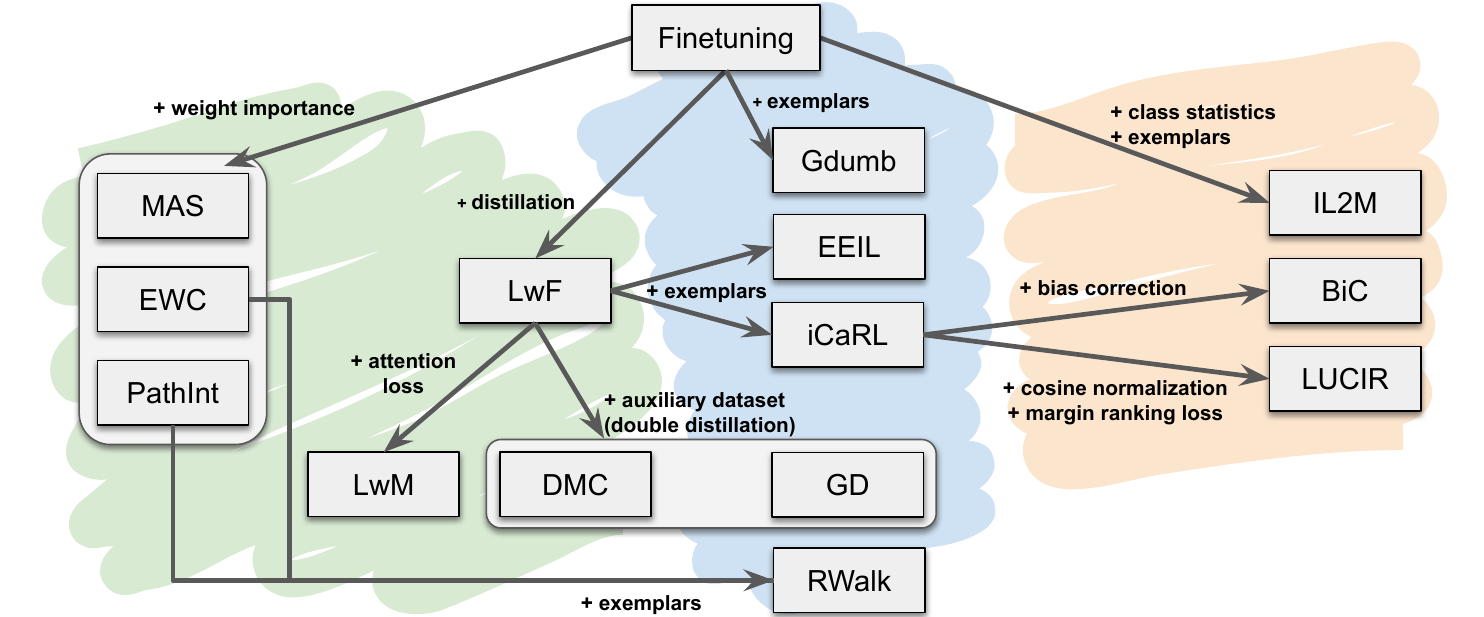}{Relation between \mbox{class-IL} methods. We distinguish three main categories: exemplar-free regularization (green), rehearsal (blue), and rehearsal with explicit bias-correction mechanisms (orange). Methods that share relations are joined in a rounded box.}

\subsection{Relation between class-incremental methods}
In previous sections we discussed the main approaches to mitigating catastrophic forgetting by incremental learning methods. We summarize their relations in Fig.~\ref{fig:approaches_graph} starting from the naive finetuning approach. In the diagram we show all methods which we compare in Sec.~\ref{sec:results}. It distinguishes between methods using exemplars to retain knowledge (blue, orange) and exemplar-free methods (green). 

Most notably, the huge impact of Learning without Forgetting (LwF)~\cite{li2017learning} upon the whole field of class-IL is clear. However, we expect that with the recent findings~\cite{belouadah2019il2m}, which show that when combined with exemplars finetuning can outperform LwF, could somewhat lessen its continuing influence. Weight regularization methods~\cite{aljundi2018memory,kirkpatrick2017overcoming,zenke2017continual}, applied frequently in the task-IL setting, are significantly less used for \mbox{class-IL}. They can also be trivially extended with exemplars and we include results of this in Sec.~\ref{sec:results}. Finally, Fig.~\ref{fig:approaches_graph} also shows the influence of iCaRL~\cite{rebuffi2017icarl} in the development of more recent methods~\cite{wu2019large,hou2019learning}.

%%%%%%%%%%%%%%%%%%%%%%%%%%%%%%%%%%%%%
\section{Related work}
\label{sec:relatedwork}
In this section we broadly review related work, focusing mainly on works not discussed in the previous section.

\minisection{Existing surveys.} The problem of catastrophic forgetting has been acknowledged for many years. Already in the eighties, McCloskey and Cohen~\cite{mccloskey1989catastrophic} showed that algorithms trained with backpropagation suffered from catastrophic forgetting. Radcliff~\cite{ratcliff1990connectionist} confirmed this finding on a wider range of tasks trained with backpropagation. An excellent review on early approaches to mitigating catastrophic forgetting is by French~\cite{french1999catastrophic}. This review also discusses how the brain prevents catastrophic forgetting and lays out possible solutions for neural network design. With the resurgence of deep learning from around 2011~\cite{krizhevsky2012imagenet} the problem of catastrophic forgetting quickly gained renewed attention~\cite{goodfellow2013empirical,kirkpatrick2017overcoming}. This led to a surge of work in incremental learning, continual learning and lifelong learning.

This surge of new research has also resulted in recent surveys on the subject. Parisi et al.~\cite{parisi2019continual} provide an extensive survey on lifelong learning. This review is especially valuable because of its in-depth discussion of how biological systems address lifelong learning. They thoroughly discuss biologically-motivated solutions, such as structural plasticity, memory replay, curriculum and transfer learning. Another review~\cite{lesort2020continual} focuses on continual learning for robotics, and puts special effort into unifying evaluation methodologies between continual learning for robotics and non-robotics applications, with the aim of increasing cross-domain progress in continual learning. These reviews, however, do not provide an experimental performance evaluation of existing methods in the field.

Some recent surveys do include evaluation of methods. Pfulb and Gepperth~\cite{pfulb2019comprehensive} propose a training and evaluation paradigm for task-IL methods, limited to two tasks. De Lange et al.~\cite{delange2021continual} perform an extensive survey of task-IL with an experimental evaluation, including an analysis of model capacity, weight decay, and dropout regularization within context of task-IL. In addition, they propose a framework for continually determining the stability-plasticity trade-off of the learner -- which we also apply in our evaluation. The more challenging setting of class-IL has led to many new methods to address its particular problems. The majority of these methods were not included by De Lange et al.~\cite{delange2021continual}, nor do they perform any evaluation in the class-IL setting.

There are two concurrent surveys on class-IL: Mai et al.~\cite{mai2021online} perform a comparison of methods in the online class-IL settings, whereas offline class-IL is the focus of our survey. Belouadah et al.~\cite{belouadah2020comprehensive} perform a comparison of several class-IL methods. Our survey and performance evaluation include methods EWC, MAS, PathInt, EEIL, RWalk, LwM, DMC and GD, all of which are excluded from their survey.

\minisection{Mask-based methods.} This type of parameter isolation methods reduce or completely eliminate catastrophic forgetting by applying masks to each parameter or to each layer's representations. However, by learning useful paths for each task in the network structure, the simultaneous evaluation of all learned tasks is not possible. This forces several forward passes with different masks, which makes such methods effective for task-aware evaluation, but impractical for task-agnostic settings~\cite{delange2021continual, masana2020ternary}. Piggyback learns masks on network weights while training a backbone~\cite{mallya2018piggyback}. PackNet learns weights and then prunes them to generate masks~\cite{mallya2018packnet}. HAT~\cite{serra2018overcoming} applies attention masks on layer activations to penalize modifications to those that are important for a specific task. DAN~\cite{rosenfeld2018incremental} combines existing filters to learn filters for new tasks. Finally, PathNet~\cite{fernando2017pathnet} learns selective routing through the weights using evolutionary strategies.

\minisection{Dynamic architectures.} The other type of parameter isolation method, called architecture growing, dynamically increases the capacity of the network to reduce catastrophic forgetting. These methods rely on promoting a more intransigent model capable of maintaining previous task knowledge, while extending that model in order to learn new tasks. This makes some of these methods impractical when the task-ID is not known, or adds too many parameters to the network which makes them unfeasible for large numbers of tasks. EG~\cite{aljundi2016expert} duplicates the model for each new task in order to completely eliminate forgetting. PNN~\cite{rusu2016progressive} extends each layer and adds lateral connections between duplicates for each task. Old weights are fixed, allowing access to that information while learning the new task. However, complexity increases with the number of tasks. To address this issue, P\&C~\cite{schwarz2018progress} proposes duplicating the network only once to keep the number of parameters fixed, and use EWC~\cite{kirkpatrick2017overcoming} to mitigate forgetting. Xiao et al.\cite{xiao2014error} group similar classes together expanding hierarchically, at the cost of an expensive training procedure and a rigid architecture. ACL~\cite{ebrahimi2020adversarial} fuses a dynamic architecture with exemplars, explicitly disentangling shared and task-specific features with an adversarial loss. This allows to learn shared features that are more robust to forgetting.

Finally, Random Path Selection (RPS)~\cite{rajasegaran2019random} provides better performance with a customized architecture by combining distillation and rehearsal-based replay. Contrary to some of the previously mentioned approaches, RPS does not need a task-ID at inference time. However, in order to learn the different paths for each task, the proposed architecture is much larger than other \mbox{class-IL} approaches. Since this approach needs to use their particular RPSNet architecture and the capacity is not comparable to the other approaches compared in this survey, we provide results in Appendix~\ref{app:suppl-results} (see Sec.~\ref{app:rps_networks} and Table~\ref{tab:rpsnets}), for an analysis on different numbers of paths and memory required.

\minisection{Online incremental learning.}
Within the field of incremental learning, online methods are based on streaming frameworks where learners are allowed to observe each example only once instead of iterating over a set of examples in a training session. Lopez-Paz~\cite{lopez2017gradient} establishes definitions and evaluation methods for this setting and describes GEM, which uses a per-task exemplar memory to constrain gradients so that the approximated loss from previous tasks is not increased. A-GEM~\cite{chaudhry2018efficient} improves on GEM in efficiency by constraining based on the average of gradients from previous class exemplars. However, Chaudhry et al.~\cite{chaudhry2019continual} show that simply training on the memorized exemplars, similar to the well-established technique in reinforcement learning~\cite{mnih2013playing, rolnick2019experience}, outperforms previous results. GSS~\cite{aljundi2019gradient} performs gradient-based exemplar selection based on the GEM and A-GEM procedure to allow training without knowing the task boundaries. MIR~\cite{aljundi2019online} trains on the exemplar memory by selecting exemplars that will have a larger loss increase after each training step. Riemer et al.~\cite{riemer2019scalable} use the memory  to store discrete latent embeddings from a Variational Autoencoder that allows generation of previous task data for training. MER~\cite{riemer2019learning} combines experience replay with a modification of the meta-learning method Reptile~\cite{nichol2018first} to select replay samples which minimize forgetting. Online methods can be extended to the offline setting by running them for multiple epochs on each training session. We show some results in Sec.~\ref{sec:exemplar-exps}.

\minisection{Variational continual learning.}
Variational continual learning is based on the Bayesian inference framework. VCL~\cite{nguyen2018variational} proposes to merge online and Monte Carlo variational inference for neural networks yielding variational continual learning. It is general and applicable to both discriminative and generative deep models. VGL~\cite{farquhar2019unifying} introduces Variational Generative Replay, a variational inference generalization of Deep Generative Replay (DGR), which is complementary to VCL. UCL~\cite{ahn2019uncertainty} proposes uncertainty-regularized continual learning based on a standard Bayesian online learning framework. It gives a fresh interpretation of the Kullback-Leibler (KL) divergence term of the variational lower bound for the Gaussian mean-field approximation case. FBCL~\cite{chen2018facilitating} proposes to use Natural Gradients and Stein Gradients to better estimate posterior distributions over the parameters and to construct coresets using approximated posteriors. IUVCL~\cite{swaroop2018improving} proposes a new best-practice approach to mean-field variational Bayesian neural networks. CLAW~\cite{adel2020continual} extends VCL by applying an attention mechanism on the whole network which allows automation of the architecture adaptation process that assigns parameters to be fixed or not after each task. UCB~\cite{ebrahimi2020uncertainty} defines uncertainty for each weight to control the change in the parameters of a Bayesian Neural Network, identifying which are the weights that should stay fixed or change. They further extend their method by using a pruning strategy together with binary masks for each task to retain performance from previous tasks. These methods normally consider only evaluation on task-IL. BGD~\cite{zeno2018task} updates the posterior in closed form and that does not require a task-ID.

\minisection{Pseudo-rehearsal methods.} In order to avoid storing exemplars and privacy issues inherent in \emph{exemplar rehearsal}, some methods learn to generate examples from previous tasks. DGR~\cite{shin2017continual} generates those synthetic samples using an unconditional GAN. An auxiliary classifier is needed to assign ground truth labels to each generated sample. An improved version is proposed in MeRGAN~\cite{wu2018memory}, where a label-conditional GAN and replay alignment are used. DGM~\cite{ostapenko2019learning} combines the advantages of conditional GANs and synaptic plasticity using neural masking. A dynamic network expansion mechanism is introduced to ensure sufficient model capacity. Lifelong GAN~\cite{zhai2019lifelong} extends image generation without catastrophic forgetting from label-conditional to image-conditional GANs. As an alternative to exemplar rehearsal, some methods perform \emph{feature replay}~\cite{kemker2017fearnet, xiang2019incremental}, which need a fixed backbone network to provide good representations.

\minisection{Incremental Learning beyond image classification.}
Shmelkov et al.~\cite{shmelkov2017incremental} propose to learn object detectors incrementally. They use Fast-RCNN~\cite{girshick2015fast} as the network and propose distillation losses on both bounding box regression and classification outputs. Additionally, they choose to distill the region proposal with the lowest background scores, which filters out most background proposals. Hao et al.~\cite{hao2019end} extend Faster-RCNN~\cite{ren2015faster} with knowledge distillation. Similarly, Michieli et al.~\cite{michieli2019incremental} propose to distill both on the output logits and on intermediate features for incremental semantic segmentation. Recently, Cermelli et al.~\cite{cermelli2020modeling} model the background by revisiting distillation-based methods and the conventional cross entropy loss. Specifically, previous classes are seen as background for the current task and current classes are seen as background for distillation. Incremental semantic segmentation has also been applied to remote sensing~\cite{tasar2019incremental} and medical data~\cite{ozdemir2018learn}.

Catastrophic forgetting has been mainly studied in feed-forward neural networks. Only recently the impact of catastrophic forgetting in recurrent LSTM networks was studied~\cite{schak2019study}. In this work, they observe that catastrophic forgetting is even more notable in recurrent networks than feed-forward networks. This is because recurrent networks amplify small changes of the weights. To address catastrophic forgetting an expansion layer technique for RNNs was proposed~\cite{Coop2013fixedexpansion}. A Net2Net technique~\cite{chen2015net2net} was combined with gradient episodic memory~\cite{Sodhani2019rnnlll}. In addition, they propose a benchmark of tasks for training and evaluating models for learning sequential problems. Finally, Del Chiaro et al.~\cite{delchiaro2020RATT} study preventing forgetting for the task of captioning.

The paper that introduced EWC~\cite{kirkpatrick2017overcoming} also considered training Deep Reinforcement Learning (DRL) agents to play multiple Atari games~\cite{mnih2013playing} over their lifetimes. Reinforcement learning (RL) is an application area of deep learning in which \emph{task specification} is usually implicit in the definition of the reward function to be optimized, and as such is another example where laboratory practice often does not completely reflect the real world since the agent's goals must evolve with the changing environment around them. Incremental task acquisition enjoys a long tradition in the RL community~\cite{moore1993prioritized}, and more recently the CLEAR approach mixes on-policy learning for plasticity with off-policy learning from replayed experiences to encourage stability with respect to tasks acquired in the past~\cite{rolnick2019experience}.

%%%%%%%%%%%%%%%%%%%%%%%%%%%%%%%%%%%%%
\section{Experimental setup}
\label{sec:exp-setup}
In this section, we explain the experimental setup and how we evaluate the approaches. We also introduce the baselines and the experimental scenarios used to gather the results presented in Sec.~\ref{sec:results}. More details on the implementation of the methods are described in Appendix~\ref{app:hyperparams}.

\subsection{Code framework}
In order to make a fair comparison between the different approaches, we implemented a versatile and extensible framework. Datasets are split into the same partitions and data is queued in the same order at the start of each task. All library calls related to randomness are synchronized and set to the same seed so that the initial conditions for all methods are the same. Data from previous tasks (excluding exemplar memory) is not available during training, thus requiring selection of any stability-plasticity-based trade-off before a training session of a task is completed (see also Sec.~\ref{sec:hyperparameters}).

The current version of the code includes implementations of several baselines and the following methods:
EWC~\cite{kirkpatrick2017overcoming},
MAS~\cite{aljundi2018memory},
PathInt~\cite{zenke2017continual}, RWalk~\cite{chaudhry2018riemannian},
LwM~\cite{dhar2019learning},
DMC~\cite{zhang2020class},
GD~\cite{lee2019overcoming},
GDumb~\cite{prabhu2020gdumb},
LwF~\cite{li2017learning},
iCaRL~\cite{rebuffi2017icarl}, EEIL~\cite{castro2018end},
BiC~\cite{wu2019large},
LUCIR~\cite{hou2019learning},
and IL2M~\cite{belouadah2019il2m}. The framework includes extending most exemplar-free methods with the functionality of exemplars, facilitates using a wide variety of network architectures, and allows running the various experimental scenarios we perform in this paper. As such, our framework contributes to the wider availability and comparability of existing methods, which will facilitate future research and comparisons of \mbox{class-IL} methods.

\subsection{Datasets}
We study the effects of CL methods for image classification on nine different datasets whose statistics are summarized in Appendix~\ref{app:hyperparams}. First, we compare the three main categories of approaches described in Sec.~\ref{sec:approaches} on the \mbox{CIFAR-100} dataset~\cite{krizhevsky2009learning}. Next, we use several fine-grained classification datasets: Oxford Flowers~\cite{nilsback2008automated}, MIT Indoor Scenes~\cite{quattoni2009recognizing}, CUB-200-2011 Birds~\cite{wah2011caltech}, Stanford Cars~\cite{krause20133d}, FGVC Aircraft~\cite{maji2013fine}, and Stanford Actions~\cite{yao2011human}. These provide higher resolution and allow studying the effects on larger domain shifts when used as different tasks. To study the effects on smaller domain shifts, we use the VGGFace2 dataset~\cite{cao2018vggface2}. Since the original dataset has no standard splits for our setting, we keep the 1,000 classes that have the most samples and split the data following the setup from~\cite{belouadah2019il2m}. This means that this dataset is not totally balanced, but at least all used classes have a large enough pool of samples. Finally, the ImageNet dataset~\cite{russakovsky2015imagenet} is used as a more realistic and large-scale scenario. It consists of 1,000 diverse object classes with different numbers of samples per class. Since this dataset takes time and needs a lot of resources, we also use the reduced ImageNet-Subset, which contains the first 100 classes from ImageNet as in~\cite{rebuffi2017icarl}.

In order to apply a patience learning rate schedule and a hyperparameter selection framework, an additional class-balanced split of 10\% from training is assigned to validation for those datasets. In the case of Flowers and Aircraft, we fuse the official train and validation splits and then randomly extract a class-balanced 10\% validation split.

\subsection{Metrics}
In incremental learning, $a_{t,k} \in [0, 1]$ denotes the accuracy of task $k$ after learning task $t$ ($k \leq t$), which provides precise information about the incremental process. In order to compare the overall learning process, the \emph{average accuracy} is defined as $A_t = \frac{1}{t} \sum_{i=1}^{t} a_{t,i}$ at task $t$. This measure is used to compare performances of different methods with a single value. When tasks have different numbers of classes, a class frequency weighted version is used.

Additional metrics focusing on several aspects of IL such as \emph{forgetting} and \emph{intransigence}~\cite{chaudhry2018riemannian} have also been proposed. \emph{Forgetting} estimates how much the model forgot about previous tasks. Another measure, \emph{intransigence} quantifies a model's inability to learn a new task. Both can be considered complementary measures that help understand the stability-plasticity dilemma. These measures were originally proposed for task-IL. However, their use in class-IL was not found equally useful. When adding new tasks the performance of previous tasks drops because the learner has to perform the more complex task of classifying data in all seen classes. This effect will incorrectly contribute to the forgetting measure. Therefore, in this survey we use \emph{average accuracy} as the main metric.

All reported CIFAR-100 results are averages over 10 runs, while the domain shift and different architecture results are averages over 5 runs. Each run uses a different random seed, but these are fixed across the different approaches so that the comparison is on identical splits generated from the same set of seeds.

\subsection{Baselines}
\label{sec:baselines}
Training with only a cross-entropy loss (see Eq.~\ref{eq:CE}) is the default Finetuning (FT) baseline common in most IL works. This learns each task incrementally while not using any data or knowledge from previous tasks and is often used to illustrate the severity of catastrophic forgetting. However, when moving to a \mbox{class-IL} scenario where all previous and new classes are evaluated, other finetuning variants can be considered. We might not update the weights corresponding to the outputs of previous classes (FT+), which avoids the slow forgetting due to not having samples for those classes (see Eq.~\ref{eq:CE2}). As seen in Table~\ref{tab:baselines}, this simple modification has an impact on the baseline performance. Since previous classes will not be seen, freezing the weights associated with them avoids biased modifications based only on new data. Furthermore, in the proposed scenarios approaches usually make use of an exemplar memory, which helps improve overall performance and avoid catastrophic forgetting by replaying previously seen classes. Therefore, as an additional baseline we also consider extending FT with the same exemplar memory as exemplar-based approaches (FT-E). The result of this is quite clearly more beneficial than the other FT baselines, and makes the baseline more comparable with approaches using the same memory.

In the case of Freezing (FZ), the baseline is also simple: we freeze all layers except the last one (corresponding to the classification layer or head of the network) after the first task is learned. Similarly to FT, we can also make the simple modification of not updating the weights directly responsible for the previous classes outputs (FZ+). This extends freezing to that specific group of weights which we know will not receive a gradient from previous class samples. As seen in Table~\ref{tab:baselines}, this leads to a more robust baseline. However, if we add exemplars (FZ-E) the performance decreases with respect to FT-E. We have also observed that, when starting from a larger first task, freezing can achieve much better performance since the learned representations before freezing are more robust.

Finally, we also use as an upper bound the joint training over all seen data (Joint). In order to have this baseline comparable over all learned tasks, we perform incremental joint training which uses all seen data at each task, starting from the model learned for the previous one. This baseline gives us an upper bound reference for all learned tasks.

\subsection{Hyperparameter selection}
\label{sec:hyperparameters}
For a fair comparison of IL methods, two main issues with non-IL evaluation need to be addressed. The first is that choosing the best hyperparameters for the sequence of tasks after those are learned is not a realistic scenario in that information from future tasks is used. A better comparison under an IL setting is to search for the best hyperparameters as the tasks are learned with the information at hand for each of them. Second, it makes the comparison very specific to the scenario, and in particular to the end of the specific sequence of tasks. It provides a less robust evaluation of the results over the rest of tasks, which means that other task sequence lengths are not taken into account. We feel that a broader evaluation of CL methods should include results over all tasks as if each of them were the last one for hyperparameter selection purposes.

In order to provide this more robust evaluation, we use the Continual Hyperparameter Framework proposed by De Lange et al.~\cite{delange2021continual}. This framework assumes that at each task, only the data for that task is available, as in a real scenario. For each task, a \textit{Maximal Plasticity Search} phase is used with Finetuning, and a \textit{Stability Decay} phase is used with the corresponding method. This allows to establish a reference performance first and then find the best stability-plasticity trade-off~\cite{delange2021continual} (see also \mbox{Appendix~\ref{app:hyperparams}}). The hyperparameters that have no direct correspondence with the intransigence-forgetting duality are set to the recommended values for each of the methods. A list of those, together with the values can be found in \mbox{Appendix~\ref{app:hyperparams}}.

\tabbaselines

\subsection{Network architectures}
As suggested by He et al.~\cite{he2016deep}, ResNet-32 and ResNet-18 are commonly used in the literature for \mbox{CIFAR-100} and datasets with larger resolution (input sizes of around $224 \times 224 \times 3$), respectively. Therefore, we use those architectures trained from scratch for most of the experiments, but we also include an analysis on different architectures in Sec.~\ref{sec:exp-networks} and \mbox{Appendix~\ref{app:smalldomain}}.

\tabregularizationmethods

\subsection{Experimental scenarios}
To make the following section easier to read, we define a few experimental scenarios here. We denote a dataset with $A$ tasks of $B$ classes each as $(A/B)$. We indicate scenarios having a different number of classes in the first task ($C$) with $(A/C$-$B)$. For example, a \mbox{(10/10)} experiment refers to splitting the dataset into 10 tasks of 10 classes each. Another setting is \mbox{(11/50-5)}, which means that the first task has 50 classes, and the remaining 10 tasks have 5 classes each. These are the two main settings proposed for evaluating the different approaches and their characteristics. In our evaluation, we do not consider the case where a task consists of only a single class. This is because several methods cannot be straightforwardly applied to this scenario, mainly because they train a cross-entropy loss on only the last task (e.g. BiC, DMC). Adding tasks with multiple classes is the most common scenario considered in \mbox{class-IL} literature.

%%%%%%%%%%%%%%%%%%%%%%%%%%%%%%%%%%%%%
\section{Experimental results}
\label{sec:results}
In this section, we evaluate a large number of incremental learning methods in terms of many aspects of incremental learning on a broad variety of datasets.

\subsection{On regularization methods}
Most of the regularization approaches have been proposed for a task-IL setting where the task-ID is known at inference time~\cite{jung2016less, kirkpatrick2017overcoming, lee2017overcoming, li2017learning, rannen2017encoder}. Since regularization is applied to weights or representations, they can be easily extended to a \mbox{class-IL} setting without much or any modification. This makes for a more challenging problem, and several more recent regularization methods already show results for \mbox{class-IL}~\cite{chaudhry2018riemannian, dhar2019learning, zhang2020class}. Similarly to the baselines in Sec.~\ref{sec:baselines}, when not using exemplars, methods can freeze the weights of the final layer associated with previous classes to improve performance based on the assumption that only data from new classes is used during a training session. This helps the problem of vanishing weights from learned classes and the task-recency bias, especially when using weight decay.

In Table~\ref{tab:regul-methods} we compare regularization-based methods for both task-IL and \mbox{class-IL}. Two methods that apply data regularization (LwF, LwM) and four weight regularization methods (EWC, PathInt, MAS, RWalk) are compared on \mbox{CIFAR-100} (10/10). The ten tasks are learned sequentially, and each method and setting shows average accuracy at the second, fifth and final tasks to illustrate different sequence lengths. We start by comparing the regularization methods without using exemplars. Results clearly show a significant drop in performance due to the lack of task-ID, especially after 5 and 10 tasks. LwF obtains better results than weight-based regularization methods, which might explain why distillation has been the dominant approach for most rehearsal methods~\cite{castro2018end, hou2019learning, rebuffi2017icarl, wu2019large}.

We also expand the regularization methods with exemplars to see how it affects their performance. Note that these methods are originally proposed without exemplars, except for RWalk. In Table~\ref{tab:regul-methods}, we include results with two types of memories. The baseline method FT-E obtains the best results, even better than LwF-E, as also noticed by Belouadah et al.~\cite{belouadah2019il2m}. The results show that exemplars are not straightforwardly combined with regularization, and can often hurt performance. The only exception is LwM-E (originally not proposed with exemplars) that outperforms FT-E in some short sequence cases. However, it should be noted that in some of the next experiments we find that weight regularization and exemplars can actually achieve good results. Note that in the remainder of the survey we mainly include results for these methods when they improve over FT-E -- our main baseline.

\figBiasCorrection

\subsection{On bias-correction}
As seen in Fig.~\ref{fig:cm}, there exists a clear bias towards recent tasks. Here we evaluate the success of \mbox{class-IL} methods to address the task-recency bias. To allow for a better visualization, we use \mbox{CIFAR-100} (5/20) with ResNet-32 trained from scratch and a fixed memory of 2,000 exemplars. In the text, we also give in brackets the average accuracy after the last task for all methods we considered.

We show the task confusion matrix for different bias-correction approaches in Fig.~\ref{fig:cm} and Fig.~\ref{fig:bias-correction-exp}. The FT-E~(40.9) baseline, despite having improved performance due to the use of rehearsal strategies, still has a clear task-recency bias. iCaRL~(43.5) clearly benefits from using the NME classifier, removing most task-recency bias, although at the cost of having slightly worse performance than the other approaches. EEIL~(47.6) ignores the task-recency bias during training of new tasks, however at the end of each training session it performs balanced training based only on the exemplars. This method obtains good performance, as balanced training calibrates all outputs from previous classes and thus removes a large part of the task-recency bias. BiC~(45.7) does a very good job at avoiding the bias while maintaining a good performance. It is clear that the newer tasks have less inter-task classification errors. However, it seems like the small pool of samples used for learning the $\alpha$ and $\beta$ parameters (see Eq.~\ref{eq:bic}) leads to having the opposite effect, and BiC appears to over-compensate toward previous tasks. LUCIR~(47.3) shows a more gradual task-recency bias while maintaining good performance. This could be related to the change in experimental scenario. LUCIR was shown to work better when having a larger first task followed by some smaller ones. In the more challenging setup used here their bias-correction struggles to obtain good results. Finally, IL2M~(45.6) overcomes task-recency bias while improving on iCaRL, although the task confusion matrix seems to point towards more inter-task miss-classifications.

These results show that the two methods that have better performance (EEIL, LUCIR) suffer from task-recency bias, while approaches that have a better solution for it (iCaRL, BiC, IL2M) still have a margin for performance improvement. This leaves room for future work to create new approaches that can both have better overall performance while simultaneously addressing the bias-correction issue.

\subsection{On exemplar usage}
\label{sec:exemplar-exps}
Here, we study the effects of different characteristics related to exemplars. The number of exemplars to store is limited by the type and amount of memory available, and exemplars are selected at the end of each training session following a sampling strategy. Note, that in Section~\ref{sec:online}, we will discuss the results of online exemplar-based approaches (namely GDumb~\cite{prabhu2020gdumb}, ER~\cite{chaudhry2019continual} and MIR~\cite{aljundi2019online}).

\figExpSingleColumn{1.0}{0 11 0 20}{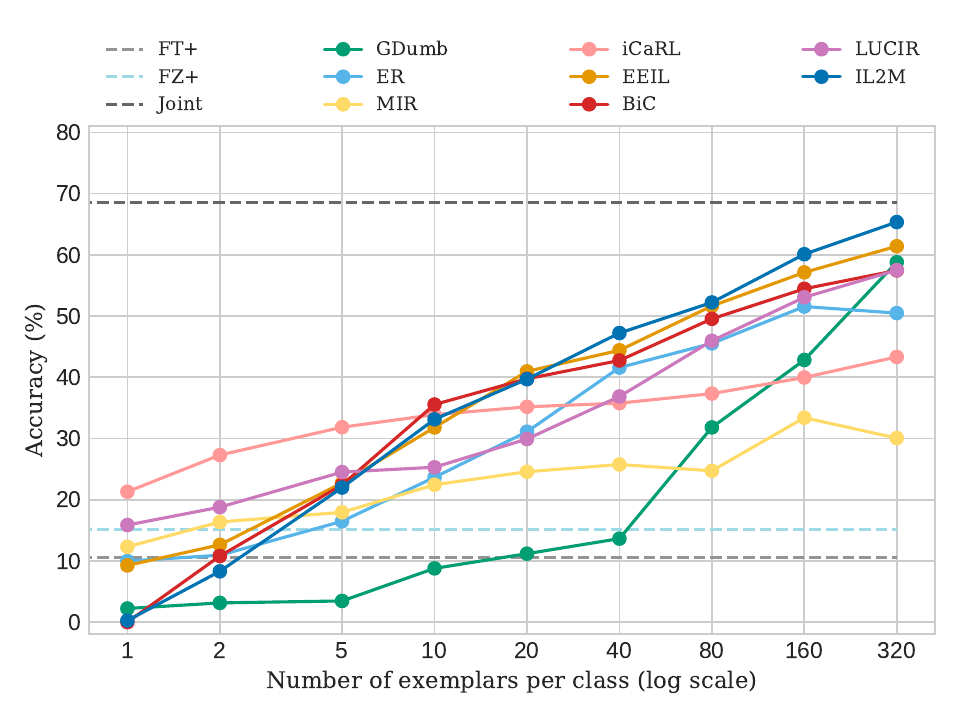}{Results for \mbox{CIFAR-100} (10/10) on \mbox{ResNet-32} trained from scratch with different exemplar memory sizes.}

\figExpDoubleColumnTwoImage{0.48}{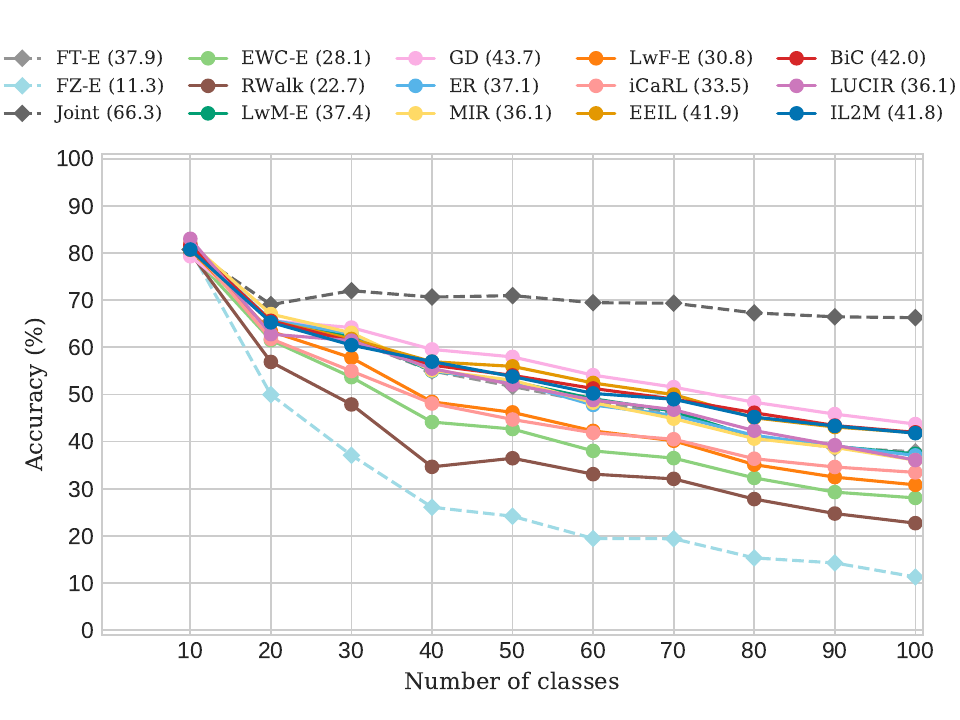}{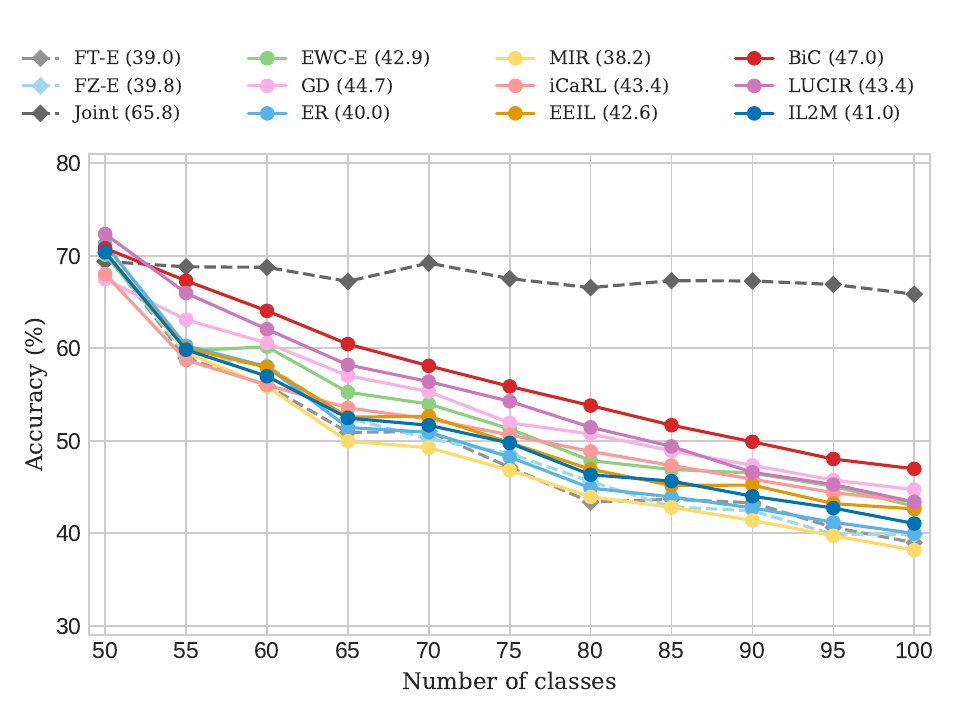}{\mbox{CIFAR-100} (10/10) with 2,000 exemplar fixed memory (left), and \mbox{CIFAR-100} (11/50-5) with 2,000 exemplar fixed memory (right). Results with 20 exemplars per class growing memory are available in Appendix~\ref{app:more-cifar}.}

\minisection{On memory size:}
We first analyze how the number of exemplars per class affects performance as we expand the exemplar memory. In Fig.~\ref{fig:number_exemplars} we compare several rehearsal methods with different numbers of exemplars per class in a growing memory. As expected, in almost all cases performance increases as more exemplars are added. LUCIR and iCaRL always perform equal to or better than FT+ and FZ+. When using few exemplars per class, the weights of the last layer can be modified by large gradients coming from new classes while very little to no variability of gradients comes from previous ones. We found that the freezing of the last layer weights as used in FZ+ provides a larger advantage than is obtained with only a few exemplars (see results with fewer than five exemplars for EEIL, BiC, and IL2M).

Adding more samples becomes more costly after 20 exemplars per class in comparison to the gain in performance obtained. As an example, expanding the memory from 10 to 20 samples per class on BiC yields a 6.2 point gain in average accuracy. Expanding from 20 to 40 yields a 4.8 point gain at the cost of doubling the memory size. For the other methods, these gains are similar or worse. Although starting with better performance with fewer exemplars per class, iCaRL has a slight slope, which makes the cost of expanding the memory less beneficial. LUCIR follows with a similar curve, and both seem to be further away from Joint training (upper bound), probably due to the differences in how the classification layer is defined (NME and cosine normalization, respectively). Finally, IL2M and EEIL are quite close to Joint training when using a third of the data as memory (160 out of 500 maximum samples per class). To maintain a realistic memory budget, and given the lower performance gains from increasing said memory, we fix growing memories to use 20 exemplars per class.

\minisection{On sampling strategies:}
As introduced in Sec.~\ref{subsec:rehearsal}, for rehearsal approaches, there are different strategies to select which exemplars to keep. In Table~\ref{tab:ex-sampling} we compare the FT-E baseline, the two most common regularization-based methods (LwF-E, EWC-E), and two of the latest bias-correction methods (EEIL, BiC). We use the four different sampling strategies introduced in Sec.~\ref{subsec:rehearsal}: random, herding (mean of features), entropy-based, and plane distance-based. These methods and strategies are evaluated under our two main proposed scenarios: \mbox{CIFAR-100} (10/10) and (11/50-5)---the second one available in \mbox{Appendix~\ref{app:sampling_strats}}.

Results generally show a slight preference across all approaches for the herding sampling strategy, with random sampling closely tied or slightly close. Both these strategies clearly outperform the others in both scenarios. When only evaluating after two tasks for the (10/10) scenario, the gap between them is even smaller, probably due to the large number of exemplars available at that point (2,000). The differences between random and herding are not statistically significant except for some methods on the longer 10-task sequence (see Appendix~\ref{app:sampling_strats}).

\tabexemplarsampling

\subsection{On different scenarios}
\label{sec:scenarios-exps}
In addition, we evaluate the method on two alternative scenarios that have been reported in the literature.

\minisection{On different starting scenarios:}
We explore two scenarios with different numbers of classes in the starting task. The first one compares methods on \mbox{CIFAR-100} (10/10), with classes equally split across all tasks. For the second scenario, we compare methods on \mbox{CIFAR-100} (11/50-5) which is similar to having the first task being a pretrained starting point with more classes and a richer feature representation before the subsequent 10 smaller tasks are learned. In Fig.~\ref{fig:cifar100_fixd_plot_cifar100_fixd_lft_plot} both scenarios are evaluated with fixed memory (2,000 total exemplars) with herding as the sampling strategy (results for the growing scenario are provided in Appendix~\ref{app:more-cifar}). Note that for RWalk we use the version with exemplars.

In Fig.~\ref{fig:cifar100_fixd_plot_cifar100_fixd_lft_plot} (left), the methods GD, BiC, EEIL and IL2M achieve the best results. Note that for GD we use the version without external data. Some methods have different starting points on task 1 since they do not have the same initial conditions as the other approaches (e.g. LUCIR uses cosine linear layers, while BiC uses fewer data during training because it stores some for bias-correction parameter training). It is quite clear that the approaches that tackle task-recency bias have an overall better performance than the others. Furthermore, as already noted~\cite{belouadah2019il2m}, FT-E achieves competitive performance similar to the lowest performance of that family.

Fig.~\ref{fig:cifar100_fixd_plot_cifar100_fixd_lft_plot} (right) shows that, in general, all methods improve when starting from a larger number of classes, probably because anchoring to the first task already yields more diverse features. This is especially noticeable in the case of FZ-E. The results show the importance of comparing to this baseline when doing experiments with pretrained models or a very strong first task. Both LUCIR and EWC-E also seem to perform much better in this scenario.

\minisection{On external data:}
Some methods use external data (e.g., ImageNet) to perform knowledge distillation from the previous model. DMC uses external data instead of exemplars and GD uses both external data and exemplars. We found that the gain obtained by distillation from an external dataset to be rather small. A detailed comparison of these approaches is given in Appendix~\ref{app:external_data}.

\subsection{On online approaches for offline scenarios}
\label{sec:online}
An active field of research is online class-IL (see also Sec.~\ref{sec:relatedwork}). Methods that are developed for the online scenario can be applied to the offline setting by allowing these algorithms to cycle several epochs over the data. Here, we do this for three methods, namely ER~\cite{chaudhry2019continual}, ER-MIR~\cite{aljundi2019online} and GDumb~\cite{prabhu2020gdumb}. The results for ER and ER-MIR are available in Figs.~\ref{fig:number_exemplars} and~\ref{fig:cifar100_fixd_plot_cifar100_fixd_lft_plot}. In Fig.~\ref{fig:number_exemplars} we can see that MIR outperforms ER for small memory sizes. However, when applied to large memories, the method obtains inferior results to any of the compared methods.
On CIFAR-100 (10/10) with a 2000 exemplar fixed memory, ER obtains an average accuracy of 37.1\% (see Fig.~\ref{fig:cifar100_fixd_plot_cifar100_fixd_lft_plot}). The ER results are very similar to the baseline method FT-E, and its performance is also similar. ER-MIR obtains an average accuracy of 36.1\% which is slightly below ER, showing that the reported gain for a single epoch is not maintained when data is revisited multiple times. GDumb has only been added in Fig.~\ref{fig:number_exemplars} since it obtained inferior results for the class-IL setting, as also reported by Prabhu et al.~\cite{prabhu2020gdumb}. In conclusion, the compared online methods obtain similar results as the strong baseline FT-E for offline class-IL.

\subsection{On the effect of domain shift}
\label{sec:domain-shift}
Up to this point, experiments have been on a dataset with a small input size and a wide variety of classes from a similar distribution. In this experiment, we study the effects of using tasks which have different degrees of domain shifts between them and whose images have higher resolution.

\figExpSingleColumn{1.0}{0 80 0 45}{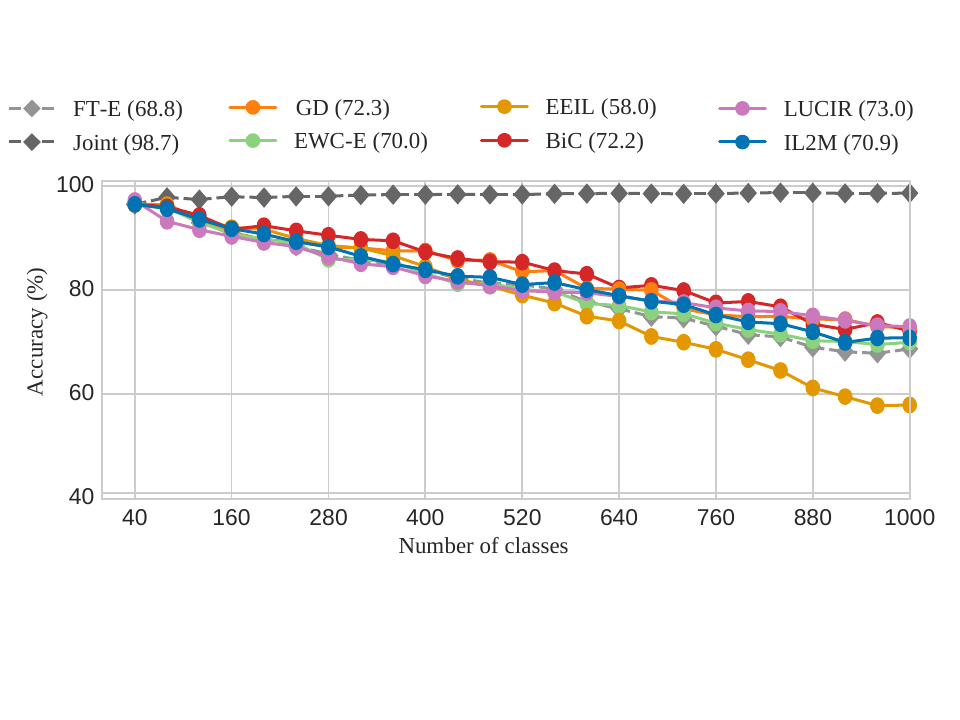}{Small domain shifts on VGGFace2 (25/40) with \mbox{ResNet-18} (scratch) and a 5,000 exemplar fixed memory.}

\minisection{Smaller domain shift:}
We first conduct experiments on very small domain shifts between different classes and tasks, as is the case for VGGFace2~\cite{cao2018vggface2}. We divide the 1,000 classes equally into 25 tasks of 40 classes, store 5,000 exemplars in a fixed memory, and train ResNet-18 from scratch. In Fig.~\ref{fig:vggface2}~(also see Appendix~\ref{app:smalldomain}) we see that LUCIR, GD and BiC perform the best among all methods. In particular, LUCIR achieves 73.0\% average accuracy after 25 tasks, which is relatively high compared to previous experiments on \mbox{CIFAR-100}. This indicates that LUCIR might be more indicated for smaller domain shifts. \mbox{FT-E} performs only 4.2 points lower than LUCIR and close to \mbox{EWC-E}, which also performs well with small domain shifts between tasks. EEIL shows competitive performance on the first 13 tasks, but starts to decline for the remaining ones.

\minisection{Larger domain shift:}
We are the first to compare \mbox{class-IL} methods to incrementally learn classes from multiple datasets. As a consequence, tasks have large domain shifts and different numbers of classes. We use six fine-grained datasets (Flowers, Scenes, Birds, Cars, Aircraft and Actions) learned sequentially using ResNet-18 from scratch with a growing memory of 5 exemplars per class. The number of classes varies among the tasks, but the classes within each of them are closely related. In Fig.~\ref{fig:multi_plot} we see that most approaches have a similar performance, unlike in previous experiments. It is noticeable that bias-correction methods do not have a clear advantage compared to other approaches. It seems that when the domain shift between tasks is large, inter-task confusion becomes the major cause for catastrophic forgetting. Solving the task-recency bias provides a lower performance advantage than in other scenarios and only improves the outputs of the corresponding task. Forgetting caused by the large weight and activation drift deriving from the large domain shifts seems to dominate.
The use of triple distillation from GD (without external data) is less effective, unlike in other scenarios. This is probably because activation-based regularization methods are less effective when large domain shifts occur between tasks~\cite{aljundi2016expert}. The fact that no method clearly outperforms the FT-E baseline shows that scenarios with large domain shifts, where catastrophic forgetting is caused by inter-task confusion, are still an important direction of study since most proposed methods focus on weight drift, activation drift, or task-recency bias.

\figExpSingleColumn{1.0}{10 10 10 20}{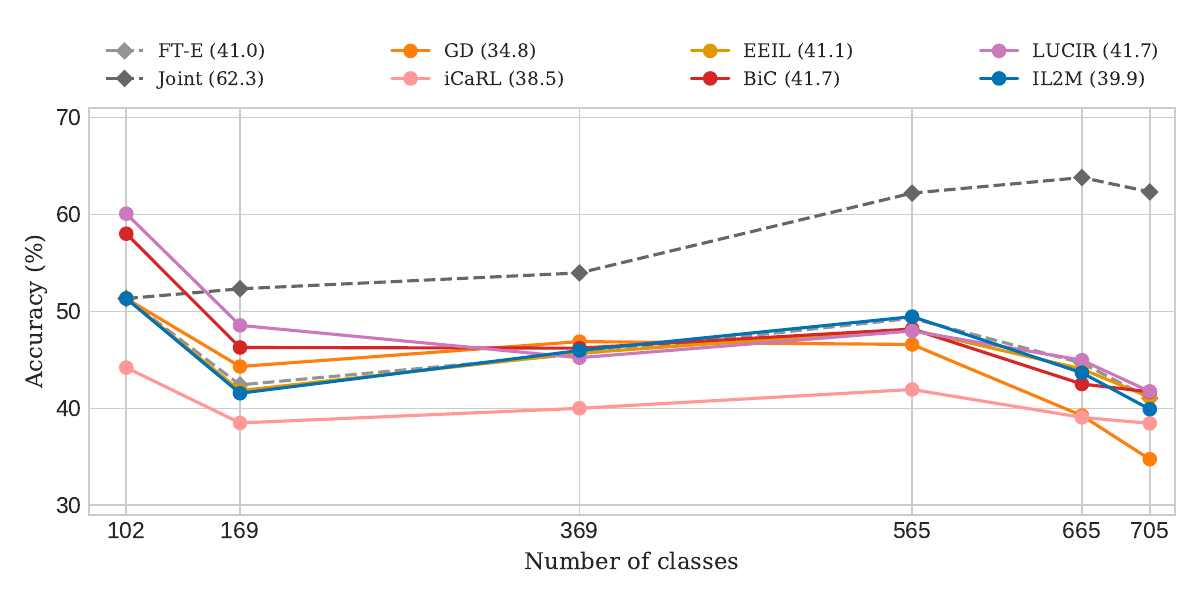}{Large domain shifts with multiple fine-grained datasets (Flowers, Scenes, Birds, Cars, Aircraft, Actions).}

\figInterspersed{Forgetting when revisiting old domains with new classes from different fine-grained datasets on AlexNet.}

\subsection{On ``interspersed'' domains:}
We propose another scenario not yet explored in \mbox{class-IL}: revisiting learned distributions to learn new classes. We learn four fine-grained datasets split into four tasks of ten classes each for a total of 16 tasks and 160 classes. A group consists of four tasks, one from each dataset in this order: Flowers, Birds, Actions, Aircraft. The experiment consists of four group repetitions, where each group contains different classes. This allows us to analyze how \mbox{class-IL} methods perform when similar tasks re-appear after learning different tasks. We refer to this scenario as ``interspersed'' domains since classes from each domain are distributed across tasks.

Results of forgetting on the first group during the whole sequence are presented in Fig.~\ref{fig:interspersed_forg}. LUCIR suffers quite a large loss on the first task at the beginning of the sequence and after the second group is learned, never recovering any performance for that task. However, LUCIR shows very little forgetting for the remaining tasks in the sequence. This seems to be related to the preference of LUCIR to have a larger first task with more diverse feature representations, as also observed in earlier experiments. For the remaining methods, the first task has a lot of variation with a general decaying trend. BiC has an initial drop right after learning each of the other tasks, but manages to prevent further forgetting, though with some variability on the first Aircraft task. LwF-E and EEIL have a more cyclic pattern of forgetting and recovering. Forgetting is more pronounced when the task being learned is of the same dataset as the current one, and seems to slightly recover when learning less similar tasks. Finally, the forgetting of IL2M shows a lot of variation, which might be related to the lack of a distillation loss keeping new representations closer to previous ones.

\subsection{On network architectures}
\label{sec:exp-networks}
We compare the four most competitive methods over a range of different network architectures in Fig.~\ref{fig:diff-nets}. Specifically, we use AlexNet~\cite{krizhevsky2012imagenet}, ResNet-18~\cite{he2016deep}, VGG-11~\cite{simonyan2014very}, GoogleNet~\cite{szegedy2015going} and MobileNet~\cite{sandler2018mobilenetv2}. An interesting observation is that for different networks, the performance rankings of the methods can change completely. For instance, in architectures which do not use skip connections (AlexNet, VGG-11), iCaRL performs the best. On the other hand, BiC performs worse without skip connections, but performs the best with architectures that have them (ResNet-18, MobileNet and GoogleNet). IL2M is more consistent compared to other methods using different networks, never having the best nor the worst performance. Networks without skip connections seem to reduce forgetting for iCaRL and IL2M. EEIL suffers more forgetting compared to other methods across different networks.

ResNet-18 obtains the best result among all networks with BiC. Note that in most of the literature, ResNet-18 is used as the default network for this scenario and similar ones. However, as shown above, it seems that methods benefit from architectures differently. Another interesting observation is that MobileNet, which has the lowest number of parameters/operations and can run on devices with limited capacity, has very competitive results compared to the other networks. These results show that existing IL approaches can be applied to different architectures with comparable results to the scenarios presented in the literature.

\figdifferentnetworks{0 0 0 0}{Average accuracy after 10 tasks on ImageNet-Subset-100 (10/10) with different networks trained from scratch. From left to right: MobileNet (2017), GoogleNet (2014), ResNet-18 (2015), AlexNet (2012) and VGG-11 (2014).}

\figExpSingleColumn{1.0}{0 35 0 75}{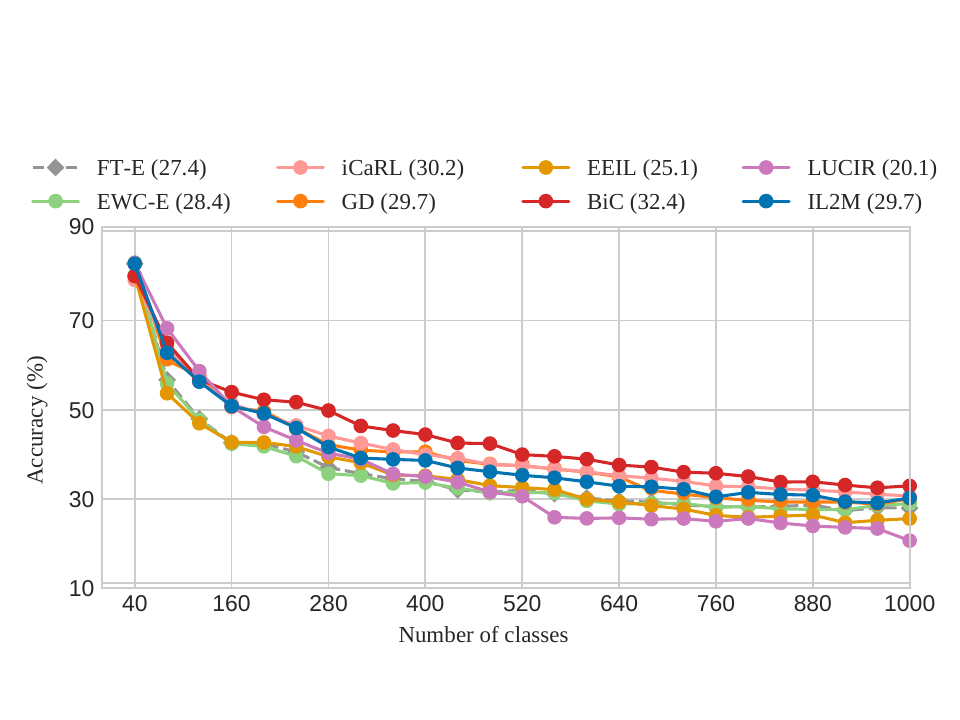}{ImageNet (25/40) on ResNet-18 with growing memory of 20 exemplars per class and herding sampling.}

\subsection{On large-scale scenarios}
Finally, we compare different methods using ResNet-18 on ImageNet (25/40) with a growing memory of 20 exemplars per class. Fig.~\ref{fig:imagenet_plot} shows that BiC and iCaRL achieve the best performance with 32.4\% and 30.2\% average accuracy after 25 tasks, respectively. Surprisingly, \mbox{EWC-E} and \mbox{FT-E} outperform LUCIR, EEIL and LwF-E (19.8\%) in this setting. Note that in other settings, IL2M and LUCIR often perform better than \mbox{EWC-E} and \mbox{FT-E}. We note that BiC, iCaRL, IL2M, GD and LUCIR avoid a larger initial drop in performance during the first four tasks compared to other methods and continue learning without major drops in performance, with the exception of LUCIR. Of the rest of the methods, \mbox{EWC-E}, \mbox{FT-E} and EEIL seem to stabilize after the initial drop and exhibit less forgetting as new tasks are added. In scenarios with a larger number of classes and more variability, methods which can easily handle early tasks will perform better afterwards. On the second half of the sequence, most approaches have a stable behaviour since the network has learned a robust representation from the initial tasks.

%%%%%%%%%%%%%%%%%%%%%%%%%%%%%%%%%%%%%
\section{Emerging trends in class-IL}\label{sec:trends}
Here we discuss some recent developments in \mbox{class-IL} that we think will play an important role in the coming years.

\minisection{Exemplar learning.} Recently, an exciting new direction has emerged that parametrizes exemplars and optimizes them to prevent forgetting~\cite{chaudhry2021using, liu2020mnemonics}. This enables much more efficient use of available storage. Liu et al.~\cite{liu2020mnemonics} propose Mnemonics Training, a method that trains the parametrized exemplars. The exemplars are optimized to prevent the forgetting when evaluated on the current task data. Chaudry et al.~\cite{chaudhry2021using} generalize the theory to a streaming setting, where the learning of exemplars does not require multiple loops over the data for every task. Optimizing the available storage by computing more efficient exemplars is expected to attract more research in the coming years.

\minisection{Feature rehearsal.}
Pseudo-rehearsal is a good alternative to storing exemplars~\cite{ostapenko2019learning,shin2017continual,wu2018memory}. It learns a separate network that generates images of previous tasks. However, current state-of-the-art image generation methods struggle to realistically generate complex image data, and therefore this approach has been applied to  simple datasets and is known to obtain unsatisfying results on complex ones. To address this problem, some works have proposed to perform \emph{feature} replay instead of image replay~\cite{xiang2019incremental, liu2020generative, iscen2020memory}, where instead a generator is trained to generate features at some hidden layer of the network. In this way, rehearsal can also be applied to complex datasets. Another closely related line of research is based on the observation that storing feature exemplars is much more compact than storing images~\cite{hayes2020remind}. Moving away from image replay towards different variants of feature replay is expected to gain traction.

\minisection{Self- and unsupervised incremental learning.} Being able to incrementally learn representations from an unsupervised data stream is a desirable feature in any learning system. This direction applied to \mbox{class-IL} has received relatively little attention to date. Rao et al.~\cite{rao2019continual} propose a method that performs explicit task classification and fits a mixture of Gaussians on the learned representations. They also explore scenarios with smooth transitions from one task to another. Still in its infancy, more research on unsupervised incremental learning is expected in coming years. In addition, leveraging the power of self-supervised representation learning~\cite{jing2020selfsuper} is only little explored within the context of IL, and is expected to gain interest. 

\minisection{Beyond cross-entropy loss.} Several recent works show that the cross-entropy loss might be responsible for high levels of catastrophic forgetting~\cite{yu2020semantic,li2020energy}. Less forgetting has been reported by replacing the cross-entropy loss with a metric learning loss~\cite{yu2020semantic} or by using an energy-based method~\cite{li2020energy}. Combining these methods with the other class-IL categories, such as bias-correction and rehearsal, is expected to result in highly competitive methods.

\minisection{Meta-learning.} Meta-learning aims to learn new tasks leveraging information accrued while solving related tasks~\cite{schmidhuber1987evolutionary}. Riemer et al.~\cite{riemer2019learning} show that such a method can learn parameters that reduce interference of future gradients and improves transfer based on future gradients. Javed and White~\cite{javed2019meta} explicitly learn a representation for continual learning that avoids interference and promotes future learning. These initial works have shown the potential of meta-learning on small datasets. However, we expect these techniques to be further developed in the coming years, and will start to obtain results on more complex datasets like the ones considered in our evaluation.

\minisection{Task-free settings.} Many practical applications do not fit well into the experimental setup with non-overlapping tasks. A more realistic scenario is one where there are no clear task boundaries and the distribution over classes changes gradually. This scenario is expected to receive increased attention in near future. This setting was studied in several early task-aware continual learning works, including EWC~\cite{kirkpatrick2017overcoming} and P\&C~\cite{schwarz2018progress}. The transition to the task-free setting is not straight-forward, since many methods have inherent operations that are performed on the task boundaries: replacing the old model, updating of importance weights, etc.  Recently, several works for \mbox{class-IL} started addressing this setting~\cite{aljundi2019task,li2020energy,rajasegaran2020itaml}.

%%%%%%%%%%%%%%%%%%%%%%%%%%%%%%%%%%%%%
\section{Conclusions}\label{sec:conclusions}
We performed an extensive survey of class-incremental learning. We organized the proposed approaches along three main lines: regularization, rehearsal, and bias-correction. In addition, we provided extensive experiments in which we compare thirteen methods on a wide range of incremental learning scenarios. Here we briefly enumerate the main conclusions from these experiments:
\begin{itemize}
    \item When comparing exemplar-free methods, LwF obtains the best results (see Table~\ref{tab:regul-methods}). Among the other regularization methods, data regularization (LwM) obtains superior results compared to weight regularization (EWC and MAS). Exemplar-free methods can currently not compete with exemplar rehearsal methods, and given the more restrictive setting in which they operate, we advocate comparing them separately.
    \item When combining LwF with exemplars, we confirm previous results~\cite{belouadah2019il2m} showing that the added regularization does not improve results and the baseline method of finetuning with exemplars performs better (see Table~\ref{tab:regul-methods}).
    \item Allowing a model to specialize on the current task and then using distillation to combine this new knowledge with the knowledge of the previous tasks, as is done by GD, obtains excellent results for class-IL (see Fig.~\ref{fig:cifar100_fixd_plot_cifar100_fixd_lft_plot}).
    \item We found that, in several scenarios, weight regularization method EWC-E outperforms data regularization method LwF-E significantly (see Figs.~\ref{fig:cifar100_fixd_plot_cifar100_fixd_lft_plot}, ~\ref{fig:vggface2} and ~\ref{fig:imagenet_plot}), showing that the IL community choice of data regularization with LwF (see Fig.~\ref{fig:approaches_graph}) instead of weight regularization should be reconsidered.
    \item Herding is, on average, marginally better than random exemplar sampling for longer sequences of tasks (see Table~\ref{tab:ex-sampling}). However this is only statistically significant for some methods (see Appendix~\ref{app:sampling_strats}).
    \item Methods that explicitly address the task-recency bias obtain better performance for \mbox{class-IL} (see Figs.~\ref{fig:cifar100_fixd_plot_cifar100_fixd_lft_plot},~\ref{fig:vggface2},~\ref{fig:multi_plot},~\ref{fig:imagenet_plot}): we found that BiC obtains state-of-the-art on several experiments (notably on ImageNet). IL2M obtains consistent good performance on most datasets. Also, iCaRL and EEIL obtain good performance on several datasets, but fail to outperform the baseline FT-E on others. Methods like LUCIR require a good starting representation -- for example in the scenario with the larger first task or smaller domain shifts, LUCIR can be state-of-the-art.
    \item Current methods have mainly presented results on datasets with small domain shifts (typically random class orderings from a single dataset). When considering large domain shifts none of the methods significantly outperform the baseline FT-E (see Fig.~\ref{fig:multi_plot}). 
    Large domain shift scenarios have been considered for task-IL, but our results show that they require new techniques to obtain satisfactory results in \mbox{class-IL} settings.
    \item We are the first to compare \mbox{class-IL} methods on a wide range of network architectures, showing that current \mbox{class-IL} works on a variety of networks. Results show that most are sensitive to architecture and rankings change depending on the network used. It is quite clear that using a network with skip connections favors some methods, while their absence favors others.
\end{itemize}

%%%%%%%%%%%%%%%%%%%%%%%%%%%%%%%%%%%%%
\section*{Acknowledgments}
We acknowledge the support from Huawei Kirin Solution. Masana acknowledges grant \mbox{2019-FI\_B2-00189} from Generalitat de Catalunya, Van de Weijer the Spanish project \mbox{PID2019-104174GB-I00/AEI/10.13039/501100011033}, and Bagdanov the European Commission, Horizon 2020 Programme, grant number 951911 - AI4Media.

%%%%%%%%%%%%%%%%%%%%%%%%%%%%%%%%%%%%%
\bibliographystyle{IEEEtran}
\bibliography{IEEEabrv, refs}

\begin{thebibliography}{100}
\providecommand{\url}[1]{#1}
\csname url@samestyle\endcsname
\providecommand{\newblock}{\relax}
\providecommand{\bibinfo}[2]{#2}
\providecommand{\BIBentrySTDinterwordspacing}{\spaceskip=0pt\relax}
\providecommand{\BIBentryALTinterwordstretchfactor}{4}
\providecommand{\BIBentryALTinterwordspacing}{\spaceskip=\fontdimen2\font plus
\BIBentryALTinterwordstretchfactor\fontdimen3\font minus
  \fontdimen4\font\relax}
\providecommand{\BIBforeignlanguage}[2]{{%
\expandafter\ifx\csname l@#1\endcsname\relax
\typeout{** WARNING: IEEEtran.bst: No hyphenation pattern has been}%
\typeout{** loaded for the language `#1'. Using the pattern for}%
\typeout{** the default language instead.}%
\else
\language=\csname l@#1\endcsname
\fi
#2}}
\providecommand{\BIBdecl}{\relax}
\BIBdecl

\bibitem{rebuffi2017icarl}
S.-A. Rebuffi, A.~Kolesnikov, G.~Sperl, and C.~H. Lampert, ``icarl: Incremental
  classifier and representation learning,'' in \emph{Proc. IEEE Conference on
  Computer Vision and Pattern Recognition}, 2017.

\bibitem{thrun1996learning}
S.~Thrun, ``Is learning the n-th thing any easier than learning the first?'' in
  \emph{Proc. Adv. Neural Information Processing Systems}, 1996.

\bibitem{french1999catastrophic}
R.~M. French, ``Catastrophic forgetting in connectionist networks,''
  \emph{Trends in cognitive sciences}, 1999.

\bibitem{goodfellow2013empirical}
I.~J. Goodfellow, M.~Mirza, D.~Xiao, A.~Courville, and Y.~Bengio, ``An
  empirical investigation of catastrophic forgetting in gradient-based neural
  networks,'' in \emph{Proc. International Conference on Learning
  Representations}, 2014.

\bibitem{kirkpatrick2017overcoming}
J.~Kirkpatrick, R.~Pascanu, N.~Rabinowitz, J.~Veness, G.~Desjardins, A.~A.
  Rusu, K.~Milan, J.~Quan, T.~Ramalho, A.~Grabska-Barwinska \emph{et~al.},
  ``Overcoming catastrophic forgetting in neural networks,'' \emph{National
  Academy of Sciences}, 2017.

\bibitem{mccloskey1989catastrophic}
M.~McCloskey and N.~J. Cohen, ``Catastrophic interference in connectionist
  networks: The sequential learning problem,'' in \emph{Psychology of learning
  and motivation}, 1989.

\bibitem{chaudhry2018riemannian}
A.~Chaudhry, P.~K. Dokania, T.~Ajanthan, and P.~H. Torr, ``Riemannian walk for
  incremental learning:~understanding forgetting and intransigence,'' in
  \emph{Proc. European Conference on Computer Vision}, 2018.

\bibitem{vandeven2019three}
G.~M. van~de Ven and A.~S. Tolias, ``Three scenarios for continual learning,''
  in \emph{Proc. Adv. Neural Information Processing Systems Workshop on
  Continual Learning}, 2018.

\bibitem{thrun1995lifelong}
S.~Thrun, ``A lifelong learning perspective for mobile robot control,'' in
  \emph{Intelligent Robots and Systems}, 1995.

\bibitem{chen2018lifelong}
Z.~Chen and B.~Liu, ``Lifelong machine learning,'' \emph{Synthesis Lectures on
  Artificial Intelligence and Machine Learning}, 2018.

\bibitem{aljundi2016expert}
R.~Aljundi, P.~Chakravarty, and T.~Tuytelaars, ``Expert gate: Lifelong learning
  with a network of experts,'' in \emph{Proc. IEEE Conference on Computer
  Vision and Pattern Recognition}, 2017.

\bibitem{chaudhry2018efficient}
A.~Chaudhry, M.~Ranzato, M.~Rohrbach, and M.~Elhoseiny, ``Efficient lifelong
  learning with a-gem,'' in \emph{Proc. International Conference on Learning
  Representations}, 2019.

\bibitem{delange2021continual}
M.~Delange, R.~Aljundi, M.~Masana, S.~Parisot, X.~Jia, A.~Leonardis,
  G.~Slabaugh, and T.~Tuytelaars, ``A continual learning survey: Defying
  forgetting in classification tasks,'' \emph{IEEE Transactions on Pattern
  Analysis and Machine Intelligence}, 2021.

\bibitem{lesort2020continual}
T.~Lesort, V.~Lomonaco, A.~Stoian, D.~Maltoni, D.~Filliat, and
  N.~D{\'\i}az-Rodr{\'\i}guez, ``Continual learning for robotics: Definition,
  framework, learning strategies, opportunities and challenges,''
  \emph{Information Fusion}, 2020.

\bibitem{mcclure2018distributed}
P.~McClure, C.~Y. Zheng, J.~R. Kaczmarzyk, J.~A. Lee, S.~S. Ghosh, D.~Nielson,
  P.~Bandettini, and F.~Pereira, ``Distributed weight consolidation: a brain
  segmentation case study,'' in \emph{Proc. Adv. Neural Information Processing
  Systems}, 2018.

\bibitem{strubell2019energy}
E.~Strubell, A.~Ganesh, and A.~McCallum, ``Energy and policy considerations for
  deep learning in nlp,'' in \emph{Assoc. for Comput. Linguistics}, 2019.

\bibitem{sharir2020cost}
O.~Sharir, B.~Peleg, and Y.~Shoham, ``The cost of training nlp models: A
  concise overview,'' \emph{arXiv}, 2020.

\bibitem{lopez2017gradient}
D.~Lopez-Paz and M.~Ranzato, ``Gradient episodic memory for continual
  learning,'' in \emph{Proc. Adv. Neural Information Processing Systems}, 2017.

\bibitem{mai2021online}
Z.~Mai, R.~Li, J.~Jeong, D.~Quispe, H.~Kim, and S.~Sanner, ``Online continual
  learning in image classification: An empirical survey,''
  \emph{Neurocomputing}, vol. 469, pp. 28--51, 2022.

\bibitem{li2017learning}
Z.~Li and D.~Hoiem, ``Learning without forgetting,'' \emph{IEEE Transactions on
  Pattern Analysis and Machine Intelligence}, 2017.

\bibitem{hou2019learning}
S.~Hou, X.~Pan, C.~C. Loy, Z.~Wang, and D.~Lin, ``Learning a unified classifier
  incrementally via rebalancing,'' in \emph{Proc. IEEE International Conference
  on Computer Vision}, 2019.

\bibitem{wu2019large}
Y.~Wu, Y.~Chen, L.~Wang, Y.~Ye, Z.~Liu, Y.~Guo, and Y.~Fu, ``Large scale
  incremental learning,'' in \emph{Proc. IEEE Conference on Computer Vision and
  Pattern Recognition}, 2019.

\bibitem{castro2018end}
F.~M. Castro, M.~J. Mar{\'\i}n-Jim{\'e}nez, N.~Guil, C.~Schmid, and K.~Alahari,
  ``End-to-end incremental learning,'' in \emph{Proc. European Conference on
  Computer Vision}, 2018.

\bibitem{belouadah2019il2m}
E.~Belouadah and A.~Popescu, ``Il2m:~class incremental learning with dual
  memory,'' in \emph{Proc. IEEE International Conference on Computer Vision},
  2019.

\bibitem{schwarz2018progress}
J.~Schwarz, J.~Luketina, W.~M. Czarnecki, A.~Grabska-Barwinska, Y.~W. Teh,
  R.~Pascanu, and R.~Hadsell, ``Progress \& compress: A scalable framework for
  continual learning,'' in \emph{Proc. International Conference on Machine
  Learning}, 2018.

\bibitem{rusu2016progressive}
A.~A. Rusu, N.~C. Rabinowitz, G.~Desjardins, H.~Soyer, J.~Kirkpatrick,
  K.~Kavukcuoglu, R.~Pascanu, and R.~Hadsell, ``Progressive neural networks,''
  \emph{arXiv}, 2016.

\bibitem{rajasegaran2019random}
J.~Rajasegaran, M.~Hayat, S.~Khan, F.~S. Khan, and L.~Shao, ``Random path
  selection for incremental learning,'' in \emph{Proc. Adv. Neural Information
  Processing Systems}, 2019.

\bibitem{hayes2020remind}
T.~L. Hayes, K.~Kafle, R.~Shrestha, M.~Acharya, and C.~Kanan, ``Remind your
  neural network to prevent catastrophic forgetting,'' in \emph{Proc. European
  Conference on Computer Vision}, 2020.

\bibitem{mallya2018packnet}
A.~Mallya and S.~Lazebnik, ``Packnet: Adding multiple tasks to a single network
  by iterative pruning,'' in \emph{Proc. IEEE Conference on Computer Vision and
  Pattern Recognition}, 2018.

\bibitem{shin2017continual}
H.~Shin, J.~K. Lee, J.~Kim, and J.~Kim, ``Continual learning with deep
  generative replay,'' in \emph{Proc. Adv. Neural Information Processing
  Systems}, 2017.

\bibitem{nguyen2018variational}
C.~V. Nguyen, Y.~Li, T.~D. Bui, and R.~E. Turner, ``Variational continual
  learning,'' in \emph{Proc. International Conference on Learning
  Representations}, 2018.

\bibitem{mermillod2013stability}
M.~Mermillod, A.~Bugaiska, and P.~Bonin, ``The stability-plasticity dilemma:
  Investigating the continuum from catastrophic forgetting to age-limited
  learning effects,'' \emph{Frontiers in psychology}, 2013.

\bibitem{aljundi2018memory}
R.~Aljundi, F.~Babiloni, M.~Elhoseiny, M.~Rohrbach, and T.~Tuytelaars, ``Memory
  aware synapses: Learning what (not) to forget,'' in \emph{Proc. European
  Conference on Computer Vision}, 2018.

\bibitem{zenke2017continual}
F.~Zenke, B.~Poole, and S.~Ganguli, ``Continual learning through synaptic
  intelligence,'' in \emph{Proc. International Conference on Machine Learning},
  2017.

\bibitem{jung2016less}
H.~Jung, J.~Ju, M.~Jung, and J.~Kim, ``Less-forgetting learning in deep neural
  networks,'' \emph{arXiv}, 2016.

\bibitem{wu2018memory}
C.~Wu, L.~Herranz, X.~Liu, Y.~Wang, J.~van~de Weijer, and B.~Raducanu, ``Memory
  replay {GANs}: learning to generate images from new categories without
  forgetting,'' in \emph{Proc. Adv. Neural Information Processing Systems},
  2018.

\bibitem{xiang2019incremental}
Y.~Xiang, Y.~Fu, P.~Ji, and H.~Huang, ``Incremental learning using conditional
  adversarial networks,'' in \emph{Proc. IEEE International Conference on
  Computer Vision}, 2019.

\bibitem{lee2017overcoming}
S.-W. Lee, J.-H. Kim, J.~Jun, J.-W. Ha, and B.-T. Zhang, ``Overcoming
  catastrophic forgetting by incremental moment matching,'' in \emph{Proc. Adv.
  Neural Information Processing Systems}, 2017.

\bibitem{liu2018rotate}
X.~Liu, M.~Masana, L.~Herranz, J.~Van~de Weijer, A.~M. Lopez, and A.~D.
  Bagdanov, ``Rotate your networks: Better weight consolidation and less
  catastrophic forgetting,'' in \emph{International Conference on Pattern
  Recognition}, 2018.

\bibitem{rannen2017encoder}
A.~Rannen, R.~Aljundi, M.~B. Blaschko, and T.~Tuytelaars, ``Encoder based
  lifelong learning,'' in \emph{Proc. IEEE Conference on Computer Vision and
  Pattern Recognition}, 2017.

\bibitem{silver2002task}
D.~L. Silver and R.~E. Mercer, ``The task rehearsal method of life-long
  learning: Overcoming impoverished data,'' in \emph{Conference of the Canadian
  Society for Computational Studies of Intelligence}, 2002.

\bibitem{zhang2020class}
J.~Zhang, J.~Zhang, S.~Ghosh, D.~Li, S.~Tasci, L.~Heck, H.~Zhang, and C.-C.~J.
  Kuo, ``Class-incremental learning via deep model consolidation,'' in
  \emph{Proc. IEEE Winter Conference on Applications of Computer Vision}, 2020.

\bibitem{lee2019overcoming}
K.~Lee, K.~Lee, J.~Shin, and H.~Lee, ``Overcoming catastrophic forgetting with
  unlabeled data in the wild,'' in \emph{Proc. IEEE International Conference on
  Computer Vision}, 2019.

\bibitem{lee2020continual}
J.~Lee, H.~G. Hong, D.~Joo, and J.~Kim, ``Continual learning with extended
  kronecker-factored approximate curvature,'' in \emph{Proc. IEEE Conference on
  Computer Vision and Pattern Recognition}, 2020.

\bibitem{bucilua2006model}
C.~Buciluǎ, R.~Caruana, and A.~Niculescu-Mizil, ``Model~compres-sion,'' in
  \emph{International Conference on Knowledge Discovery and Data Mining}, 2006.

\bibitem{hinton2014distilling}
G.~Hinton, O.~Vinyals, and J.~Dean, ``Distilling the knowledge in a neural
  network,'' in \emph{Proc. Adv. Neural Information Processing Systems Workshop
  on Deep Learning}, 2014.

\bibitem{liu2020mnemonics}
Y.~Liu, A.-A. Liu, Y.~Su, B.~Schiele, and Q.~Sun, ``Mnemonics training:
  Multi-class incremental learning without forgetting,'' in \emph{Proc. IEEE
  Conference on Computer Vision and Pattern Recognition}, 2020.

\bibitem{zagoruyko2016paying}
S.~Zagoruyko and N.~Komodakis, ``Paying more attention to attention: Improving
  the performance of convolutional neural networks via attention transfer,'' in
  \emph{Proc. International Conference on Learning Representations}, 2017.

\bibitem{dhar2019learning}
P.~Dhar, R.~V. Singh, K.-C. Peng, Z.~Wu, and R.~Chellappa, ``Learning without
  memorizing,'' in \emph{Proc. IEEE Conference on Computer Vision and Pattern
  Recognition}, 2019.

\bibitem{selvaraju2017grad}
R.~R. Selvaraju, M.~Cogswell, A.~Das, R.~Vedantam, D.~Parikh, and D.~Batra,
  ``Grad-cam: Visual explanations from deep networks via gradient-based
  localization,'' in \emph{Proc. IEEE International Conference on Computer
  Vision}, 2017.

\bibitem{Hou_2018_ECCV}
S.~Hou, X.~Pan, C.~C. Loy, Z.~Wang, and D.~Lin, ``Lifelong learning via
  progressive distillation and retrospection,'' in \emph{Proc. European
  Conference on Computer Vision}, 2018.

\bibitem{javed2018revisiting}
K.~Javed and F.~Shafait, ``Revisiting distillation and incremental classifier
  learning,'' in \emph{Asian Conference on Computer Vision}, 2018.

\bibitem{ostapenko2019learning}
O.~Ostapenko, M.~Puscas, T.~Klein, P.~J{\"a}hnichen, and M.~Nabi, ``Learning to
  remember: A synaptic plasticity driven framework for continual learning,'' in
  \emph{Proc. IEEE Conference on Computer Vision and Pattern Recognition},
  2019.

\bibitem{kemker2017fearnet}
R.~Kemker and C.~Kanan, ``Fearnet: Brain-inspired model for incremental
  learning,'' in \emph{Proc. International Conference on Learning
  Representations}, 2018.

\bibitem{welling2009herding}
M.~Welling, ``Herding dynamical weights to learn,'' in \emph{Proc.
  International Conference on Machine Learning}, 2009.

\bibitem{prabhu2020gdumb}
A.~Prabhu, P.~H. Torr, and P.~K. Dokania, ``Gdumb: A simple approach that
  questions our progress in continual learning,'' in \emph{Proc. European
  Conference on Computer Vision}.\hskip 1em plus 0.5em minus 0.4em\relax
  Springer, 2020, pp. 524--540.

\bibitem{titsias2020functional}
M.~K. Titsias, J.~Schwarz, A.~G. d.~G. Matthews, R.~Pascanu, and Y.~W. Teh,
  ``Functional regularisation for continual learning with gaussian processes,''
  in \emph{Proc. International Conference on Learning Representations}, 2020.

\bibitem{pan2020continual}
P.~Pan, S.~Swaroop, A.~Immer, R.~Eschenhagen, R.~E. Turner, and M.~E. Khan,
  ``Continual deep learning by functional regularisation of memorable past,''
  in \emph{Proc. Adv. Neural Information Processing Systems}, 2020.

\bibitem{ratcliff1990connectionist}
R.~Ratcliff, ``Connectionist models of recognition memory: constraints imposed
  by learning and forgetting functions.'' \emph{Psychological review}, 1990.

\bibitem{krizhevsky2012imagenet}
A.~Krizhevsky, I.~Sutskever, and G.~E. Hinton, ``Imagenet classification with
  deep convolutional neural networks,'' in \emph{Proc. Adv. Neural Information
  Processing Systems}, 2012.

\bibitem{parisi2019continual}
G.~I. Parisi, R.~Kemker, J.~L. Part, C.~Kanan, and S.~Wermter, ``Continual
  lifelong learning with neural networks: A review,'' \emph{Neural Networks},
  2019.

\bibitem{pfulb2019comprehensive}
B.~Pf{\"u}lb and A.~Gepperth, ``A comprehensive, application-oriented study of
  catastrophic forgetting in dnns,'' in \emph{Proc. International Conference on
  Learning Representations}, 2019.

\bibitem{belouadah2020comprehensive}
E.~Belouadah, A.~Popescu, and I.~Kanellos, ``A comprehensive study of class
  incremental learning algorithms for visual tasks,'' \emph{Neural Networks},
  2020.

\bibitem{masana2020ternary}
M.~Masana, T.~Tuytelaars, and J.~van~de Weijer, ``Ternary feature masks:
  continual learning without any forgetting,'' in \emph{Proc. IEEE Conference
  on Computer Vision and Pattern Recognition Workshops}, 2021.

\bibitem{mallya2018piggyback}
A.~Mallya, D.~Davis, and S.~Lazebnik, ``Piggyback: Adapting a single network to
  multiple tasks by learning to mask weights,'' in \emph{Proc. European
  Conference on Computer Vision}, 2018.

\bibitem{serra2018overcoming}
J.~Serra, D.~Suris, M.~Miron, and A.~Karatzoglou, ``Overcoming catastrophic
  forgetting with hard attention to the task,'' in \emph{Proc. International
  Conference on Machine Learning}, 2018.

\bibitem{rosenfeld2018incremental}
A.~Rosenfeld and J.~K. Tsotsos, ``Incremental learning through deep
  adaptation,'' \emph{IEEE Transactions on Pattern Analysis and Machine
  Intelligence}, 2018.

\bibitem{fernando2017pathnet}
C.~Fernando, D.~Banarse, C.~Blundell, Y.~Zwols, D.~Ha, A.~A. Rusu, A.~Pritzel,
  and D.~Wierstra, ``Pathnet: Evolution channels gradient descent in super
  neural networks,'' \emph{arXiv}, 2017.

\bibitem{xiao2014error}
T.~Xiao, J.~Zhang, K.~Yang, Y.~Peng, and Z.~Zhang, ``Error-driven incremental
  learning in deep convolutional neural network for large-scale image
  classification,'' in \emph{ACM International Conference on Multimedia}, 2014.

\bibitem{ebrahimi2020adversarial}
S.~Ebrahimi, F.~Meier, R.~Calandra, T.~Darrell, and M.~Rohrbach, ``Adversarial
  continual learning,'' in \emph{Proc. European Conference on Computer Vision},
  2020.

\bibitem{chaudhry2019continual}
A.~Chaudhry, M.~Rohrbach, M.~Elhoseiny, T.~Ajanthan, P.~K. Dokania, P.~H. Torr,
  and M.~Ranzato, ``Continual learning with tiny episodic memories,'' in
  \emph{Proc. International Conference on Machine Learning}, 2019.

\bibitem{mnih2013playing}
V.~Mnih, K.~Kavukcuoglu, D.~Silver, A.~Graves, I.~Antonoglou, D.~Wierstra, and
  M.~Riedmiller, ``Playing atari with deep reinforcement learning,'' in
  \emph{Proc. Adv. Neural Information Processing Systems Workshop on Deep
  Learning}, 2013.

\bibitem{rolnick2019experience}
D.~Rolnick, A.~Ahuja, J.~Schwarz, T.~P. Lillicrap, and G.~Wayne, ``Experience
  replay for continual learning,'' in \emph{Proc. Adv. Neural Information
  Processing Systems}, 2019.

\bibitem{aljundi2019gradient}
R.~Aljundi, M.~Lin, B.~Goujaud, and Y.~Bengio, ``Gradient based sample
  selection for online continual learning,'' in \emph{Proc. Adv. Neural
  Information Processing Systems}, 2019.

\bibitem{aljundi2019online}
R.~Aljundi, E.~Belilovsky, T.~Tuytelaars, L.~Charlin, M.~Caccia, M.~Lin, and
  L.~Page-Caccia, ``Online continual learning with maximal interfered
  retrieval,'' in \emph{Proc. Adv. Neural Information Processing Systems},
  2019.

\bibitem{riemer2019scalable}
M.~Riemer, T.~Klinger, D.~Bouneffouf, and M.~Franceschini, ``Scalable
  recollections for continual lifelong learning,'' in \emph{Proc. AAAI
  Conference Artificial Intelligence}, 2019.

\bibitem{riemer2019learning}
M.~Riemer, I.~Cases, R.~Ajemian, M.~Liu, I.~Rish, Y.~Tu, and G.~Tesauro,
  ``Learning to learn without forgetting by maximizing transfer and minimizing
  interference,'' in \emph{Proc. International Conference on Learning
  Representations}, 2019.

\bibitem{nichol2018first}
A.~Nichol, J.~Achiam, and J.~Schulman, ``On first-order meta-learning
  algorithms,'' \emph{arXiv}, 2018.

\bibitem{farquhar2019unifying}
S.~Farquhar and Y.~Gal, ``A unifying bayesian view of continual learning,'' in
  \emph{Proc. Adv. Neural Information Processing Systems Workshop on Deep
  Learning}, 2019.

\bibitem{ahn2019uncertainty}
H.~Ahn, S.~Cha, D.~Lee, and T.~Moon, ``Uncertainty-based continual learning
  with adaptive regularization,'' in \emph{Proc. Adv. Neural Information
  Processing Systems}, 2019.

\bibitem{chen2018facilitating}
Y.~Chen, T.~Diethe, and N.~Lawrence, ``Facilitating bayesian continual learning
  by natural gradients and stein gradients,'' in \emph{Proc. Adv. Neural
  Information Processing Systems Workshop on Continual Learning}, 2018.

\bibitem{swaroop2018improving}
S.~Swaroop, C.~V. Nguyen, T.~D. Bui, and R.~E. Turner, ``Improving and
  understanding variational continual learning,'' in \emph{Proc. Adv. Neural
  Information Processing Systems Workshop on Continual Learning}, 2018.

\bibitem{adel2020continual}
T.~Adel, H.~Zhao, and R.~E. Turner, ``Continual learning with adaptive weights
  (claw),'' in \emph{Proc. International Conference on Learning
  Representations}, 2020.

\bibitem{ebrahimi2020uncertainty}
S.~Ebrahimi, M.~Elhoseiny, T.~Darrell, and M.~Rohrbach, ``Uncertainty-guided
  continual learning with bayesian neural networks,'' in \emph{Proc.
  International Conference on Learning Representations}, 2020.

\bibitem{zeno2018task}
C.~Zeno, I.~Golan, E.~Hoffer, and D.~Soudry, ``Task agnostic continual learning
  using online variational bayes,'' in \emph{Proc. Adv. Neural Information
  Processing Systems Workshop on Bayesian Deep Learning}, 2018.

\bibitem{zhai2019lifelong}
M.~Zhai, L.~Chen, F.~Tung, J.~He, M.~Nawhal, and G.~Mori, ``Lifelong gan:
  Continual learning for conditional image generation,'' in \emph{Proc. IEEE
  International Conference on Computer Vision}, 2019.

\bibitem{shmelkov2017incremental}
K.~Shmelkov, C.~Schmid, and K.~Alahari, ``Incremental learning of object
  detectors without catastrophic forgetting,'' in \emph{Proc. IEEE
  International Conference on Computer Vision}, 2017.

\bibitem{girshick2015fast}
R.~Girshick, ``Fast r-cnn,'' in \emph{IEEE International Conference on Computer
  Vision}, 2015.

\bibitem{hao2019end}
Y.~Hao, Y.~Fu, Y.-G. Jiang, and Q.~Tian, ``An end-to-end architecture for
  class-incremental object detection with knowledge distillation,'' in
  \emph{International Conference on Multimedia and Expo}, 2019.

\bibitem{ren2015faster}
S.~Ren, K.~He, R.~Girshick, and J.~Sun, ``Faster r-cnn: Towards real-time
  object detection with region proposal networks,'' in \emph{Proc. Adv. Neural
  Information Processing Systems}, 2015.

\bibitem{michieli2019incremental}
U.~Michieli and P.~Zanuttigh, ``Incremental learning techniques for semantic
  segmentation,'' in \emph{Proc. IEEE International Conference on Computer
  Vision Workshops}, 2019.

\bibitem{cermelli2020modeling}
F.~Cermelli, M.~Mancini, S.~R. Bulo, E.~Ricci, and B.~Caputo, ``Modeling the
  background for incremental learning in \mbox{semantic} segmentation,'' in
  \emph{IEEE Conference on Computer Vision and Pattern Recognition}, 2020.

\bibitem{tasar2019incremental}
O.~Tasar, Y.~Tarabalka, and P.~Alliez, ``Incremental learning for semantic
  segmentation of large-scale remote sensing data,'' \emph{IEEE Applied Earth
  Observations and Remote Sensing}, 2019.

\bibitem{ozdemir2018learn}
F.~Ozdemir, P.~Fuernstahl, and O.~Goksel, ``Learn the new, keep the old:
  Extending pretrained models with new anatomy and images,'' in
  \emph{International Conference on Medical Image Computing and
  Computer-Assisted Intervention}, 2018.

\bibitem{schak2019study}
M.~Schak and A.~Gepperth, ``A study on catastrophic forgetting~in deep lstm
  networks,'' in \emph{International Conference on Artificial Neural Networks},
  2019.

\bibitem{Coop2013fixedexpansion}
R.~Coop and I.~Arel, ``Mitigation of catastrophic forgetting in recurrent
  neural networks using a fixed expansion layer,'' in \emph{International Joint
  Conference on Neural Networks}, 2013.

\bibitem{chen2015net2net}
T.~Chen, I.~Goodfellow, and J.~Shlens, ``Net2net: Accelerating learning via
  knowledge transfer,'' in \emph{Proc. International Conference on Learning
  Representations}, 2016.

\bibitem{Sodhani2019rnnlll}
S.~Sodhani, S.~Chandar, and Y.~Bengio, ``Toward training recurrent neural
  networks for lifelong learning,'' \emph{Neural Computation}, 2019.

\bibitem{delchiaro2020RATT}
R.~Del~Chiaro, B.~Twardowski, A.~D. Bagdanov, and J.~Van~de Weijer, ``Ratt:
  Recurrent attention to transient tasks for continual image captioning,'' in
  \emph{International Conference on Machine Learning Workshop LifelongML},
  2020.

\bibitem{moore1993prioritized}
A.~W. Moore and C.~G. Atkeson, ``Prioritized sweeping: Reinforcement learning
  with less data and less time,'' \emph{Machine learning}, 1993.

\bibitem{krizhevsky2009learning}
A.~Krizhevsky, ``Learning multiple layers of features from tiny images,''
  Citeseer, Tech. Rep., 2009.

\bibitem{nilsback2008automated}
M.-E. Nilsback and A.~Zisserman, ``Automated flower classification over a large
  number of classes,'' in \emph{Indian Conference on Computer Vision, Graphics
  \& Image Processing}, 2008.

\bibitem{quattoni2009recognizing}
A.~Quattoni and A.~Torralba, ``Recognizing indoor scenes,'' in \emph{Proc. IEEE
  Conference on Computer Vision and Pattern Recognition}, 2009.

\bibitem{wah2011caltech}
C.~Wah, S.~Branson, P.~Welinder, P.~Perona, and S.~Belongie, ``The caltech-ucsd
  birds-200-2011 dataset,'' California Institute of Technology, Tech. Rep.
  CNS-TR-2011-001, 2011.

\bibitem{krause20133d}
J.~Krause, M.~Stark, J.~Deng, and L.~Fei-Fei, ``3d object representations for
  fine-grained categorization,'' in \emph{Proc. IEEE International Conference
  on Computer Vision Workshops}, 2013.

\bibitem{maji2013fine}
S.~Maji, J.~Kannala, E.~Rahtu, M.~Blaschko, and A.~Vedaldi, ``Fine-grained
  visual classification of aircraft,'' eprint arXiv:1306.5151, Tech. Rep.,
  2013.

\bibitem{yao2011human}
B.~Yao, X.~Jiang, A.~Khosla, A.~L. Lin, L.~Guibas, and L.~Fei-Fei, ``Human
  action recognition by learning bases of action attributes and parts,'' in
  \emph{Proc. IEEE International Conference on Computer Vision}, 2011.

\bibitem{cao2018vggface2}
Q.~Cao, L.~Shen, W.~Xie, O.~M. Parkhi, and A.~Zisserman, ``Vggface2: A dataset
  for recognising faces across pose and age,'' in \emph{International
  Conference on Automatic Face \& Gesture Recognition}, 2018.

\bibitem{russakovsky2015imagenet}
O.~Russakovsky, J.~Deng, H.~Su, J.~Krause, S.~Satheesh, S.~Ma, Z.~Huang,
  A.~Karpathy, A.~Khosla, M.~Bernstein \emph{et~al.}, ``Imagenet large scale
  visual recognition challenge,'' \emph{International Journal of Computer
  Vision}, 2015.

\bibitem{he2016deep}
K.~He, X.~Zhang, S.~Ren, and J.~Sun, ``Deep residual learning for image
  recognition,'' in \emph{Proc. IEEE Conference on Computer Vision and Pattern
  Recognition}, 2016.

\bibitem{simonyan2014very}
K.~Simonyan and A.~Zisserman, ``Very deep convolutional networks for
  large-scale image recognition,'' in \emph{Proc. International Conference on
  Learning Representations}, 2015.

\bibitem{szegedy2015going}
C.~Szegedy, W.~Liu, Y.~Jia, P.~Sermanet, S.~Reed, D.~Anguelov, D.~Erhan,
  V.~Vanhoucke, and A.~Rabinovich, ``Going deeper with convolutions,'' in
  \emph{IEEE Conference on Computer Vision and Pattern Recognition}, 2015.

\bibitem{sandler2018mobilenetv2}
M.~Sandler, A.~Howard, M.~Zhu, A.~Zhmoginov, and L.-C. Chen, ``Mobilenetv2:
  Inverted residuals and linear bottlenecks,'' in \emph{Proc. IEEE Conference
  on Computer Vision and Pattern Recognition}, 2018.

\bibitem{chaudhry2021using}
A.~Chaudhry, A.~Gordo, P.~K. Dokania, P.~Torr, and D.~Lopez-Paz, ``Using
  hindsight to anchor past knowledge in continual learning,'' in \emph{Proc.
  AAAI Conference Artificial Intelligence}, 2021.

\bibitem{liu2020generative}
X.~Liu, C.~Wu, M.~Menta, L.~Herranz, B.~Raducanu, A.~D. Bagdanov, S.~Jui, and
  J.~van~de Weijer, ``Generative feature replay for class-incremental
  learning,'' in \emph{Proc. IEEE Conference on Computer Vision and Pattern
  Recognition Workshops}, 2020.

\bibitem{iscen2020memory}
A.~Iscen, J.~Zhang, S.~Lazebnik, and C.~Schmid, ``Memory-efficient incremental
  learning through feature adaptation,'' in \emph{Proc. European Conference on
  Computer Vision}, 2020.

\bibitem{rao2019continual}
D.~Rao, F.~Visin, A.~Rusu, R.~Pascanu, Y.~W. Teh, and R.~Hadsell, ``Continual
  unsupervised representation learning,'' in \emph{Proc. Adv. Neural
  Information Processing Systems}, 2019.

\bibitem{jing2020selfsuper}
L.~Jing and Y.~Tian, ``Self-supervised visual feature learning with deep neural
  networks: A survey,'' \emph{IEEE Transactions on Pattern Analysis and Machine
  Intelligence}, 2020.

\bibitem{yu2020semantic}
L.~Yu, B.~Twardowski, X.~Liu, L.~Herranz, K.~Wang, Y.~Cheng, S.~Jui, and
  J.~v.~d. Weijer, ``Semantic drift compensation for class-incremental
  learning,'' in \emph{Proc. IEEE Conference on Computer Vision and Pattern
  Recognition}, 2020.

\bibitem{li2020energy}
S.~Li, Y.~Du, G.~M. van~de Ven, A.~Torralba, and I.~Mordatch, ``Energy-based
  models for continual learning,'' in \emph{Proc. International Conference on
  Learning Representations}, 2021.

\bibitem{schmidhuber1987evolutionary}
J.~Schmidhuber, ``Evolutionary principles in self-referential learning,''
  \emph{Diploma thesis, Tech. Univ. Munich}, 1987.

\bibitem{javed2019meta}
K.~Javed and M.~White, ``Meta-learning representations for continual
  learning,'' in \emph{Proc. Adv. Neural Information Processing Systems}, 2019.

\bibitem{aljundi2019task}
R.~Aljundi, K.~Kelchtermans, and T.~Tuytelaars, ``Task-free continual
  learning,'' in \emph{IEEE Conference on Computer Vision and Pattern
  Recognition}, 2019.

\bibitem{rajasegaran2020itaml}
J.~Rajasegaran, S.~Khan, M.~Hayat, F.~S. Khan, and M.~Shah, ``{iTAML}: An
  incremental task-agnostic meta-learning approach,'' in \emph{Proc. IEEE
  Conference on Computer Vision and Pattern Recognition}, 2020.

\bibitem{de2020continual}
M.~De~Lange and T.~Tuytelaars, ``Continual prototype evolution: Learning online
  from non-stationary data streams,'' in \emph{Proc. IEEE International
  Conference on Computer Vision}, 2020.

\bibitem{kunstner2019limitations}
F.~Kunstner, L.~Balles, and P.~Hennig, ``Limitations of the empirical fisher
  approximation for natural gradient descent,'' in \emph{Proc. Adv. Neural
  Information Processing Systems}, 2019.

\bibitem{torralba2008tiny}
A.~{Torralba}, R.~{Fergus}, and W.~T. {Freeman}, ``80 million tiny images: A
  large data set for nonparametric object and scene recognition,'' \emph{IEEE
  Transactions on Pattern Analysis and Machine Intelligence}, 2008.

\bibitem{tinyIM}
\BIBentryALTinterwordspacing
Stanford. (CS231N) Tiny imagenet challenge, cs231n course. [Online]. Available:
  \url{https://tiny-imagenet.herokuapp.com/}
\BIBentrySTDinterwordspacing

\bibitem{van2018inaturalist}
G.~Van~Horn, O.~Mac~Aodha, Y.~Song, Y.~Cui, C.~Sun, A.~Shepard, H.~Adam,
  P.~Perona, and S.~Belongie, ``The inaturalist species classification and
  detection dataset,'' in \emph{Proc. IEEE Conference on Computer Vision and
  Pattern Recognition}, 2018.

\bibitem{masana2020onclass}
M.~Masana, B.~Twardowski, and J.~Van~de Weijer, ``On class orderings for
  incremental learning,'' in \emph{International Conference on Machine Learning
  Workshop on Continual Learning}, 2020.

\bibitem{bengio2009curriculum}
Y.~Bengio, J.~Louradour, R.~Collobert, and J.~Weston, ``Curriculum learning,''
  in \emph{Proc. International Conference on Machine Learning}, 2009.

\bibitem{sergey2016wide}
S.~Zagoruyko and N.~Komodakis, ``Wide residual networks,'' in \emph{Proc.
  British Machine Vision Conference}, 2016.

\end{thebibliography}

%%%%%%%%%%%%%%%%%%%%%%%%%%%%%%%%%%%%%
\newpage
\begin{IEEEbiographynophoto}{\footnotesize Marc~Masana} \footnotesize obtained his Ph.D. in computer vision at the Computer Vision Center from Universitat Autònoma de Barcelona, Spain in 2020 and is currently holding a post-doc position at TU Graz, Austria. His research interests concern machine learning, network compression, continual learning and novelty detection.
\end{IEEEbiographynophoto}
\vspace{-10mm}
\begin{IEEEbiographynophoto}{\footnotesize Xialei~Liu} \footnotesize is currently an associate professor at Nankai University, China. Before that, he was a post-doc research associate at University of Edinburgh, UK. He obtained his PhD in 2020 from Universitat Autònoma de Barcelona, Spain. His research interests include lifelong learning, self-supervised learning, few-shot learning and long-tailed learning.
\end{IEEEbiographynophoto}
\vspace{-10mm}
\begin{IEEEbiographynophoto}{\footnotesize Bart{\l}omiej~Twardowski} \footnotesize obtained his Ph.D. in computer science from Warsaw University of Technology in 2018. Currently holding a post-doc position in Computer Vision Center, Universitat Autònoma de Barcelona.
\end{IEEEbiographynophoto}
\vspace{-10mm}
\begin{IEEEbiographynophoto}{\footnotesize Mikel~Menta} \footnotesize obtained his M.Sc. in Computer Vision from Universitat Autònoma de Barcelona in 2018. Currently working as an applied researcher for Wide Eyes Technologies.
\end{IEEEbiographynophoto}
\vspace{-10mm}
\begin{IEEEbiographynophoto}{\footnotesize Andrew~D.~Bagdanov} \footnotesize Andrew D. Bagdanov is Associate Professor at the University of Florence, Italy. He received the Ph.D. from the University of Amsterdam in 2004, after which he was a postdoc at the University of Florence and the Universitat Autònoma de Barcelona, and was a Ramon y Cajal Fellow at the Computer Vision Center, Barcelona. His research spans a broad spectrum of computer vision, image processing and machine learning.
\end{IEEEbiographynophoto}
\vspace{-10mm}
\begin{IEEEbiographynophoto}{\footnotesize Joost~van~de~Weijer} \footnotesize Joost van de Weijer received the Ph.D. degree from the University of Amsterdam in 2005. He was a Marie Curie Intra-European Fellow at INRIA Rhone-Alpes, France, and from 2008 to 2012 was a Ramon y Cajal Fellow at the Universitat Autònoma de Barcelona, Spain, where he is currently a Senior Scientist at the Computer Vision Center and leader of the LAMP Team.
\end{IEEEbiographynophoto}
\vfill
%%%%%%%%%%%%%%%%%%%%%%%%%%%%%%%%%%%%%

% Tables and figures for supplementary start with S
\renewcommand{\thetable}{S\arabic{table}}
\renewcommand{\thefigure}{S\arabic{figure}}
\setcounter{table}{0}
\setcounter{figure}{0}

\appendices
%%%%%%%%%%%%%%%%%%%%%%%%%%%%%%%%%%%%%
\section{Implementation and hyperparameters}
\label{app:hyperparams}
We study the effects of CL methods for image classification on nine different datasets whose statistics are summarized in Table~\ref{tab:datasets}. CIFAR-100 contains $32 \times 32$ colour images for 100 classes, with 600 samples for each class divided into 500 for training and 100 for testing. For data augmentation, a padding of 4 is added to each side, and crops of $32 \times 32$ are randomly selected during training and the center cropped is used during testing. For all datasets except \mbox{CIFAR-100}, images are resized to $256 \times 256$ with random crops of $224 \times 224$ for training and center crops for testing. Input normalization and random horizontal flipping are performed for all datasets.

As described in Section~5.5, the Continual Hyperparameter Framework (CHF)~\cite{de2020continual} is used for the stability-plasticity trade-off hyperparameters that are associated to intransigence and forgetting when learning a new task. The CHF first performs a learning rate (LR) search with Finetuning on the new task. This corresponds to the \textit{Maximal Plasticity Search} phase.

The LR search is limited to \{5e-1, 1e-1, 5e-2\} on the first task since all experiments are trained from scratch. For the remaining tasks, the LR search is limited to the three values immediately lower than the one chosen for the first task from this set: \{1e-1, 5e-2, 1e-2, 5e-3, 1e-3\}. We use a patience scheme as a LR scheduler where the patience is fixed to 10, the LR factor to 3 (LR is divided by it each time the patience is exhausted), and the stopping criteria is either having a LR below 1e-4 or if 200 epochs have passed (100 for VGGFace2 and ImageNet). We also do gradient clipping at 10,000, which is mostly negligible for most training sessions except the first one. We use SGD with momentum set to 0.9 and weight decay fixed to 0.0002. Batch size is 128 for most experiments except 32 for fine-grained datasets and 256 for ImageNet and VGGFace2. All code is implemented using Pytorch.

Once the shared hyperparameters are searched, the best ones are fixed and the accuracy for the first phase is stored as a reference. The hyperparameter directly related to the stability-plasticity trade-off is set to a high value which represents a heavy intransigence to learn the new task, close to freezing the network so that knowledge is preserved. At each search step, the performance is evaluated on the current task and compared to the reference accuracy from the \textit{Maximal Plasticity Search} phase. If the method accuracy is above the 80\% of the reference accuracy, we keep the model and trade-off as the ones for that task. If the accuracy is below the threshold, the trade-off is reduced in half and the search continues. As the trade-off advances through the search, it becomes less intransigence and slowly converges towards higher forgetting, which ultimately would correspond to the Finetuning of the previous phase. This corresponds to the \textit{Stability Decay} phase.
\smallskip

\tabledatasets{Summary of datasets used. We use a random 10\% from the train set for validation.}

\noindent The methods have the following implementations:
\begin{itemize}
    \item \textbf{LwF}: we implement the $\mathcal{L}_{dis}$ distillation loss following \mbox{Eqs. 5-6}, and fix the temperature scaling parameter to $T=2$ as proposed in the original work (and used in most of the literature). When combining LwF with examplars the distillation loss is also applied to the exemplars of previous classes~\cite{castro2018end, hou2019learning, rebuffi2017icarl, wu2019large}. This loss is combined with the $\mathcal{L}_{c}$ cross-entropy loss from Eqs. 2-3 with a trade-off that is chosen using the CHF and starts with a value of 10. In our implementation we choose to duplicate the older model for training to evaluate the representations (instead of saving them at the end of the previous session) to benefit from the data augmentation. That older model can be removed after the training session to avoid overhead storage.
    \item \textbf{EWC}: the fusion of the old and new importance weights is done with $\alpha=0.5$ (chosen empirically) to avoid the storage of the importance weights for each task. The Fisher Information Matrix (FIM) is calculated by using all samples from the current task and is based on the predicted class. Referring to the definitions from~\cite{kunstner2019limitations}, we have implementations of the empirical and real Fisher Information Matrix (FIM) in our experimental framework, the difference being using either a fixed label or sampling from the model’s predictive distribution when computing the FIM. In the manuscript we report results for EWC using the FIM estimated using the maximum probability class output, which is a variant described in~\cite{vandeven2019three}. The loss introduced in Eq. 4 is combined with the $\mathcal{L}_{c}$ cross-entropy loss with a trade-off chosen using the CHF and with a starting value of 10,000.
    \item \textbf{PathInt}: we fix the damping parameter to 0.1 as proposed in the original work. As in LwF and EWC, the trade-off between the quadratic surrogate loss and the cross-entropy loss is chosen using the CHF with a starting value of 1.
    \item \textbf{MAS}: we implement MAS in the same way as EWC, with $\alpha=0.5$ and the same Fisher Information Matrix setting. The trade-off between the importance weights penalty and the cross-entropy loss is chosen using the CHF and a starting value of 400.
    \item \textbf{RWalk}: since it is a fusion of EWC and PathInt, the same parameters $\alpha=0.5$, Fisher Information Matrix setting and $\text{damping}=0.1$ are fixed. The starting value for the CHF on the trade-off between their proposed objective loss and the cross-entropy loss is 10.
    \item \textbf{DMC}: we implement the $\mathcal{L}_{DD}$ double distillation loss from Eqs. 10-11. We use a $32\times32$ resized version of Imagenet as auxiliary dataset and set its batch size to 128. The student is neither initialized from the previous tasks or new task models but random, as proposed in the original work.
    \item \textbf{GD}: we implement both the training and the sampling algorithms as described in~\cite{lee2019overcoming}. Due to the withdrawal of Tiny Images \cite{torralba2008tiny} (the auxiliary dataset used in the original implementation), we use a $32\times32$ resized Imagenet as auxiliary dataset and we set its batch size to 128.
    \item \textbf{GDumb}: we use random sampling to select exemplars, reinitialize the model and apply cutmix regularization, as described in~\cite{prabhu2020gdumb}.
    \item \textbf{LwM}: we combine the cross-entropy loss with the distillation loss and $\mathcal{L}_{AD}$ attention distillation using the $\beta$ and $\gamma$ trade-offs respectively. The $\beta$ trade-off is the one that balances the stability-plasticity dilemma and we chose it using the CHF with a starting value of 2. The $\gamma$ trade-off is fixed to 1 since it does not directly affect the stability-plasticity dilemma. Since there is no mention in the original work on which are the better values to balance the three losses, that last value was chosen after a separate test with values $\gamma\in (0,2]$ and fixed for all scenarios in Section~6.
    \item \textbf{iCaRL}: we implement the five algorithms that comprise iCaRL. The distillation loss is combined with the cross-entropy loss during the training sessions and chosen using the CHF with a starting value of 4. However, during evaluation, the NME is used instead of the softmax outputs.
    \item \textbf{EEIL}: we implement EEIl with the balanced and unbalanced training phases. The unbalanced phase uses the hyperparameters shared across all methods. However, for the balanced phase the LR is reduced by 10 and the number of training epochs to 40. As with LwF, $T=2$ and the trade-off is chosen using the CHF starting at 10. However, we apply a slight modification to the original work by not using the addition of noise to the gradients. Our preliminary results with this method showed that it was consistently detrimental to performance, which provided a worse representation of the capabilities of the method.
    \item \textbf{BiC}: the distillation stage is implemented the same as LwF, as in the original paper, with $T=2$. However, the trade-off between distillation and cross-entropy losses is not chosen using the CHF. The authors already propose to set it to $\frac{n}{n+m}$, where $n$ is the number of previous classes, and $m$ is the number of new classes, and we keep that decision. On the bias correction stage, also following the original work, we fix the percentage of validation split used from the total amount of exemplar memory to be 10\%.
    \item \textbf{LUCIR}: for this method we make two changes on the architecture of the model. First, we replace the classifier layer by a cosine normalization layer following Eq. 14; and second we remove the ReLU from the penultimate layer to allow features to take both positive and negative values. However, since this procedure is only presented in the original work for ResNet models, we do not extend it to other architectures. The original code used a technique called imprint weights during the initialization of the classifier. However, since it was not mentioned in the original paper, and preliminary experiments showed no significant difference, we decided to not include it in our implementation.
    
    The cross-entropy loss is combined with the $\mathcal{L}_{lf}$ less-forget constraint from Eq. 12 and the $\mathcal{L}_{mr}$ margin ranking loss from Eq. 15. The number of new class embeddings chosen as hard negatives and the margin threshold are fixed to $K=2$ and $m=0.5$ as in the original work. The margin ranking loss is combined with the cross-entropy loss in a one-to-one ratio, while the less-forget constraint is chosen using the CHF with a starting value of 10, as is the trade-off related to the stability-plasticity dilemma.
    \item \textbf{IL2M}: since it only stores some statistics on the classes and applies them after the training is done in the same way as Finetuning, there is no hyperparameter to tune for this method.
\end{itemize}

Finally, the Finetuning, Freezing and Joint training baselines have no hyperparameters associated to them, reducing the Continual Hyperparameter Framework to only performing the learning rate search for each task before doing the final training.

%%%%%%%%%%%%%%%%%%%%%%%%%%%%%%%%%%%%%
\section{Supplemental results}
\label{app:suppl-results}
\subsection{More on CIFAR-100}
\label{app:more-cifar}
Experiments on CIFAR-100 are evaluated on 10 fixed random seeds that are the same for all approaches, to make the comparison fair. In Table~\ref{tab:more-cifar}, we show mean and standard deviation for average accuracy after learning the last task on the CIFAR-100 scenarios from Sec.~\ref{sec:exemplar-exps},~\ref{sec:scenarios-exps} and~\ref{sec:online}. For most approaches and scenarios, the standard deviation seems to be below 2.5. However, some regularization-based methods (MAS-E, PathInt-E, RWalk) and iCaRL, seem to have much more variation when used in the initial larger task scenario. In the case of regularization-based methods, some runs struggle to learn new tasks properly after the initial ones, obtaining quite low performance and therefore resulting in high variability in the results. In the case of iCaRL, the variability seems to be related on how well the output features do on the initial task, since performance stays quite stable on the remaining ones. It is also notable that among the bias-correction methods, IL2M is the more stable one. We also provide results for the online class-IL approaches (GDumb, ER and MIR) applied to our offline scenarios. We allow those online methods to revisit the data of a task for several epochs to be competitive in our proposed scenarios. As shown in the original paper~\cite{prabhu2020gdumb}, GDumb performs worse than most class-IL approaches for offline class-IL. ER and MIR show results similar to those of FT-E on their offline version.

\tabmorecifar{Mean and standard deviation of average accuracy over 10 runs for different CIFAR-100 scenarios.}

Next to the fixed memory scenario evaluated in the main paper, we here also provide results for the growing memory scenario (20 examplars per class) with herding as the sampling strategy. In Figure~\ref{fig:cifar100_grow_plot_cifar100_grow_lft_plot}, GD, BiC, EEIL and IL2M achieve the best results after learning 10 tasks with a growing memory, just as it was for a fixed memory. In general, most methods seem to suffer less catastrophic forgetting when using a fixed memory that allows storing more exemplars during early tasks. That is the case for BiC, GD and LUCIR, which have much better performance with a fixed memory. For some approaches, the difference is quite considerable after learning 5 tasks and slightly better after the full 10-task sequence.

\figExpDoubleColumnTwoImage{0.48}{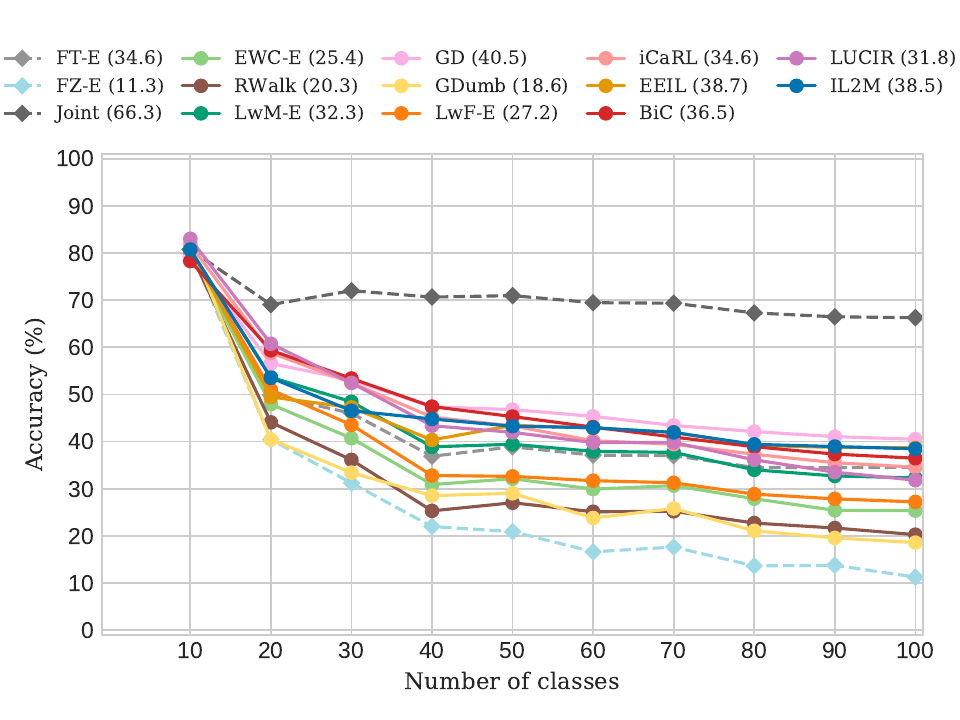}{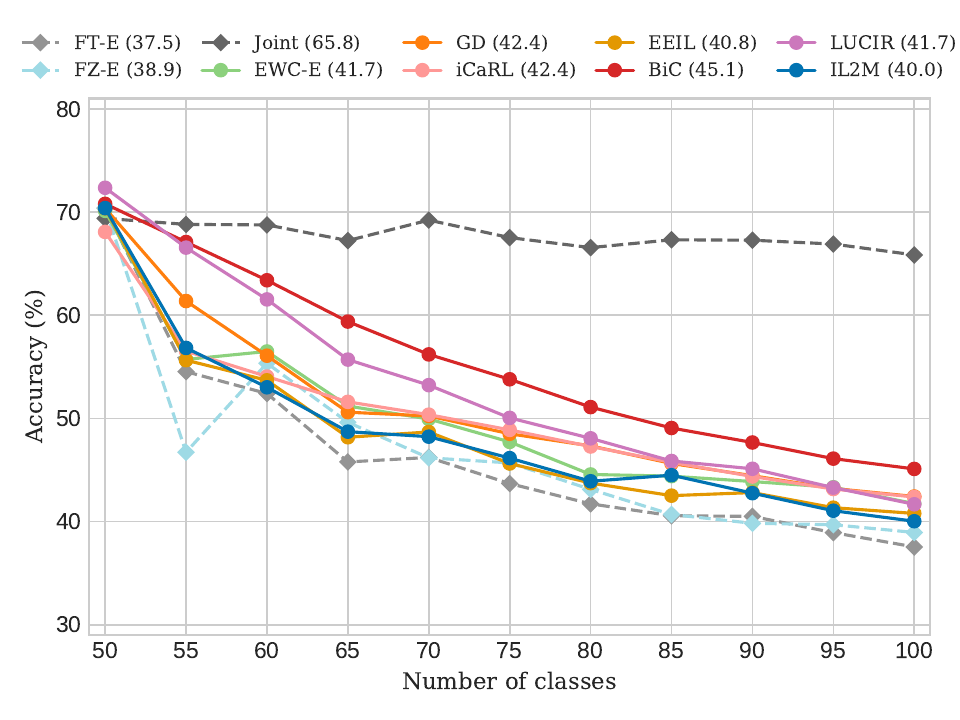}{\mbox{CIFAR-100} (10/10) with 20 exemplars per class growing memory (left), and \mbox{CIFAR-100} (11/50-5) with 20 exemplars per class growing memory (right).}

\subsection{On Random Path Selection}
\label{app:rps_networks}
Although Random Path Selection (RPS)~\cite{rajasegaran2019random} is not a fixed network architecture approach, it is one of the better performing methods from the dynamic architectures family. In Table~\ref{tab:rpsnets} we provide a comparison with different number of paths and a range of baselines. The original \mbox{CIFAR-100} (10/10) experiment was proposed with a variation of ResNet-18, however, to make it comparable with the experiments in Sec.~\ref{sec:results} we compare it using a customized ResNet-32, which is even more memory efficient than the original ResNet-18 with 3.72M instead of 89.56M parameters. As expected, performance decreases when reducing the number of paths, making this approach very competitive if memory restrictions for the network are not an issue. With a comparable network size, it becomes less competitive in comparison to other approaches such as finetuning with exemplars (FT-E). We also report the average time per epoch for all tasks, and it is clear that the original RPS with ResNet-18 computational cost is much larger than other methods. When we change the network to ResNet-32 (with significantly fewer parameters), both performance and running time reduce dramatically, but the running time is still much more than other methods due to the execution of different paths in parallel.

\figrpsnets{Comparison of Random Path Selection (RPS) on \mbox{CIFAR-100} (10/10) with fixed 2,000-exemplar memory.}

\figClassOrdering

\subsection{On semantic tasks}
\label{app:semantic_tasks}
The popularity of iCaRL and the interest in comparing with it makes it quite common to utilize the random class ordering for experiments based on \mbox{CIFAR-100}~\cite{krizhevsky2009learning}. The authors of iCaRL use a random order of classes which is fixed in the iCaRL code by setting the random seed to 1993 just before shuffling the classes. However, this gives very little insight on class orderings which make use of the coarse labels from that dataset to group classes into sharing similar semantic concepts. This was explored for the tinyImageNet (Stanford, CS231N~\cite{tinyIM}) dataset in~\cite{de2020continual, masana2020ternary}, where the authors show that some methods report different results based on different semantics-based class orderings. In~\cite{de2020continual}, the iNaturalist~\cite{van2018inaturalist} dataset is split into tasks according to supercategories and are ordered using a relatedness measure. Having tasks with different semantic distributions and learning tasks in different orders is interesting for real-world applications where subsequent tasks are based on correlated data instead of fully random. Recently,~\cite{masana2020onclass} also brings attention to the learning variability between using different class orderings when learning a sequence of tasks incrementally.

In joint training, specific features in the network can be learned that focus on differentiating two classes that are otherwise easily confused. However, in an IL setting those discriminative features become more difficult to learn or can be modified afterwards, especially when the classes belong to different tasks. Thus, the difficulty of the task can be perceived differently in each scenario. Depending on the method, this issue may be handled differently and therefore lead to more catastrophic forgetting. This setting is different from the one proposed in Curriculum Learning~\cite{bengio2009curriculum}, since the objective here is not to find the best order to learn tasks efficiently, but rather to analyze incremental learning settings (in which the order is not known in advance) and analyze the robustness of methods under different task orderings.

In order to investigate robustness to class orderings, we use the 20 coarse-grained labels provided in the \mbox{CIFAR-100} dataset to arrive at semantically similar groups of classes. Then, we order these groups based on their classification difficulty. To assess performance we trained a dedicated model with all \mbox{CIFAR-100} data in a single training session and use this model accuracy as a proxy value for classification difficulty. Finally, we order them from easier to harder (Dec. Acc.) and the other way around (Inc. Acc.). Results are presented in Fig.~\ref{fig:class-ordering-exp} for two methods without exemplars (FT+, LwF), and two methods with exemplars (FT-E, BiC). Performance can be significantly lower when using a semantics-based ordering compared to random one. In the examplar-free cases, special care of the used task ordering should be taken as the final performance after learning all classes can have quite some variability as seen in the LwF case. However, the variation with respect to the orderings is mitigated by the use of exemplars. Therefore, evaluating methods which use exemplars with randomized task orderings often suffices.

\tabexemplarsamplinggrowing

\tabexemplarsamplingstats

\subsection{More on sampling strategies}
\label{app:sampling_strats}
The performance achieved by the different sampling methods is very similar in the CIFAR-100 (11/50-5) scenario. As seen in Table~\ref{tab:ex-sampling-2}, for longer task sequences herding has a slight benefit over the other sampling strategies when using class-incremental learning methods. In the case of shorter sequences, similar to transfer learning, performance does not seem to specifically favour any sampling strategy. We also add a variation of entropy and distance sampling which chooses the samples furthest away from the task boundaries to observe the effect of choosing the least confusing samples instead. We denote these as \emph{inv-entropy} and \emph{inv-distance}. It is notable that for shorter task sequences, entropy- and distance-based perform similar to the proposed inverse versions. However, for larger sequences of tasks, the inverse versions perform better. This could be due to samples further away from the boundaries (and closer to the class centers) becoming more relevant when the number of classes and their diversity increases.

To provide further context on the results between the different sampling strategies, we extend Table~\ref{tab:ex-sampling} with the standard deviations in Table~\ref{tab:ex-sampling-3}. Furthermore, we apply a Mann-Whitney U-test to compare random and herding on the different sequence lengths to verify if the slight performance increase of herding is significant or not. Results are shown in Table~\ref{tab:sampling-stats}, which indicate that there is no significant difference between using herding or random for exemplar selection in most cases. However, there is a tendency for longer sequences to be more favourable towards herding, with methods LwF-E and EEIL being significantly better when using herding for 10 task sequences.

\tabsamplingstats{Mann–Whitney U-test between random and herding sampling strategies from Table~\ref{tab:ex-sampling-3}.}

\subsection{More on external data}
\label{app:external_data}
DMC~\cite{zhang2020class} uses external data instead of exemplars and GD~\cite{lee2019overcoming} uses both external data and exemplars. For GD we use the convention that $\text{GD}^*$ refers to the version with external data, and when using GD we replace the external data with the standard exemplar memory. Same for $\text{DMC}^*$ and DMC. We use the 300 classes from ImageNet-32 as external data. The results are included in Table~\ref{tab:more-cifar}. $\text{DMC}^*$ obtains an average accuracy of 25.9\% on CIFAR-100 (10/10). However, if the auxiliary data is not used, and it uses exemplars instead, performance drops to 20.6\%. The method provides privacy-preserving properties at the cost of some performance. $\text{GD}^*$, which also uses exemplars, obtains the excellent result of 44.6\%, outperforming all methods. However, the gain with respect to GD without exemplars is relatively small (which obtained 43.7\%). In conclusion, we found that the gain obtained by distillation from an additional dataset is rather small.

\subsection{More on small domain shifts}
\label{app:smalldomain}
For the experiment in Sec.~\ref{sec:domain-shift} on smaller domain shifts, we extend the results shown in Fig.~\ref{fig:vggface2} in Table~\ref{tab:smalldomain}. Interestingly, in this setting, weight regularization method EWC-E (70.0\%) outperforms data regularization method LwF-E (43.5\%) by a very large margin.

\tablesmalldomain{Additional results for Fig.~\ref{fig:vggface2}. Small domain shifts on VGGFace2 (40/25) on ResNet-18 trained from scratch amd 5,000 exemplar fixed memory.}

\subsection{More on network architectures}
We have selected the networks in the experiment from Sec.~\ref{sec:exp-networks} to represent a wide variety of network architectures commonly used in deep learning, allowing us to compare them within a continual learning setting. We have chosen AlexNet and VGG-11 as architectures which start with a number of initial convolutional layers followed by several fully connected layers. ResNets have achieved superior performance in many different computer vision tasks, and we therefore consider ResNet-18. We have also included GoogleNet which uses skip-connection and $1 \times 1$ convolutions are used as a dimension reduction module to reduce computation. We are also interested to evaluate incremental learning on compact networks. We have therefore selected MobileNet, which, to better trade off latency and accuracy, propose to replace standard convolution layers by depthwise separable convolutions. This makes them suitable to run on mobile devices.

We provide more detailed results on the experiment with different architectures in Table~\ref{tab:diff-nets}. Each network architecture is evaluated on the same 10 random seeds and the accuracy and forgetting presented is an average of those runs. We also include WideResNet-50~\cite{sergey2016wide} together with the ones presented in Fig.~\ref{fig:interspersed_forg}. BiC exhibits the least forgetting among all methods, even having positive forgetting which indicates that performance improves on some tasks after learning subsequent ones. However, this result comes at the expense of having slightly lower performance for each task right after learning them.

\tabdifferentnetworks{ImageNet-Subset-100 (10/10) with different networks trained from scratch. Task accuracy when the task was learned and forgetting after learning all classes (between brackets). Final column reports the average accuracy after 10 tasks.}

\subsection{More on large-scale scenarios}
In class-IL it has become more common to report results in large-scale datasets such as ImageNet-1000. Some works~\cite{hou2019learning, wu2019large} prefer to report in shorter sequences such as ImageNet (10/100) or ImageNet (11/500-50), thus we include the results on these scenarios in Tables~\ref{tab:more-imagenet} and~\ref{tab:more-imagenet-large} respectively.

\tabmoreimagenet{Average accuracy over learned tasks for ImageNet (10/100) on ResNet-18 from scratch with growing memory of 20 exemplars per class.}

\tabmoreimagenetlarge{Average accuracy over learned tasks for ImageNet (11/500-50) on ResNet-18 from scratch with growing memory of 20 exemplars per class.}

\end{document}